\documentclass{article} % For LaTeX2e
\usepackage{float}
\usepackage{iclr2027_conference,times}

\AtBeginDocument{%
  \renewcommand{\headrulewidth}{0pt}%
}

\usepackage{amsmath,amsfonts,bm}

\def\eqref#1{equation~\ref{#1}}
\def\1{\bm{1}}

\DeclareMathAlphabet{\mathsfit}{\encodingdefault}{\sfdefault}{m}{sl}
\SetMathAlphabet{\mathsfit}{bold}{\encodingdefault}{\sfdefault}{bx}{n}

\usepackage{hyperref}
\usepackage{url}

\usepackage{booktabs}
\usepackage{multirow}
\usepackage{graphicx}
\usepackage{cleveref}
\usepackage{wrapfig}
\usepackage{pifont}
\newcommand{\cmark}{\ding{51}}
\newcommand{\xmark}{\ding{55}}
\usepackage{titletoc}  % Required for partial/isolated TOCs
\usepackage{enumitem}
\usepackage{float}

\title{GeoSCope: Grounded, Stable and \\Controllable Driving World Models}

\author{%
  \textbf{Marius Kästingschäfer}$^{1,2}$
  \quad
  \textbf{Sudhanshu Mittal}$^{2}$
  \quad
  \textbf{Sebastian Bernhard}$^{3}$
  \quad
  \textbf{Thomas Brox}$^{2}$\\
  $^1$Aumovio SE
  \quad
  $^2$University of Freiburg
  \quad
  $^3$Technische Hochschule Augsburg\\
  \texttt{marius.kaestingschaefer@gmail.com}
}

\usepackage[table]{xcolor}
\definecolor{oursrow}{gray}{0.93}

\iclrfinalcopy % Uncomment for camera-ready version, but NOT for submission.
\begin{document}

\maketitle
\vspace{-0.4cm}
\begin{abstract}
Generative driving world models produce realistic video but lack grounding in a real location. Conditioned on their own predictions, they modify road layout and surrounding structures, causing long-horizon rollouts to diverge from the recorded scene. A policy evaluated in such rollouts is effectively tested on a different scene than the one recorded. We propose GeoSCope, a method that grounds driving world models in the geometry of a real route using only a single monocular recording. A dynamics-aware SLAM pipeline reconstructs a static 3D map of the route, which is rendered at the camera pose of each frame prior to its generation and used to condition a generative world model. Although the rendered signal is sparse and imperfect, the model learns to exploit it selectively. Experiments with three world models on four datasets show that this anchoring substantially reduces long-horizon drift and preserves scene structure without suppressing dynamic objects. Since the map can be rendered along arbitrary trajectories for conditioning, the same model supports counterfactual rollouts, turning existing video-based world models into grounded, stable, and controllable neural simulators. With geometric conditioning trained only on nuPlan, the approach transfers zero-shot to Waymo, MARS and nuScenes, and sets a new state of the art on the Ann-Arbor-City-Bench multi-traversal benchmark. Code and weights will be released soon.
\end{abstract}

\vspace{-0.25cm}
\section{Introduction}
\label{introduction}
\vspace{-0.1cm}

% The need for faithful simulation.
Closed-loop training and evaluation of driving policies require a simulator that reacts to the policy's actions while remaining faithful to the real world. 
Reconstruction-based neural simulators, including those based on 3D Gaussian splatting, provide explicit geometry and direct camera control.  However, their quality deteriorates under large deviations from the observed trajectory, many require auxiliary annotations such as 3D bounding boxes or object tracks~\citep{chen2025omnire, yan2024street}, and dynamics must be modeled through separate components. Generative world models offer the complementary strength: they learn appearance, dynamics and interactions end-to-end from video. Yet nothing ties what they generate to a real place because they are not grounded.

% Why missing grounding matters
Without a persistent scene reference, a generative world model is free to hallucinate the static environment, reinventing lane geometry, intersections and buildings as the rollout proceeds. 
%Exposure bias makes this worse. The model is trained on clean context but at inference conditions on its own imperfect predictions, so errors accumulate and the rollout drifts out of distribution over long horizons.
Because each new frame is predicted from the model's own previous outputs, nothing in the loop can correct such deviations and they compound over long horizons.
This is not merely a visual-quality issue, since a prediction can look realistic without being faithful to the real scene layout. When evaluating a driving policy, the static environment is what the policy is being tested against, so it has to remain true to the real scene. If the road bends to accommodate a lateral deviation, or an intersection disappears after a few seconds, the policy is no longer evaluated on the logged scene. A learning policy may even exploit such artifacts, and its failures can no longer be told apart from failures of the simulator. One way to ground generation is to condition on HD maps~\citep{wang2023drivedreamer, gao2023magicdrive}. This fixes the road layout, but HD maps need precise annotations that are expensive to create and maintain, which is why most raw driving logs do not have them.

% Our appraoch
We introduce GeoSCope, a \textbf{G}eometry-grounded, \textbf{S}table, and \textbf{C}ontrollable method for anchoring a generative world model in the real geometry of the traversed route using only a single monocular video driving log, without LiDAR or HD maps. A dynamics-aware SLAM pipeline first reconstructs a filtered static point cloud of the route. 
At each rollout step, we render this reconstruction at the requested pose of the frame about to be generated and use the resulting pixel-aligned image to condition a pre-trained world model.  
We refer to this as prospective geometric anchoring, since the reconstruction is queried at the upcoming target pose rather than aligned only with past or current viewpoints. In this setup, the reconstruction fixes the static environment wherever it was observed, while the generative prior completes sparse and uncovered regions and generates the dynamic agents that the static reconstruction does not contain.
The mechanism assumes nothing about the backbone beyond an autoregressive rollout and a latent encoder, so it applies to existing world models without architectural redesign. We demonstrate this on three models that differ in scale and architecture, namely Orbis v1, the hierarchical Orbis v2, and Cosmos-Predict2.5.

Anchoring to a persistent reconstruction grounds generation in the real layout of the route, which substantially reduces structural drift and better preserves perceptual quality over long horizons. Across base models and datasets, our approach roughly doubles long-horizon keypoint matches and reduces long-horizon FVD and depth error substantially, without suppressing the dynamic content that the generative prior supplies. 
Beyond stabilizing the scene, the same signal provides direct control over the ego trajectory: the reconstruction can be rendered along any trajectory through the observed region, so the model can also roll out counterfactual variants of a logged route. Beyond the reconstructed region, it falls back to its learned prior and degrades gracefully toward the base model.
% robustness.
Our approach can be made robust to imprecise localization. This matters because many mapless, camera-only driving stacks have rough positions from satellite navigation, so the rendered map will not line up exactly with the camera. When trained with pose-augmented renders, the model tolerates position errors up to 3~m and still outperforms the base model without conditioning. This makes it possible to reuse a map from one drive for later drives in the same area, even when they are only roughly localized.

We summarize our contributions as follows:
\vspace{-0.20cm}

\begin{itemize}[leftmargin=1.5em] 
     
\item We introduce the \emph{geometric anchoring} approach GeoSCope, which grounds any autoregressive driving world model in a SLAM reconstruction built from a single monocular camera log. Rendered at the pose of each upcoming frame, the map supports both faithful route replay and counterfactual trajectories, while the generative prior remains responsible for all dynamics.

\item We show that these sparse and photometrically degraded renderings are sufficient for grounding and stable long-horizon prediction. The gain persists across conditioning mechanisms, at reduced coverage, and, with pose-perturbed training, under localization errors of several meters. Across three world models and four datasets, our approach roughly doubles long-horizon keypoint matches without suppressing dynamic content, transfers zero-shot to unseen datasets, and sets a new state of the art on the Ann-Arbor-City-Bench cross-traversal benchmark despite never being trained on multi-traversal data.

\end{itemize}
\vspace{-0.25cm}
\section{Related Work} % TODO: massively shorten
\label{background}

\textbf{Driving World Models.}
Early work established real-world generative driving simulation and action-conditioned autoregressive generation~\citep{Kim2021DriveGANTA,hu2023gaia1generativeworldmodel}. 
Subsequent methods improved spatial fidelity through HD maps, 3D bounding boxes, and multi-view geometric constraints~\citep{wang2023drivedreamer,lu2023wovogen,li2023drivediffusion,gao2023magicdrive}, which ground the road layout explicitly but require annotations that most raw driving logs lack.
More recent models obtain realistic, controllable rollouts through large-scale video pre-training and diffusion-based generation~\citep{yang2024genad,gao2024vista,epona2025,nvidia2025cosmos}.
Orbis v1~\citep{mousakhan2025orbis} and its hierarchical successor Orbis v2~\citep{orbis2_2026} showed that compact models trained only on monocular driving video can produce strong long-horizon rollouts.
In these models, the lack of grounding leads to structural drift in rollouts longer than a few seconds, which we address by anchoring generation to a geometric reconstruction of the traversed route.

%Despite strong initial visual fidelity, the lack of grounding leads to structural drift during rollouts longer than a few seconds across all the mentioned models.
%
%Our work targets this long-horizon drift by anchoring the generation to a geometric reconstruction of the specific route being traversed.

% \textbf{Reconstruction-based simulators.} A parallel line of work builds an explicit reconstruction of a recorded drive and renders it at new camera poses, from neural sensor simulators~\citep{yang2023unisim,tonderski2024neurad} to Gaussian-splatting methods that decompose urban scenes into a static background and per-object dynamics~\citep{yan2024street,chen2025omnire,fischer2024dynamic}. These provide exact geometry and direct camera control, but the reconstruction must also produce the output pixels, so unobserved viewpoints yield holes and artifacts.  
% StreetCrafter~\citep{yan2025streetcrafter} narrows the gap to generative models by conditioning a video diffusion model on LiDAR point-cloud renders, which also enables pixel-level scene edits. It remains a conditional view synthesizer rather than a world model, however, since a render is supplied for every generated frame and the recorded dynamics are replayed rather than predicted.

\textbf{Reconstruction-based simulators.} A parallel line of work reconstructs a recorded drive explicitly and renders it at new camera poses, from neural sensor simulators~\citep{yang2023unisim,tonderski2024neurad} to Gaussian-splatting methods that separate a static background from per-object dynamics~\citep{yan2024street,chen2025omnire,fischer2024dynamic}. These give exact geometry and direct camera control, but the reconstruction must produce the output pixels, so unobserved viewpoints yield holes and artifacts. StreetCrafter~\citep{yan2024streetcrafter} relaxes this by conditioning a video diffusion model on LiDAR renders, but it requires LiDAR and synthesizes views of a recorded clip rather than predicting a future.

\textbf{Persistent Scene Context.}
%Long-horizon video models either reduce the training--inference gap with or without external memory. 
Prior work on long-horizon video generation can be grouped by whether it maintains an explicit scene memory and by the source of that memory.
%
%An extended comparison on scene memory approaches is provided in the appendix (\Cref{tab:related_memory}).

% \textit{Implicit and distributional.} 
% %
% Methods such as Diffusion Forcing~\citep{chen2025diffusion}, Self-Forcing++~\citep{cui2025selfforcing}, History-Guided Video Diffusion~\citep{song2025histgdf}, and MilliVid~\citep{chandratreya2026millivid} improve robustness by reducing the training--inference gap without external memory. 
% %
% They, for example us per-token independent noise levels, train on the models own degraded rollouts or apply a coarse-to-fine temporal hierarchy. 
% %
% Their conditioning remains bounded by the learned prior, however, and cannot independently correct hallucinated scene content as the sequence length grows.

\textit{No explicit memory.} Some methods reduce the training--inference gap without external memory, for example by using per-token independent noise levels or by training on the model's own degraded rollouts~\citep{chen2025diffusion, cui2025selfforcing}. Their conditioning remains bounded by the learned prior and cannot correct hallucinated scene content as the sequence grows.
%Their conditioning remains bounded by the learned prior, however, and cannot correct hallucinated scene content as the sequence grows.

%\textit{Generated content memory.} 
%
%A second class maintains explicit memory built from model outputs. 
%Methods use a compressed global frame summary~\citep{henschel2024streamingt2v}, WorldMem~\citep{xiaoworldmem} retrieves previously generated frames by field-of-view overlap and SlowFastGen~\citep{hong2024slowfastgen} encodes scene-specific memory via per-scene LoRA fine-tuning. 
%
%WorldWarp~\citep{worldwarp2025}, VMem~\citep{Li_2025_vmem}, MosaicMem~\citep{mosaicmem2026}, and SpMem~\citep{wu2025videoworldmodels} lift generated frames into 3D Gaussian splats, surfel maps, 3D memory patches, and static point maps, respectively, rendering them to novel poses to condition the generation process.
%
%Because the geometric representations are derived from generated content, generation errors accumulate within the reconstructed scene, degrading future conditioning.

\textit{Memory from generated content.} Others build an explicit memory from the model's own outputs, storing compressed frame summaries, retrieved past frames, or scene-specific weights~\citep{henschel2024streamingt2v,xiaoworldmem,hong2024slowfastgen}. Others lift generated frames into 3D representations such as Gaussian splats, surfel maps or point maps, and render them at new poses to condition generation~\citep{worldwarp2025,Li_2025_vmem,mosaicmem2026,wu2025videoworldmodels}. 
Since this memory is built from generated content, errors accumulate within it and degrade future conditioning.

\textit{Memory from real observations.}
%
% The closest approaches to ours ground generation in a scene representation build from real observations.
% %
% Render-based methods reconstruct a 3D representation offline and render it at the target camera pose.
The closest approaches to ours ground generation in a representation built from real observations, reconstructed offline and rendered at the target camera pose.
GEN3C~\citep{ren2025gen3c} unprojects monocular depth from seed images into a point-cloud cache, and GeoDrive~\citep{chen2025geodrive} extracts a static point cloud from a single seed frame, so their geometry covers only the initially visible scene and cannot anchor generation once the rollout leaves that view frustum.
StreetNVS~\citep{streetnvs2025} instead fuses LiDAR reprojections with surround-view imagery, Vista4D~\citep{vista4d2025} lifts training frames into a persistent 4D point cloud, and VerseCrafter~\citep{zheng2026versecrafter} pairs a static background point cloud with per-object 3D Gaussian trajectories to control both camera and object motion.
These approaches primarily perform geometry-conditioned novel-view synthesis that replays observed dynamics.
% 
%Very recent retrieval-based systems such as CityRAG~\citep{chou2026cityrag} and SWM~\citep{seo2026grounding} query geo-registered imagery or pre-built geospatial maps, but require dense multi-pass coverage.
Retrieval-based systems such as CityRAG~\citep{chou2026cityrag} and SWM~\citep{seo2026grounding} query geo-registered imagery or pre-built geospatial maps, but require dense multi-pass coverage.
We use a single monocular SLAM pass to anchor only the static route context, leaving dynamics to the world-model prior, and render it prospectively at future poses so that the same map also supports counterfactual trajectories.
%
% We instead use a single monocular SLAM pass only to anchor static route context, leaving future dynamics to the world-model prior and making the approach scalable to ordinary single-traversal driving logs.
% %
% Prospective rendering at future poses further supports counterfactual trajectories absent from the traversal.
%
%Their coverage is therefore bounded by what the seed views contain or by the sensors needed to build it, and they primarily perform geometry-conditioned novel-view synthesis that replays observed dynamics. Retrieval-based systems~\citep{chou2026cityrag,seo2026grounding} instead query geo-registered imagery or pre-built geospatial maps, but require dense multi-pass coverage. We use a single monocular SLAM pass to anchor only the static route context, leaving dynamics to the world-model prior, and render it prospectively at future poses so that the same map also supports counterfactual trajectories.
\section{Method}
\label{sec:method}

\subsection{Problem Setup and Prospective Geometric Anchoring}
\label{sec:setup}

We consider an autoregressive driving world model that generates future frames one window at a time. Given the $N_{\text{ctx}}$ most recent frames $G_{n-N_{\text{ctx}}:n-1}$, the model predicts the next $F_{\text{pred}}$ frames and appends them to the context for the next rollout step. Since the generation is not constrained by the real environment, the model can produce any plausible future. Each window is predicted from previous predictions, so deviations accumulate and the rollout drifts away from the recorded scene.

\begin{figure}[t]
    \centering
    \includegraphics[width=1\linewidth]{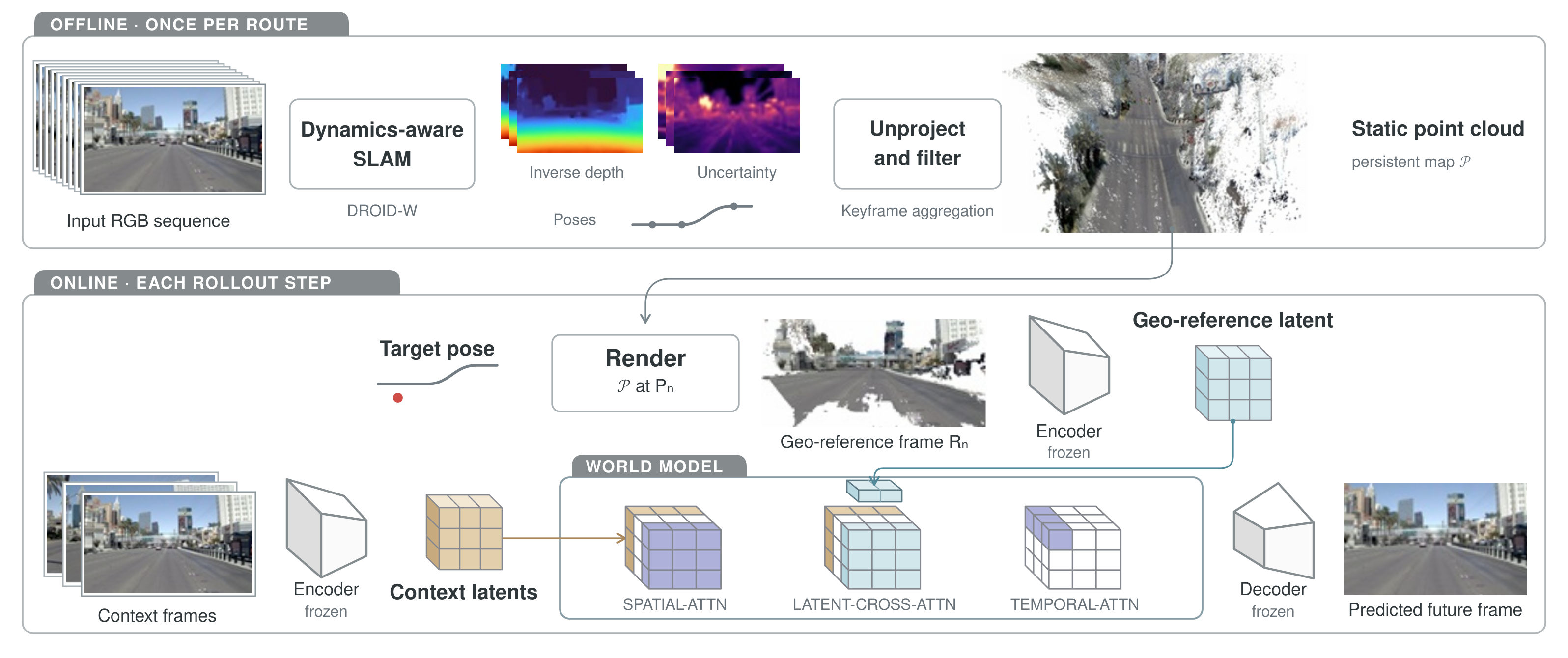}
    \vspace{-0.5cm}
    \caption{\textbf{GeoSCope overview}. Top: SLAM reconstruction and point cloud filtering (\Cref{sec:reconstruction}) are performed \emph{offline}, once per route. Bottom: the resulting point cloud is reused \emph{online} at every rollout step to render geo-reference frames (\Cref{sec:rendering}) that condition the world model (\Cref{sec:conditioning}).}
    \label{fig:method_overview}
\end{figure}

To ground the rollout, we introduce a conditioning signal that is independent of the model's predictions (\Cref{fig:method_overview}). From a single monocular recording of the route, we reconstruct a static point cloud $\mathcal{P}$ once, offline (\Cref{sec:reconstruction}). For each frame to be generated, we render $\mathcal{P}$ at the corresponding camera pose $P_n$ (\Cref{sec:rendering}), which yields a pixel-aligned geo-reference frame $R_n = \Pi(\mathcal{P}, P_n)$, where $\Pi$ denotes the rendering function. The world model is then conditioned on this frame,
\begin{equation}
    G_n \sim p_\theta\bigl(\,\cdot \mid G_{n-N_{\text{ctx}}:n-1},\, R_n\bigr).
    \label{eq:conditional}
\end{equation}
We refer to this as \emph{prospective geometric anchoring}, since the point cloud is rendered at the pose of the upcoming frame before it is generated. During training, $P_n$ is the SLAM-estimated pose of the recorded frame, so that $R_n$ is aligned with the target frame. At inference, the choice of poses determines the rollout. The recorded poses enable a replay in the original drive's environment, and modified poses produce counterfactual rollouts in the same scene.

\subsection{Static Scene Reconstruction}
\label{sec:reconstruction}

\textbf{Dynamics-aware SLAM.}
We estimate the scene geometry with DROID-W~\citep{Li2026DROIDW}, an off-the-shelf dynamics-aware SLAM system. It extends the optical-flow bundle adjustment of DROID-SLAM~\citep{teed2021droidSalm} with a monocular metric-depth prior and a learned per-pixel uncertainty, which makes the reconstruction robust to moving objects and to single-traversal monocular input. Unlike the usual online deployment of SLAM, we run DROID-W offline over the full recording, so that all frames of the route contribute to the optimization. For each selected keyframe, it returns a camera pose, an inverse-depth map and a per-pixel uncertainty map. We refer to the original work for details on tracking and optimization, and list our configuration in \Cref{tab:hp_droidw}.

\textbf{Point cloud construction and filtering.}
The keyframe poses, inverse-depth maps, and input RGB images are unprojected to a colored point cloud in world coordinates. Filtering is performed once, on the complete point cloud accumulated over the full traversal, rather than per rendered frame. Five sequential consistency filters (disparity, uncertainty, multi-view geometric and photometric, voxel-outlier) clean the raw cloud and suppress unreliable geometry arising from dynamic objects, occlusions, and mismatched correspondences. The result is a static point cloud $\mathcal{P}$ that is computed once per route. Filter definitions and thresholds are given in \Cref{tab:hp_rendering}.

\subsection{Rendering Geo-Reference Frames}
\label{sec:rendering}

While $\mathcal{P}$ is precomputed offline, rendering is performed online at each rollout step. Each frame is rendered from the complete point cloud rather than only from the points observed up to that frame, since the route has already been traversed and the reconstruction is available before the rollout begins. Points projecting to the same pixel are composited with a depth-gated weighted blend rather than a hard z-buffer, which fills small gaps and suppresses flicker on the sparse cloud. Details are given in \Cref{app:point_cloud_rendering} of the Appendix.

\subsection{Conditioning Mechanisms}
\label{sec:conditioning}

The rendered geo-reference frames have an error profile that is complementary to the generative prior. They are geometrically grounded, stable, and controllable, but sparse, incomplete, and photometrically degraded. See~\Cref{app:geo_ref_quality} for examples and quality assessment. The design question is thus how strongly the model should trust them.

\textbf{Geo-reference encoding.}
Each geo-reference frame is encoded with the frozen tokenizer of the base model. We denote by $Z^{\text{geo}} \in \mathbb{R}^{F_{\text{pred}} \times H_{\text{lat}} \times W_{\text{lat}} \times C}$ the latents of the geo-reference frames of one rollout window, and by $M$ the binary mask of latent positions covered by projected points. Further, $g_{f,i} \in \mathbb{R}^{C}$ denotes the geometry token of frame $f$ at spatial position $i$. % \in \{1, \dots, N\}$, with $N = H_{\text{lat}} W_{\text{lat}}$.

%\textbf{Geo-reference encoding.}
%Each geo-reference frame is encoded with the frozen tokenizer of the base model. For one rollout window, $Z^{\text{geo}} \in \mathbb{R}^{F_{\text{pred}} \times H_{\text{lat}} \times W_{\text{lat}} \times C}$ denotes the resulting latents, $M$ the binary mask of latent positions covered by projected points, and $g_{f,i} \in \mathbb{R}^{C}$ the geometry token of frame $f$ at spatial position $i$.

\textbf{Injection mechanisms.}
We compare mechanisms that range from hard to soft trust in the signal (\Cref{tab:conditioning_overview}). \emph{Latent blending} is training-free and interpolates the latent toward a noised version of $Z^{\text{geo}}$ in covered regions during the early, noise-dominated denoising steps. \emph{Concatenation} fuses $Z^{\text{geo}}$ with the noisy input latent along the channel axis, using a zero-initialized extension of the patch embedding. The \emph{adaLN adapter} adds per-token offsets derived from the geometry to the  global adaLN 
\begin{wraptable}{r}{0.6\textwidth}
\vspace{-0.3cm}
\centering
\caption{Overview of the geometric conditioning mechanisms, ranging from hard latent-space constraints to soft learned modulation. Parameter counts refer to Orbis~v1.}
\label{tab:conditioning_overview}
\small
\setlength{\tabcolsep}{3pt}
\begin{tabular}{l l c c}
\toprule
Variant & Spatial scope & Trust & New params \\
\midrule
Latent blending    & masked tokens only    & hard & $0$        \\
Concatenation      & local (per token)     & soft & $0.02$\,M  \\
AdaLN              & local (per token)     & soft & $43.2$\,M  \\
Cross-attn         & non-local (attention) & soft & $59.1$\,M  \\
Cross-attn + AdaLN & local + non-local     & soft & $102.3$\,M \\
\bottomrule
\end{tabular}
\end{wraptable}
modulation of every block, which gives spatially local control. \emph{Gated cross-attention} lets every token attend non-locally to the geometry and can therefore discount unreliable regions. We also test cross-attention combined with adaLN. See~\Cref{sec:appendix_cond_mechanism} for full formulations. We used cross-attention for all ablation experiments.

% \begin{table}[ht!]
% \centering
% \caption{Overview of the geometric conditioning mechanisms, ranging from hard latent-space constraints to soft learned modulation. Parameter counts refer to Orbis~v1.}
% \label{tab:conditioning_overview}
% \small
% \begin{tabular}{l l c c c c}
% \toprule
% & & & \multicolumn{2}{c}{Params} & \\
% \cmidrule(lr){4-5}
% Variant & Spatial scope & Trust & added & total & Train.-free \\
% \midrule
% Latent blending    & masked tokens only    & hard & $0$       & $512.8$\,M & \cmark \\
% Concatenation      & local (per token)     & soft & $0.02$\,M & $512.9$\,M & \xmark \\
% AdaLN              & local (per token)     & soft & $43.2$\,M & $556.1$\,M & \xmark \\
% Cross-attn         & non-local (attention) & soft & $59.1$\,M & $571.9$\,M & \xmark \\
% Cross-attn + AdaLN & local + non-local     & soft & $102.3$\,M & $615.2$\,M & \xmark \\
% \bottomrule
% \end{tabular}
% \end{table}

\paragraph{Gated cross-attention.}
Each geometry token is lifted to the hidden dimension by a shared linear projection, and the frozen sinusoidal positional embedding $p_i$ of the query tokens is added to establish a shared spatial coordinate system,
\begin{equation}
    h_{f,i} = W_{\text{proj}}\, g_{f,i} + p_i, \qquad W_{\text{proj}} \in \mathbb{R}^{D \times C}.
    \label{eq:geo_proj}
\end{equation}
A gated cross-attention sub-layer is inserted into every block of the backbone, with queries from the denoising tokens and keys and values from $h$. Queries and keys pass through a parameter-free LayerNorm on the per-head dimension, which prevents logit explosion from the heterogeneous magnitudes of the geometry features. The sub-layer is placed after the spatial self-attention and spatial MLP and before temporal attention~\citep{opensora}, so that queries carry local spatial context and temporal attention propagates the geometry-conditioned features across frames,
\begin{equation}
    x \leftarrow x + \tanh(\alpha) \cdot \mathrm{CrossAttn}\bigl(\mathrm{LN}(x), h\bigr).
    \label{eq:gated_cross_attn}
\end{equation}
The scalar gate $\alpha$ is initialized to zero, so the block computes the pretrained function at initialization, and the gate opens during fine-tuning. Attention is block-diagonal across frames, so predicted frame $j$ attends only to geo-reference frame $j$, while context frames do not participate. The final projection layer likewise applies gated cross-attention before the global adaLN.

% \subsection{Training}
% \label{sec:training}
\subsection{Integration into Pre-trained World Models}
\label{sec:base_models}

\textbf{Base world models.} We apply our conditioning to three base world models of different scales and architectures, all initialized from their released weights. Architecture details are given in \Cref{app:wm_details}.
\textbf{Orbis~v1}~\citep{mousakhan2025orbis} is a 469M-parameter STDiT that rolls out frame by frame, as described in \Cref{sec:setup}. We use it for the comparison of all conditioning mechanisms.
\textbf{Orbis~v2}~\citep{orbis2_2026} is a 1.07B-parameter hierarchical model, in which an abstract predictor generates subgoals at 2\,Hz and a detail predictor generates frames conditioned on them. We apply gated cross-attention to the abstract predictor, so that the geometric signal guides the subgoals, which the detail predictor then turns into frames.
\textbf{Cosmos-Predict2.5}~\citep{nvidia2025cosmos} is a 2B-parameter video DiT with a temporally compressing VAE, rolled out chunk by chunk. We apply gated cross-attention between the denoising target latents and the encoded geometric signal latents.

\textbf{Training.} Latent blending is applied directly to the released model and requires no training. For all learned variants, the tokenizer remains frozen, and the full backbone is fine-tuned jointly with the new modules using the flow-matching objective of the respective base model. For a fixed fraction of training samples, $Z^{\text{geo}}$ is set to zero, which enables classifier-free guidance at inference.

\textbf{Training with perturbed poses.}
To make the model tolerant to imprecise localization, we optionally train with geo-reference frames rendered from displaced viewpoints. For each frame, we sample a displacement uniformly within a circle of radius $R{=}3$\,m around the true pose, holding orientation fixed. We additionally add small yaw/pitch jitter ($\pm2^\circ$/$\pm1^\circ$). This teaches the model to trust what the rendering shows without assuming it is exactly aligned. %The model then learns to perform non-pose aligned conditioning.
%treat the signal as a soft hint rather than a hard constraint (\Cref{sec:localization}).
\section{Experiments}
\label{experiments}

\subsection{Experiment Details}
\label{experiment_details}

\textbf{Datasets.} We train on nuPlan~\citep{caesar2021nuplan}, using only the front camera without LiDAR, maps, or object annotations. 
Following Orbis v1 and Orbis v2, we test on their curated \textsc{nuPlan-Turns} subset of 400 clips which includes challenging turning maneuvers.
%on their curated \textsc{nuPlan-Turns} validation subset of 400 clips selected from the nuPlan validation split which includes challenging turning maneuvers.
%where the initial yaw rate is at least \(0.12\,\text{rad/s}\) (${\approx}1$ standard deviation), targeting challenging turning maneuvers. 
%
We also zero-shot test on three more datasets. On Waymo~\citep{sun2020waymoopen}, we use 182 full-length, non-overlapping validation clips, each the single longest contiguous window its segment supports (up to 197 raw frames at 10\,Hz, ${\approx}19.7$\,s). 
We apply the same recipe used to generate nuPlan-Turns to MARS's~\citep{mars} 66 multitraversal recordings, yielding 400 turn-event scenes: initial yaw rate exceeding 1 standard deviation of MARS's measured distribution, with the ego vehicle maintaining a mean speed of at least $1\,\text{m/s}$ over a 20\,s horizon.
On nuScenes~\citep{caesar2020nuscenes}, we roll out all 150 official validation scenes in full, without windowing.

To compare against the concurrent Seoul World Model~\citep{seo2026grounding}, we follow their protocol and construct \textsc{Ann-Arbor-City-Bench} from 30 MARS cross-traversal pairs, where the geo-reference signal is reconstructed from one traversal of a route and rendered at the camera poses of a second, independent traversal.

%Separately, to compare against the concurrent Seoul World Model~\citep{seo2026grounding} paper, we follow their evaluation protocol to construct the \textsc{Ann-Arbor-City-Bench} from 30 MARS~\citep{mars} cross-traversal pairs: the geo-reference signal is reconstructed from one (sparse) traversal of a route and rendered at a second, independent traversal's real camera poses, so the model must generalize across passes of the same route rather than merely extrapolate within a single one.

\textbf{Training details.} For each training scene, we run DROID-W and the rendering pipeline on a contiguous 300-frame subsequence and draw all training windows from it. 
%
%Across all runs the tokenizer remains frozen, and the full backbone is trained jointly with the newly added architecture parts.
%
%For a fixed fraction of training steps the geo-reference signal is dropped to later enable classifier-free guidance (cfg).
%
%All training videos are sampled at 5\,FPS or 10\,FPS depending on the base model.
%
All models are trained on 16 NVIDIA GH200 GPUs (4 nodes × 4 GPUs) with an effective batch size of 64.

\textbf{Baselines.} We compare against the released state-of-the-art driving world models Orbis v1, Orbis v2, and Cosmos2.5-predict-2B.
For Cosmos, we use the 480p variant for both the baseline and the anchored model. % due to limited compute budget.
We additionally compare against the concurrent retrieval-based method Seoul World Model~\citep{seo2026grounding}. Since no code or checkpoints are released, we follow and reproduce their evaluation protocol. We retain the other models from their table for comparison.

\subsection{Evaluation}
\label{sec:evaluation}

% Part of our evaluation replays a \emph{specific} logged route rather than hallucinating a plausible one, we can evaluate the generated video against the true observed future frame-by-frame.
% %
% We organize the evaluation along three complementary axes, each targeting a distinct claim of the paper, and report one representative metric for each failure mode.

% \textbf{Horizon axis.}
% Autoregressive drift is a temporal phenomenon: methods that look identical in the first few frames diverge sharply over long rollouts. All rollouts are assessed at 5\,FPS. Every metric is reported over a short horizon (0 to 10 seconds) and a long horizon (10 to 20 seconds). We deviate from this only in the table where we average across the four datasets where the short horizon is 0 to 8 seconds and the long one from 8 to 16 seconds. 

Our method replays logged routes, so generated videos can be compared frame by frame with the recorded future. We organize the evaluation along three complementary axes, each targeting a distinct claim of the paper. Every metric is reported over a short horizon (0--10\,s) and a long horizon (10--20\,s), as autoregressive drift only becomes visible in long rollouts. For the average across four datasets (\Cref{tab:long_horizon}), we use 0--8\,s and 8--16\,s. All rollouts are assessed at 5\,FPS.

\textbf{1.\ Consistency (reference-based).}
These metrics compare each generated frame with its temporally aligned ground-truth frame, quantifying how faithfully the model reproduces the observed future and, combined with the horizon axis, how quickly it drifts.
%
%We report three complementary metrics. \textbf{LPIPS}~\citep{lpips} for perceptual similarity, which degrades gracefully under the small viewpoint and timing offsets that pixel-aligned measures over-penalize. The number of \textbf{LightGlue keypoint matches}~\citep{lightglue} with SuperPoint keypoints, confidence ${>}0.5$ between generated and ground-truth frames, a structure-aware measure that requires local scene layout to be preserved at approximately corresponding locations. 
%Finally, \textbf{depth AbsRel} error between monocular depth predictions on both frames after per-frame median-scale alignment, isolating geometric from photometric consistency. For the latter, we use DepthAnything~V3~\citep{depthanything3}, deliberately avoiding a shared backbone with the Metric3D~\citep{metric3d} prior inside DROID-W, so that the conditioning signal and its evaluator remain independent.
%
We report three complementary metrics. \textbf{LPIPS}~\citep{lpips} measures perceptual similarity and degrades gracefully under the small viewpoint and timing offsets that pixel-aligned measures over-penalize. 
The number of \textbf{LightGlue keypoint matches}~\citep{lightglue} between generated and ground-truth frames is structure-aware and requires the local scene layout to be preserved at approximately corresponding locations. \textbf{Depth AbsRel} error between monocular depth predictions on both frames isolates geometric from photometric consistency. For the latter we use DepthAnything~v3~\citep{depthanything3}, avoiding a shared backbone with the Metric3D~\citep{metric3d} prior inside DROID-W so that the conditioning signal and its evaluator remain independent.

% \textbf{2.\ Perceptual quality (correspondence-free).}
% To test whether reduced drift comes at the cost of visual realism, we report \textbf{FVD}~\citep{fvd}, which compares spatio-temporal I3D feature distributions of generated and real clips and is therefore sensitive to temporal artifacts, blur, and collapse even when generated frames no longer align with any reference, and requires no frame-level correspondence. %it applies equally to counterfactual rollouts.

\textbf{2.\ Perceptual quality} is measured with FVD~\citep{fvd}, which requires no frame-level correspondence and tests whether reduced drift comes at the cost of visual realism. 

\textbf{3.\ Trajectory adherence (controllability).}
On the logged route, high consistency already implies trajectory adherence; this axis isolates it. 
%
%
%and, crucially, extends to \emph{counterfactual} rollouts (lane offsets, altered speed) where no ground-truth video exists but the commanded trajectory is known.
%
We recover the camera path of the generated rollout with VGGT~\citep{vggt}, align it to the commanded ego-trajectory with a Sim(3) Umeyama fit to resolve the gauge ambiguity of monocular reconstruction, and report the \textbf{ADE} (average displacement error) between the two paths.

% ---------------------------------------------------------------------------------------------------
% ---------------------------------------------------------------------------------------------------

\section{Results}
\label{results}

\subsection{GeoSCope anchoring reduces structural drift}

\begin{figure}
    \centering
    \includegraphics[width=1\linewidth]{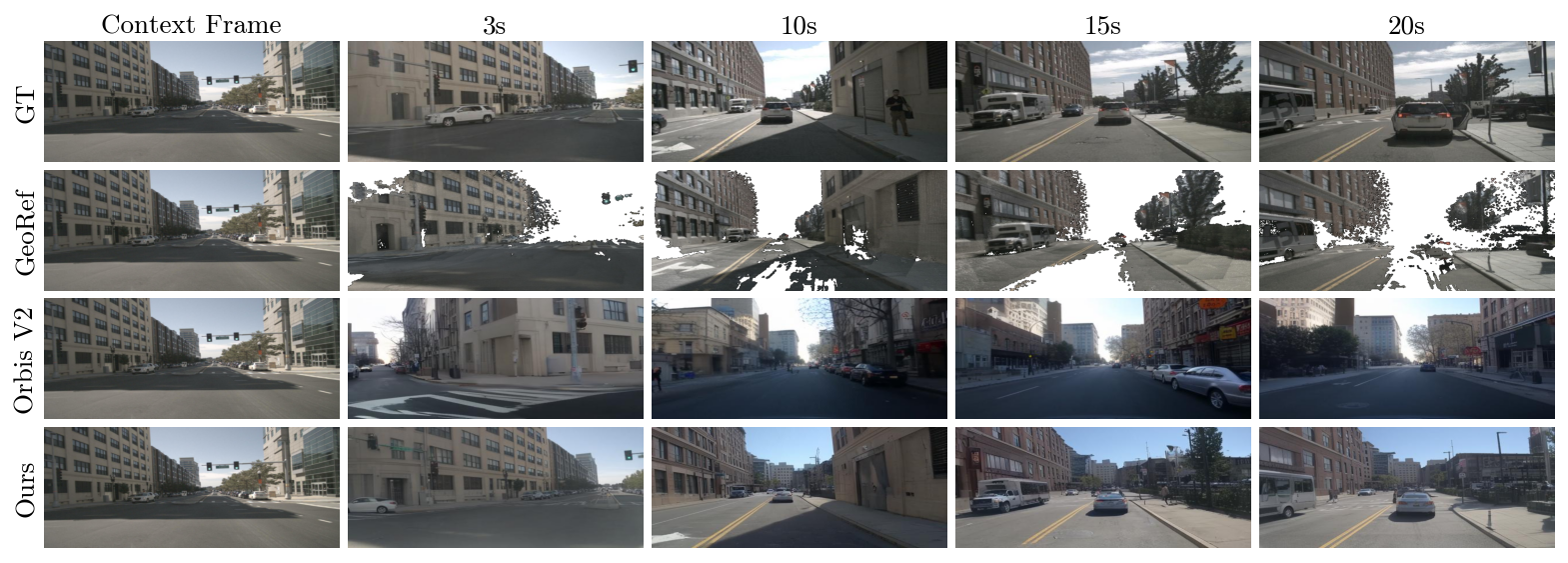}
    \caption{Qualitative long rollouts at different prediction horizons. GeoSCope follows the GeoRef geometry where it is available and fills in the gaps, while the baseline slowly drifts away from the recorded layout. Moving objects are still generated by the world model.}
    \label{fig:long_horizon}
\end{figure}

We train gated cross-attention as the ingestion mechanism across all three base models and compare each released model with its anchored variant in \Cref{tab:long_horizon}. Three of the four datasets are evaluated zero-shot, since training uses nuPlan only.
%
%The geo-reference signal alone exhibits the expected complementary error profile: it is sparse, degraded renders yield poor distributional realism (FVD ${>}1700$), yet its keypoint matches do not degrade from the short to the long horizon, confirming that the signal is geometrically stable precisely where the generative baselines drift.
%
The two information sources have complementary error profiles. The geo-reference signal alone has poor realism (FVD ${>}1700$), yet its keypoint matches do not degrade from the short to the long horizon, so it remains geometrically stable where the generative prior drifts. 
The base models fail in the opposite way, starting from strong visual realism and losing structure as the rollout proceeds, with long-horizon FVD rising by 58 to 99\% and keypoint matches dropping by more than half.
%
%All base model variants degrade sharply over the long horizon, with FVD increasing by around 60\% and keypoint matches dropping by more than half.
%
% Our model combines the strengths of both sources: it surpasses the baselines' visual realism early on and retains it over the long horizon, while matching or exceeding the geo-reference signal's temporal stability in KP matches and flattening the FVD slope significantly.
Our anchoring approach combines the strengths of both sources and improves all three evaluation axes at once. It preserves structural consistency, as measured by keypoint matches and depth error, keeps perceptual quality high by flattening the FVD slope substantially, and improves trajectory adherence.

\vspace{-0.15cm}
\begin{table}[ht!]
\centering
\caption{%
  Average across \textsc{nuPlan-Turns}, Waymo, \textsc{MARS} and nuScenes for short/long-horizon metrics; the full table is in the appendix. Each base world model is shown directly above its anchored variant (shaded). ADE in $\times10^{-3}$\,m. Best per column in \textbf{bold}.%
}
\label{tab:long_horizon}
\footnotesize
\setlength{\tabcolsep}{3pt}
\renewcommand{\arraystretch}{1.05}
\begin{tabular}{l|cc|ccc|cc|cc|cc}
\toprule
& \multicolumn{2}{c|}{LPIPS $\downarrow$} & \multicolumn{3}{c|}{FVD $\downarrow$} & \multicolumn{2}{c|}{KP Matches $\uparrow$} & \multicolumn{2}{c|}{Depth AbsRel $\downarrow$} & \multicolumn{2}{c}{ADE $\downarrow$} \\
Method & Short & Long & Short & Long & Slope & Short & Long & Short & Long & Short & Long \\
\midrule
GeoRef signal only & 0.506 & 0.493 & 2044.9 & 1755.3 & -- & 188.6 & 200.1 & 0.341 & 0.342 & 3.9 & 7.1 \\
\midrule
Orbis v1 & 0.417 & 0.508 & 323.9 & 539.4 & 64.0 & 217.3 & 104.8 & 0.214 & 0.307 & 4.8 & 14.7 \\
\quad +  fine-tuned & 0.431 & 0.518 & 342.1 & 573.8 & 59.6 & 199.5 & 88.5 & 0.225 & 0.318 & 3.6 & 10.0 \\
\rowcolor{oursrow}
\quad + GeoSCope & 0.345 & 0.411 & \textbf{227.1} & \textbf{325.0} & \textbf{27.2} & 267.8 & 184.5 & 0.157 & 0.217 & \textbf{2.3} & \textbf{6.5} \\
\cmidrule{1-12}
Orbis v2 & 0.393 & 0.494 & 282.9 & 446.8 & 45.9 & 218.3 & 98.4 & 0.202 & 0.300 & 4.0 & 11.4 \\
\rowcolor{oursrow}
\quad + GeoSCope & \textbf{0.313} & \textbf{0.379} & 229.8 & 337.0 & 30.4 & 296.1 & 236.0 & \textbf{0.126} & \textbf{0.172} & 2.6 & 6.7 \\
\cmidrule{1-12}
  Cosmos 2.5 & 0.433 & 0.514 & 336.6 & 503.9 & 43.5 & 300.7 & 136.4 & 0.215 & 0.304 & 5.7 & 10.7 \\
\rowcolor{oursrow}
\quad + GeoSCope & 0.341 & 0.412 & 285.0 & 455.1 & 42.3 & \textbf{381.0} & \textbf{278.7} & 0.143 & 0.190 & 4.2 & 8.2 \\
\bottomrule
\end{tabular}
\end{table}

\vspace{-0.2cm}
\subsection{Comparison with grounded world models}

%To compare our method against Seoul World Model (SWM) on their multi-traversal benchmark, a task our method was not explicitly trained for. We show in \Cref{tab:multi-traversal}, that our method despite never being trained on multi-traversal data performs better than all competing methods. 

We compare against the Seoul World Model (SWM) on their multi-traversal benchmark, where the geo-reference is reconstructed from one traversal of a route and rendered at the camera poses of a second independent traversal. Our model is never trained on multi-traversal data, so the setting is entirely zero-shot for it, unlike for the retrieval-based SWM.
As shown in \Cref{tab:multi-traversal}, our approach outperforms all competing methods on FID, FVD, mPSNR and mLPIPS, with SWM (TF) leading narrowly on Img.\,Q.
This is consistent with our single-traversal findings. The static structure of a route is the same on every pass, so it gives the model a stable reference to exploit, and it does not matter whether that reference was reconstructed from the same traversal or a different one.

%\vspace{-0.25cm}
\begin{table}[ht!]
    \centering
    \caption{\textbf{Comparison with other grounded world models.} We evaluate visual and temporal fidelity, and conditioning signal adherence. Values are reported on Ann-Arbor-City-Bench.}
    \resizebox{0.9\textwidth}{!}{% 
    \begin{tabular}{l ccccc}
    \toprule
    Method & FID$\downarrow$ & FVD$\downarrow$ & Img.Q.$\uparrow$ & mPSNR$\uparrow$ & mLPIPS$\downarrow$ \\
    \midrule
    %Aether~\cite{zhu2025aether} & 132.77 & 1214.84 & 0.51 & 13.03 & 0.635 \\
    %DeepVerse~\cite{chen2025deepverse} & 182.95 & 1524.97 & 0.46 & 13.43 & 0.727 \\
    Yume1.5~\citep{mao2025yume} & 85.62 & 993.62 & 0.61 & 14.15 & 0.623 \\
    HY-World1.5~\citep{hyworld2025} & 67.02 & 864.76 & 0.54 & 14.26 & 0.575 \\
    FantasyWorld~\citep{dai2025fantasyworld} & 67.72 & 917.57 & 0.49 & 11.97 & 0.592 \\
    Lingbot~\citep{team2026advancing} & 57.99 & 1039.50 & 0.60 & 12.51 & 0.641 \\
    SWM (TF)~\citep{seo2026grounding} & 56.61 & 640.17 & \textbf{0.66} & 15.18 & 0.481 \\
    SWM (SF)~\citep{seo2026grounding} & 43.97 & 779.94 & 0.57 & 14.20 & 0.573 \\
    \midrule 
    GeoRef Signal & 201.06 & 2298.45 & 0.55 & 8.10 & 0.516 \\
    Ours (Orbis v1 + GeoSCope) & \textbf{39.27} & \underline{486.69} & 0.60 & \textbf{18.61} & \textbf{0.313} \\
    Ours (Orbis v2 + GeoSCope) & \underline{39.34} & \textbf{443.16} & \underline{0.65} & \underline{17.03} & \underline{0.326} \\
    %Ours (Cosmos 2.5 + GeoSCope) & 205.54 & 2443.08 & 0.36 & \underline{18.16} & 0.416 \\
    \bottomrule
    \end{tabular}
    }
    \label{tab:multi-traversal}
\end{table}

% Counterfactual trajectories — generated by make_results.py
\subsection{Controllable counterfactual generation}

% \begin{wraptable}{r}{0.4\textwidth}
%     \vspace{-1.05cm}
%     \centering
%     \caption{Performance metrics under counterfactual trajectory perturbations.
%     }
%     \label{tab:counterfactual}
%     \small
%     \setlength{\tabcolsep}{4pt}
%     \renewcommand{\arraystretch}{1.05}
%     \begin{tabular}{l|c|c|c}
%     \toprule
%     Perturbation & FID $\downarrow$ & FVD $\downarrow$ & ADE $\downarrow$ \\
%     \midrule
%       Lateral $+1$\,m & 26.0 & 224.5 & 10.8 \\
%       Lateral $+0.5$\,m & 24.5 & 202.7 & 10.3 \\
%       Nominal & 23.5 & 188.4 & 9.9 \\
%       Lateral $-0.5$\,m & 24.5 & 202.8 & 10.2 \\
%       Lateral $-1$\,m & 26.1 & 224.0 & 11.3 \\
%       Wiggle $0.5$\,m & 24.0 & 250.4 & 11.8 \\
%       Speed $0.85\times$ & 27.5 & 208.7 & 11.9 \\
%       %Nominal & 23.5 & 188.4 & 9.9 \\
%       %Lateral $|1|$\,m & 27.0 & 234.3 & 11.1 \\
%       %Lateral $|2|$\,m & 30.0 & 272.6 & 11.7 \\
%       %Speed $0.85\times$ & 29.0 & 296.1 & 17.8 \\
%     \bottomrule
%     \end{tabular}
% \end{wraptable}

The reconstruction can be rendered along any trajectory, so the same model that replays a logged route can also execute trajectories never observed during training. We command three perturbation types: lateral offsets, a sinusoidal wiggle, and speed changes; see~\Cref{tab:counterfactual}. The model follows the 
\begin{wraptable}{r}{0.50\textwidth}
    \vspace{-0.35cm}
    \centering
    \caption{Counterfactual trajectory perturbations. Compliance measures the followed response to the command. Details in~\Cref{tab:counterfactual-compliance}.}%, in per-row units: metres of lateral, slope for wiggle and fraction of samples closer to the target speed.}
    \label{tab:counterfactual}
    \small
    \setlength{\tabcolsep}{4pt}
    \renewcommand{\arraystretch}{1.05}
    \begin{tabular}{l|c|c|c}
    \toprule
    Perturbation & FVD & ADE & Compliance \\
    \midrule
      Lateral +1\,m    & 224.5 & 10.8 & +0.66\,m \\
      Lateral +0.5\,m  & 202.7 & 10.3 & +0.47\,m \\
      Nominal            & 188.4 &  9.9 & --- \\
      Lateral -0.5\,m  & 202.8 & 10.2 & -0.42\,m \\
      Lateral -1\,m    & 224.0 & 11.3 & -0.75\,m \\
      Wiggle 0.5\,m    & 250.4 & 11.8 & 0.86 slope \\
      Speed 0.85 & 208.7 & 11.9 & $60\%$ \\
    \bottomrule
    \end{tabular}
\end{wraptable}
commanded path closely. Under lateral perturbations, trajectory adherence and realism degrade only mildly and in proportion to the perturbation magnitude (see~\Cref{tab:counterfactual}). Compliance measures use different units: metres of lateral shift, slope of the commanded sine, and fraction of samples closer to the target speed. Full statistics are included in~\Cref{tab:counterfactual-compliance}. These offsets only change the viewpoint, which the reconstruction supports directly. Speed and wiggle changes are the harder cases, as they push the rollout out of the ego motion distribution the base model was trained on.

%Lateral offsets and yaw variations degrade all metrics gracefully and proportionally to perturbation magnitude, confirming smooth generalization to nearby trajectories.

\subsection{Dynamics are not suppressed}\label{sec:dynamics}

% \begin{wrapfigure}{r}{0.4\textwidth}
%     \vspace{-1.5em}
%           \centering
%           \includegraphics[width=\linewidth]{figures/5.7_dynamics_count.png}
%           \caption{.}
% \end{wrapfigure}

% We investigate whether the static geometric reference signal suppresses or reduces the occurrence of dynamic objects such as vehicles, buses, bicycles, or pedestrians. We find that this is not the case. The number of dynamic objects does not decrease compared to the baselines. More details in \Cref{sec:dynamic_vehicles}.

\begin{figure}[H]
    \centering

    % ---------- Left figure ----------
    \begin{minipage}[t]{0.39\textwidth}
        \centering
        \includegraphics[width=\linewidth]{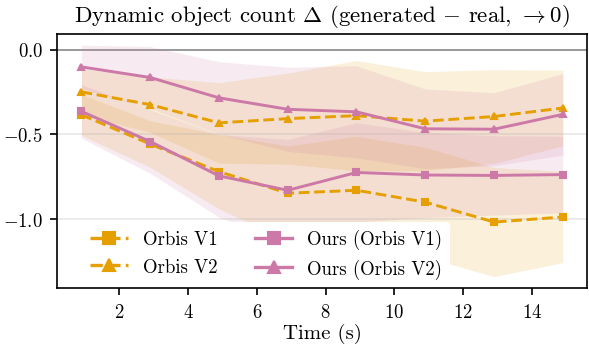}
        \caption{Dynamic object count over the rollout, measured as the difference between the generated and the recorded frames, where zero is ideal.}
        \label{fig:dynamics_count}
    \end{minipage}
    \hfill
    % ---------- Right figure ----------
    \begin{minipage}[t]{0.58\textwidth}
        \centering
        \includegraphics[width=\linewidth]{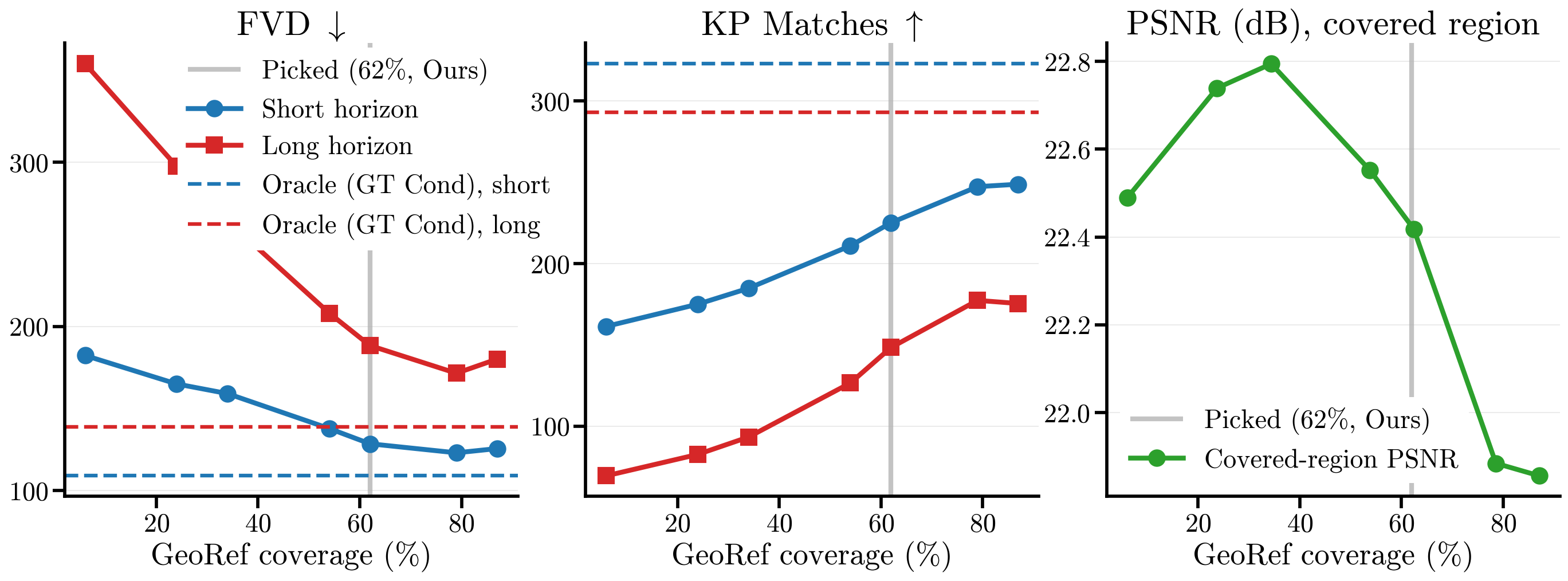}
        \caption{Effect of geo-reference coverage on generation quality. Left and center show rollout quality at the short and long horizon; right shows PSNR over the covered region. GeoSCope operates at 62\%. }%Effect of geometric signal coverage on generation quality and of coverage on signal quality.}
        \label{fig:coverage}
    \end{minipage}
\end{figure}

We investigate whether the static geometric reference signal suppresses the occurrence of dynamic objects such as vehicles, buses, bicycles, or pedestrians. 
We find that throughout the rollout their number stays within the range of the unconditioned base models rather than decreasing, so the static reference constrains the scene while also generating dynamic agents moving through it. The dynamic object count is compared with the base models in~\Cref{fig:dynamics_count}. More details in~\cref{sec:dynamic_vehicles}.

\subsection{Robustness to Localization Error}\label{sec:localization}

% We test whether the model tolerates imprecise localization at inference — a realistic constraint since many driving stacks rely on GPS positions that can be off by several meters. We compare a model trained on aligned renders against one trained with renders displaced by up to 3,m, evaluating each under both conditions.

% The model trained only on aligned signals degrades sharply under a 3,m displacement: long-horizon ADE nearly triples (9.9,$\to$,28.0) and FVD nearly quadruples (188.4,$\to$,728.4), as the model attempts to follow the displaced geometry rather than discounting it. Training with misaligned renders resolves this: the misaligned-trained model performs similarly regardless of whether the evaluation signal is aligned or not (FVD 285.3 vs.\ 299.3 long; ADE 11.5 vs.\ 12.1 long), indicating that it has learned to treat the signal as a soft hint rather than a hard constraint. The cost relative to the aligned-trained model is moderate — long-horizon FVD rises from 188.4 to 285.3 — while the gain over the unconditioned Orbis~v1 backbone (FVD 501.3 long) is largely preserved. Full results are in \Cref{app:tab_alignment}.

In all preceding experiments the renderings come from the same SLAM run that produced the evaluation poses, so the map and the camera are aligned by construction. This favorable case does not hold when poses come from satellite navigation or otherwise have inaccurate localization. We test whether the anchoring mechanism tolerates a reference signal that is misaligned with the ego pose. We compare a model trained on aligned renders against one trained with pose perturbations up to 3\,m, and evaluate each under both conditions.

% GeoRef alignment — minimal 2x2, FVD and ADE (long) only, side (wrap) table
\begin{wraptable}{r}{0.5\textwidth}
    \vspace{-0.75cm}
    \centering
    \caption{%
      Effect of localization errors when trained with and without localization error.%
    }
    \label{tab:alignment_minimal}
    \small
    \setlength{\tabcolsep}{4pt}
    \renewcommand{\arraystretch}{1.05}
    \resizebox{0.5\textwidth}{!}{%
    \begin{tabular}{l|cc|cc}
    \toprule
     & \multicolumn{4}{c}{\textbf{Evaluation}} \\
     & \multicolumn{2}{c|}{Pose error 0m} & \multicolumn{2}{c}{Pose error 0--3m} \\
    \textbf{Training} & FVD $\downarrow$ & ADE $\downarrow$ & FVD $\downarrow$ & ADE $\downarrow$ \\
    \midrule
    Pose error 0m & 188.4 & 9.9 & 728.4 & 28.0 \\
    Pose error 0--3m & 285.3 & 11.5 & 299.3 & 12.1 \\
    \bottomrule
    \end{tabular}
    }
\end{wraptable}

Trained on aligned renderings alone, the model attempts to follow the displaced geometry rather than discounting it, and degrades sharply once the signal is misaligned. Augmenting training with pose perturbations resolves this, and the perturbation-trained model performs similarly whether the evaluation signal is aligned or not, indicating that it has learned to treat the signal as a soft hint rather than a hard constraint. The cost relative to the aligned-trained model is moderate, while the gain over the unconditioned world model is largely preserved. Full results on Orbis~v1 are in \Cref{app:tab_alignment}.

\subsection{How much should the model trust the signal?}

\paragraph{Comparing conditioning mechanisms.}

%A central question is how best to inject the geo-reference signal into the base world models. \Cref{tab:ingestion} compares the four conditioning mechanisms against the two reference points on long-horizon frames.

\Cref{tab:ingestion} compares the four injection mechanisms on long-horizon frames, against the unconditioned backbone and the training-free variant. % visible in bad FVD performance.
%
%The training-free anchor obtains the best keypoint matching results, but its hard latent blend also interits artifacts from the noisy geo-reference signal, as evidenced by its weaker FVD.
\begin{wraptable}{r}{0.50\textwidth}
    \vspace{-0.45cm}
        \centering
        \caption{Conditioning mechanism comparison. Best per column in \textbf{bold}, second best \underline{underlined}.
        }
        \label{tab:ingestion}
        \footnotesize
        \setlength{\tabcolsep}{4pt}
        \renewcommand{\arraystretch}{1.05}
            \begin{tabular}{l|c|c|c}
            \toprule
            Method & FVD $\downarrow$ & KP Matches $\uparrow$ & ADE $\downarrow$ \\
            \midrule
              Orbis v1 & 501.3 & 61.4 & 16.9 \\
              Latent blending & 433.5 & 153.1 & 15.1 \\
            \midrule
              Concat & 197.1 & \underline{144.5} & 10.2 \\
              AdaLN & \underline{195.5} & 137.7 & \underline{10.0} \\
              CrossAttn & \textbf{188.4} & \textbf{148.4} & \textbf{9.9} \\
              CA + AdaLN & 197.5 & 142.8 & 10.1 \\
            \bottomrule
        \end{tabular}
\end{wraptable}
Latent blending improves keypoint matching substantially without any training, but its hard latent blend also inherits artifacts from the noisy geo-reference signal.
% All methods consistently improve long-horizon stability.
% %
% Cross-attention's ability to discount unreliable geometry and propagate valid structure into uncovered regions again gives it a slight advantage over the other methods. Notably, even ingesting the reference information via channel-concatenation leads to a significant improvement across all metrics. The reason is likely that unfreezing the full network lets the architecture incorporate the new information well, somewhat independently of the specific injection mechanism. 
All learned mechanisms improve long-horizon stability by a wide margin over the unconditioned backbone, and the differences between them are small. 
Cross-attention is best on all three metrics and channel concatenation recovers most of the gain with three orders of magnitude fewer added parameters. 
%Combining adaLN and cross-attention yields no additional gain, and during training the adaLN gates open early while the cross-attention gates remain closed.
Combining adaLN and cross-attention yields no gain, and the adaLN gates open early while the cross-attention gates remain closed.

%Interestingly, combining adaLN and cross-attention yields no additional gain. We find that during training, the adaLN gates open early, causing the cross-attention gates to remain closed.

% GeoRef coverage ablation — generated by make_results.py
\paragraph{Coverage of the geometric reference}

% \begin{wrapfigure}{r}{0.55\textwidth}
%     \vspace{-1.00em}
%           \centering
%           \includegraphics[width=\linewidth]{figures/5.5_coverage_fvd_kp_psnr_largefont.png}
%           \caption{Effect of geometric signal coverage on generation quality and of coverage on signal quality.}
% \end{wrapfigure}

We ablate the fraction of the frame covered by the projected point cloud by sweeping the multi-view depth-consistency threshold. All metrics improve monotonically as coverage increases from 6\% to 79\%, with the largest gains on long-horizon FVD and keypoint matches, precisely where the generative baseline drifts most. 
Beyond 79\%, performance plateaus and slightly reverses: loosening the threshold further admits geometrically inconsistent points that introduce noise rather than useful structure. Our operating point at 62\% lies on the conservative side of this trade-off between coverage and point cloud quality.

%Our operating point at 62\% lies on the conservative portion of this curve and provides a robust trade-off between coverage and point cloud quality.
%
%The Oracle row reveals substantial remaining headroom — long-horizon KP matches nearly double (148 $\to$ 293), and FVD drops a further 50 points — indicating that the current bottleneck is reconstruction density rather than the conditioning mechanism, and motivating future work on denser monocular reconstruction. Full results are in this \Cref{app:tab_ablation-coverage}.
%
Conditioning on ground-truth geometry reveals substantial remaining headroom, nearly doubling long-horizon keypoint matches and lowering FVD further. The bottleneck is therefore the density of the reconstruction rather than the conditioning mechanism, which motivates future work on denser monocular reconstruction. Full results are in \Cref{app:tab_ablation-coverage}.
\section{Discussion} 
\label{discussion}

Our experiments show that anchoring an autoregressive driving world model to a prospectively rendered SLAM reconstruction substantially reduces structural drift while turning a single monocular pass into a controllable neural simulation.

\textbf{Limitations.}
Our conditioning signal is static and upper-bounded by the SLAM system's quality. It anchors scene geometry, while agent dynamics, including interactions and reactions to the ego, come from the base model.
Refining the input reconstructions via a diffusion prior, or substituting higher-fidelity 3DGS or mesh representations, remains open for future work.
Finally, we measure rollout fidelity against logged or counterfactual trajectories, not closed-loop policy performance; we leave this to future work.

\textbf{Acknowledgment.} This work was funded by the German Federal Ministry for Economic Affairs and Energy within the project “NXT GEN AI METHODS" (19A23014R). The project was accomplished under GCS compute grants - nxtaim-1.

\section*{AI use statement}
We used generative AI tools for implementing methods, for surveying and retrieving related work, and for editing the paper to improve readability. They were not used for developing the method or conceptual framework, for designing experiments, or for interpreting results. All AI-assisted code was reviewed line by line by the authors and validated by reproducing the reported experiments end-to-end, all references retrieved with AI assistance were verified against the original publications, and all AI-edited text was checked by the authors against the underlying experimental results. We take responsibility for the final content of this work, including text, claims and artifacts produced with the aid of generative AI.

\section*{Ethics statement}
This work uses only public, anonymized driving datasets (nuPlan, Waymo, MARS, nuScenes) under their research licenses, with no new human-subject data collection. We note the dual-use consideration that realistic synthetic driving video could be misused for deceptive media, within our project our intended use is closed-loop policy evaluation and simulation. More broadly, we contribute to the development and evaluation of world models for autonomous driving, technology that could improve road safety and mobility access, but that also raises concerns around job displacement, liability, and over-reliance on imperfect automation should it be deployed prematurely.

\section*{Reproducibility Statement}
We will publicly release the adjusted SLAM processing pipeline, the point-cloud rendering code, and pretrained weights for all three base models upon acceptance. The SLAM configuration is given in \Cref{tab:hp_droidw} and the point-cloud filtering and rendering settings in \Cref{tab:hp_rendering}. The conditioning mechanisms are specified with equations in \Cref{sec:conditioning} and \Cref{sec:appendix_cond_mechanism}. 
%The training and evaluation setup is described in \Cref{experiment_details, evaluation}, with further details in the appendix.
The training and evaluation setup is described in
\Cref{experiment_details,sec:evaluation}, with hyperparameters in
\Cref{app:hp} and metric definitions in
\Cref{app:metrics}.

\bibliography{iclr2027_conference}

@inproceedings{wang2023drivedreamer,
author = {Wang, Xiaofeng and Zhu, Zheng and Huang, Guan and Chen, Xinze and Zhu, Jiagang and Lu, Jiwen},
title = {DriveDreamer: Towards Real-World-Drive World Models for Autonomous Driving},
year = {2024},
booktitle = {ECCV},
pages = {55–72},
numpages = {18},
}

@misc{hu2023gaia1generativeworldmodel,
      title={GAIA-1: A Generative World Model for Autonomous Driving}, 
      author={Anthony Hu and Lloyd Russell and Hudson Yeo and Zak Murez and George Fedoseev and Alex Kendall and Jamie Shotton and Gianluca Corrado},
      year={2023},
      eprint={2309.17080},
      archivePrefix={arXiv},
      primaryClass={cs.CV},
      url={https://arxiv.org/abs/2309.17080}, 
}

@inproceedings{lu2023wovogen,
  title={WoVoGen: World Volume-aware Diffusion for Controllable Multi-camera Driving Scene Generation},
  author={Lu, Jiachen and Huang, Ze and Zhang, Jiahui and Yang, Zeyu and Zhang, Li},
  booktitle={European Conference on Computer Vision (ECCV)},
  year={2024},
}

@inproceedings{li2023drivediffusion,
  title={DiffusionDrive: Truncated Diffusion Model for End-to-End Autonomous Driving},
  author={Bencheng Liao and Shaoyu Chen and Haoran Yin and Bo Jiang and Cheng Wang and Sixu Yan and Xinbang Zhang and Xiangyu Li and Ying Zhang and Qian Zhang and Xinggang Wang},
  booktitle    = {{IEEE/CVF} Conference on Computer Vision and Pattern Recognition,
                  {CVPR} 2025, Nashville, TN, USA, June 11-15, 2025},
  pages        = {12037-12047},
  publisher    = {Computer Vision Foundation / {IEEE}},
  year         = {2025}
}

@inproceedings{gao2023magicdrive,
  title={MagicDrive: Street View Generation with Diverse 3D Geometry Control},
  author={Gao, Ruiyuan and Chen, Kai and Xie, Enze and Hong, Lanqing and Li, Zhenguo and Yeung, Dit-Yan and Xu, Qiang},
  booktitle={International Conference on Learning Representations (ICLR)},
  year={2024}
}

@inproceedings{gao2024vista,
 title={Vista: A Generalizable Driving World Model with High Fidelity and Versatile Controllability}, 
 author={Shenyuan Gao and Jiazhi Yang and Li Chen and Kashyap Chitta and Yihang Qiu and Andreas Geiger and Jun Zhang and Hongyang Li},
 booktitle={Advances in Neural Information Processing Systems (NeurIPS)},
 year={2024}
}

@article{Kim2021DriveGANTA,
  title={DriveGAN: Towards a Controllable High-Quality Neural Simulation},
  author={Seung Wook Kim and Jonah Philion and Antonio Torralba and Sanja Fidler},
  journal={2021 IEEE/CVF Conference on Computer Vision and Pattern Recognition (CVPR)},
  year={2021},
  pages={5816-5825}
}

@misc{chou2026cityrag,
        title     = {CityRAG: Stepping Into a City via Spatially-Grounded Video Generation},
        author    = {Chou, Gene and Herrmann, Charles and Genova, Kyle and Deng, Boyang and Peng, Songyou and Hariharan, Bharath and Zhang, Jason Y. and Snavely, Noah and Henzler, Philipp},
        year         = {2026},
        eprint       = {2604.19741},
        archivePrefix= {arXiv},
        primaryClass = {cs.CV},
        url = {https://arxiv.org/abs/2604.19741}
      }

@inproceedings{ke2021musiq,
  title={Musiq: Multi-scale image quality transformer},
  author={Ke, Junjie and Wang, Qifei and Wang, Yilin and Milanfar, Peyman and Yang, Feng},
  booktitle={2021 IEEE/CVF International Conference on Computer Vision (ICCV)},
  pages={5128--5137},
  year={2021},
  organization={IEEE}
}

@inproceedings{seo2026grounding,
  title={{Seoul World Model: Grounding World Simulation Models in a Real-World Metropolis}},
  author={Seo, Junyoung and Choi, Hyunwook and Kwon, Minkyung and Choi, Jinhyeok and Jin, Siyoon and Lee, Gayoung and Kim, Junho and Lee, JoungBin and Gu, Geonmo and Han, Dongyoon and Yun, Sangdoo and Kim, Seungryong and Kim, Jin-Hwa},
  booktitle={European Conference on Computer Vision (ECCV)},
  year={2026}
}

@article{mosaicmem2026,
  title   = {MosaicMem: Hybrid Spatial Memory for Controllable Video World Models},
  author  = {Wei Yu and Runjia Qian and Yumeng Li and Liquan Wang and Songheng Yin and Sri Siddarth Chakaravarthy P and Dennis Anthony and Yang Ye and Yidi Li and Weiwei Wan and Animesh Garg},
  journal = {arXiv preprint arXiv:2603.17117},
  year    = {2026}
}

@InProceedings{Li_2025_vmem,
    author    = {Li, Runjia and Torr, Philip and Vedaldi, Andrea and Jakab, Tomas},
    title     = {VMem: Consistent Interactive Video Scene Generation with Surfel-Indexed View Memory},
    booktitle = {Proceedings of the IEEE/CVF International Conference on Computer Vision (ICCV)},
    month     = {October},
    year      = {2025},
    pages     = {25690-25699}
  }

@inproceedings{xiaoworldmem,
  title={WorldMem: Long-term Consistent World Simulation with Memory},
  author={Xiao, Zeqi and Yushi, LAN and Zhou, Yifan and Ouyang, Wenqi and Yang, Shuai and Zeng, Yanhong and Pan, Xingang},
  booktitle={The Thirty-ninth Annual Conference on Neural Information Processing Systems},
  year={2025}
}

@inproceedings{epona2025,
        author = {Zhang, Kaiwen and Tang, Zhenyu and Hu, Xiaotao and Pan, Xingang and Guo, Xiaoyang and Liu, Yuan and Huang,
        Jingwei and Yuan, Li and Zhang, Qian and Long, Xiao-Xiao and Cao, Xun and Yin, Wei},
        title = {Epona: Autoregressive Diffusion World Model for Autonomous Driving},
        booktitle = {Proceedings of the IEEE/CVF International Conference on Computer Vision (ICCV)},
        year = {2025}
      }

@InProceedings{mousakhan2025orbis,
  author       = "A. Mousakhan and S. Mittal and S. Galesso and K. Farid and T. Brox",
  title        = "Orbis: Overcoming Challenges of Long-Horizon Prediction in Driving World Models",
  booktitle    = "Advances in Neural Information Processing Systems (NeurIPS)",
  month        = " ",
  year         = "2025"
}

@inproceedings{chen2025diffusion,
author = {Chen, Boyuan and Mons\'{o}, Diego Mart\'{\i} and Du, Yilun and Simchowitz, Max and Tedrake, Russ and Sitzmann, Vincent},
title = {Diffusion forcing: next-token prediction meets full-sequence diffusion},
year = {2024},
isbn = {9798331314385},
publisher = {Curran Associates Inc.},
address = {Red Hook, NY, USA},
booktitle = {Proceedings of the 38th International Conference on Neural Information Processing Systems (NeurIPS)},
articleno = {759},
numpages = {45},
location = {Vancouver, BC, Canada},
}

@misc{streetnvs2025,
      title={Effective Multi-sensor Conditioning for Street-view Novel-view Synthesis}, 
      author={Zhengfei Kuang and Adam Sun and Liyuan Zhu and Tong Wu and Shengqu Cai and Jonathan Tremblay and Iro Armeni and Ehsan Adeli and Lior Yariv and Gordon Wetzstein},
      year={2026},
      eprint={2606.01590},
      archivePrefix={arXiv},
      primaryClass={cs.CV},
      url={https://arxiv.org/abs/2606.01590}, 
}

@InProceedings{vista4d2025,
    author    = {Lin, {Kuan Heng} and Liu, Zhizheng and Salamanca, Pablo and Kant, Yash and Burgert, Ryan and Xu, Yuancheng and Namekata, Koichi and Zhao, Yiwei and Zhou, Bolei and Goldblum, Micah and Debevec, Paul and Yu, Ning},
    title     = {{Vista4D}: Video Reshooting with 4D Point Clouds},
    booktitle = {Proceedings of the IEEE/CVF Conference on Computer Vision and Pattern Recognition (CVPR)},
    month     = {June},
    year      = {2026},
    pages     = {32671--32682}
}

@misc{nvidia2025cosmos,
      title={World Simulation with Video Foundation Models for Physical AI}, 
      author={NVIDIA and : and Arslan Ali and Junjie Bai and Maciej Bala and Yogesh Balaji and Aaron Blakeman and Tiffany Cai and Jiaxin Cao and Tianshi Cao and Elizabeth Cha and Yu-Wei Chao and Prithvijit Chattopadhyay and Mike Chen and Yongxin Chen and Yu Chen and Shuai Cheng and Yin Cui and Jenna Diamond and Yifan Ding and Jiaojiao Fan and Linxi Fan and Liang Feng and Francesco Ferroni and Sanja Fidler and Xiao Fu and Ruiyuan Gao and Yunhao Ge and Jinwei Gu and Aryaman Gupta and Siddharth Gururani and Imad El Hanafi and Ali Hassani and Zekun Hao and Jacob Huffman and Joel Jang and Pooya Jannaty and Jan Kautz and Grace Lam and Xuan Li and Zhaoshuo Li and Maosheng Liao and Chen-Hsuan Lin and Tsung-Yi Lin and Yen-Chen Lin and Huan Ling and Ming-Yu Liu and Xian Liu and Yifan Lu and Alice Luo and Qianli Ma and Hanzi Mao and Kaichun Mo and Seungjun Nah and Yashraj Narang and Abhijeet Panaskar and Lindsey Pavao and Trung Pham and Morteza Ramezanali and Fitsum Reda and Scott Reed and Xuanchi Ren and Haonan Shao and Yue Shen and Stella Shi and Shuran Song and Bartosz Stefaniak and Shangkun Sun and Shitao Tang and Sameena Tasmeen and Lyne Tchapmi and Wei-Cheng Tseng and Jibin Varghese and Andrew Z. Wang and Hao Wang and Haoxiang Wang and Heng Wang and Ting-Chun Wang and Fangyin Wei and Jiashu Xu and Dinghao Yang and Xiaodong Yang and Haotian Ye and Seonghyeon Ye and Xiaohui Zeng and Jing Zhang and Qinsheng Zhang and Kaiwen Zheng and Andrew Zhu and Yuke Zhu},
      year={2025},
      eprint={2511.00062},
      archivePrefix={arXiv},
      primaryClass={cs.CV},
      url={https://arxiv.org/abs/2511.00062}, 
}

@misc{cui2025selfforcing,
      title={Self-Forcing++: Towards Minute-Scale High-Quality Video Generation}, 
      author={Justin Cui and Jie Wu and Ming Li and Tao Yang and Xiaojie Li and Rui Wang and Andrew Bai and Yuanhao Ban and Cho-Jui Hsieh},
      year={2025},
      eprint={2510.02283},
      archivePrefix={arXiv},
      primaryClass={cs.CV},
      url={https://arxiv.org/abs/2510.02283}, 
}

@InProceedings{henschel2024streamingt2v,
    author    = {Henschel, Roberto and Khachatryan, Levon and Poghosyan, Hayk and Hayrapetyan, Daniil and Tadevosyan, Vahram and Wang, Zhangyang and Navasardyan, Shant and Shi, Humphrey},
    title     = {StreamingT2V: Consistent, Dynamic, and Extendable Long Video Generation from Text},
    booktitle = {Proceedings of the IEEE/CVF Conference on Computer Vision and Pattern Recognition (CVPR)},
    month     = {June},
    year      = {2025},
    pages     = {2568-2577}
}

@article{wu2025videoworldmodels,
  title={Video world models with long-term spatial memory},
  author={Wu, Tong and Yang, Shuai and Po, Ryan and Xu, Yinghao and Liu, Ziwei and Lin, Dahua and Wetzstein, Gordon},
  journal={Advances in Neural Information Processing Systems (NeurIPS)},
  volume={38},
  pages={49371--49393},
  year={2026}
}

@inproceedings{ren2025gen3c,
  title={GEN3C: 3D-Informed World-Consistent Video Generation with Precise Camera Control},
  author={Ren, Xuanchi and Shen, Tianchang and Huang, Jiahui and Ling, Huan and Lu, Yifan and Nimier-David, Merlin and M\"uller, Thomas and Keller, Alexander and Fidler, Sanja and Gao, Jun},
  booktitle={Proceedings of the IEEE/CVF Conference on Computer Vision and Pattern Recognition (CVPR)},
  year={2025}
}

@misc{worldwarp2025,
      title={WorldWarp: Propagating 3D Geometry with Asynchronous Video Diffusion}, 
      author={Hanyang Kong and Xingyi Yang and Xiaoxu Zheng and Xinchao Wang},
      year={2025},
      eprint={2512.19678},
      archivePrefix={arXiv},
      primaryClass={cs.CV},
      url={https://arxiv.org/abs/2512.19678}, 
}

@inproceedings{yang2024genad,
  title={Generalized Predictive Model for Autonomous Driving},
  author={Jiazhi Yang and Shenyuan Gao and Yihang Qiu and Li Chen and Tianyu Li and Bo Dai and Kashyap Chitta and Penghao Wu and Jia Zeng and Ping Luo and Jun Zhang and Andreas Geiger and Yu Qiao and Hongyang Li},
  booktitle={Proceedings of the IEEE/CVF Conference on Computer Vision and Pattern Recognition (CVPR)},
  year={2024}
}

@article{peebles2023dit,
  title={Scalable Diffusion Models with Transformers},
  author={William S. Peebles and Saining Xie},
  journal={IEEE/CVF International Conference on Computer Vision (ICCV)},
  year={2023},
  pages={4172-4182}
}

@inproceedings{park2019SPADE,
  title={Semantic Image Synthesis with Spatially-Adaptive Normalization},
  author={Park, Taesung and Ming-Yu Liu and Ting-Chun Wang and Jun-Yan Zhu},
  booktitle={Proceedings of the IEEE Conference on Computer Vision and Pattern Recognition (CVPR)},
  year={2019}
}

@inproceedings{hong2024slowfastgen,
  title={SlowFast-VGen: Slow-Fast Learning for Action-Driven Long Video Generation},
  author={Hong, Yining and Liu, Beide and Wu, Maxine and Zhai, Yuanhao and Chang, Kai-Wei and Li, Linjie and Lin, Kevin and Lin, Chung-Ching and Wang, Jianfeng and Yang, Zhengyuan and Wu, Ying Nian and Wang, Lijuan},
  booktitle={International Conference on Learning Representations (ICLR)},
  year={2025}
}

@article{oquab2023dinov2,
  title={{DINOv2}: Learning Robust Visual Features without Supervision},
  author={Oquab, Maxime and Darcet, Timoth{\'e}e and Moutakanni, Th{\'e}o and Vo, Huy and Szafraniec, Marc and Khalidov, Vasil and Fernandez, Pierre and Haziza, Daniel and Massa, Francisco and El-Nouby, Alaaeldin and others},
  journal={Transactions on Machine Learning Research (TMLR)},
  year={2024}
}

@inproceedings{lipman2022flow,
  title={Flow Matching for Generative Modeling},
  author={Lipman, Yaron and Chen, Ricky T. Q. and Ben-Hamu, Heli and Nickel, Maximilian and Le, Matt},
  booktitle={International Conference on Learning Representations (ICLR)},
  year={2023}
}

@inproceedings{liu2022flow,
  title={Flow Straight and Fast: Learning to Generate and Transfer Data with Rectified Flow},
  author={Liu, Xingchao and Gong, Chengyue and Liu, Qiang},
  booktitle={International Conference on Learning Representations (ICLR)},
  year={2023}
}

@inproceedings{esser2024scaling,
  title={Scaling Rectified Flow Transformers for High-Resolution Image Synthesis},
  author={Esser, Patrick and Kulal, Sumith and Blattmann, Andreas and Entezari, Rahim and M{\"u}ller, Jonas and Saini, Harry and Levi, Yam and Lorenz, Dominik and Sauer, Axel and Boesel, Frederic and Podell, Dustin and Dockhorn, Tim and English, Zion and Rombach, Robin},
  booktitle={International Conference on Machine Learning (ICML)},
  year={2024}
}

@article{bai2024factorized,
  title={Factorized Visual Tokenization and Generation},
  author={Bai, Zechen and Gao, Jianxiong and Gao, Ziteng and Wang, Pichao and Zhang, Zheng and He, Tong and Shou, Mike Zheng},
  journal={arXiv preprint arXiv:2411.16681},
  year={2024}
}

@INPROCEEDINGS{caesar2021nuplan, 
  title={NuPlan: A closed-loop ML-based planning benchmark for autonomous vehicles},
  author={Holger Caesar and Juraj Kabzan and Kok Seang Tan and Whye Kit Fong and Eric Wolff and Alex Lang and Luke Fletcher and Oscar Beijbom and Sammy Omari},
  booktitle={CVPR ADP3 workshop},
  year=2021
}

@InProceedings{sun2020waymoopen, 
  author={Ettinger, Scott and Cheng, Shuyang and Caine, Benjamin and Liu, Chenxi and Zhao, Hang and Pradhan, Sabeek and Chai, Yuning and Sapp, Ben and Qi, Charles R. and Zhou, Yin and Yang, Zoey and Chouard, Aur'elien and Sun, Pei and Ngiam, Jiquan and Vasudevan, Vijay and McCauley, Alexander and Shlens, Jonathon and Anguelov, Dragomir}, 
  title={Large Scale Interactive Motion Forecasting for Autonomous Driving: The Waymo Open Motion Dataset}, 
  booktitle={Proceedings of the IEEE/CVF International Conference on Computer Vision (ICCV)}, 
  month={October}, 
  year={2021}, 
  pages={9710-9719} 
}

@InProceedings{zheng2026versecrafter,
  author    = {Zheng, Sixiao and Yin, Minghao and Hu, Wenbo and Li, Xiaoyu and Shan, Ying and Fu, Yanwei},
  title     = {VerseCrafter: Dynamic Realistic Video World Model with 4D Geometric Control},
  booktitle = {Proceedings of the IEEE/CVF Conference on Computer Vision and Pattern Recognition (CVPR)},
  month     = {June},
  year      = {2026},
  pages     = {40277-40290}
}

@inproceedings{lpips,
  title={The unreasonable effectiveness of deep features as a perceptual metric},
  author={Zhang, Richard and Isola, Phillip and Efros, Alexei A and Shechtman, Eli and Wang, Oliver},
  booktitle={Proceedings of the IEEE Conference on Computer Vision and Pattern Recognition (CVPR)},
  pages={586--595},
  year={2018}
}

@inproceedings{fid,
  title={GANs Trained by a Two Time-Scale Update Rule Converge to a Local Nash Equilibrium},
  author={Heusel, Martin and Ramsauer, Hubert and Unterthiner, Thomas and Nessler, Bernhard and Hochreiter, Sepp},
  booktitle={Advances in Neural Information Processing Systems (NeurIPS)},
  volume={30},
  year={2017}
}

@inproceedings{fvd,
  title={{FVD}: A new Metric for Video Generation},
  author={Unterthiner, Thomas and van Steenkiste, Sjoerd and Kurach, Karol and Marinier, Rapha\"{e}l and Michalski, Marcin and Gelly, Sylvain},
  booktitle={ICLR Workshop on Deep Generative Models for Higher-Order Tensors in Multi-Modal Applications},
  year={2019},
}

@inproceedings{vggt,
  title={VGGT: Visual Geometry Grounded Transformer},
  author={Wang, Jianyuan and Chen, Minghao and Karaev, Nikita and Vedaldi, Andrea and Rupprecht, Christian and Novotny, David},
  booktitle={Proceedings of the IEEE/CVF Conference on Computer Vision and Pattern Recognition (CVPR)},
  year={2025}
}

@inproceedings{lightglue,
  title     = {{LightGlue: Local Feature Matching at Light Speed}},
  author    = {Lindenberger, Philipp and Sarlin, Paul-Edouard and Pollefeys, Marc},
  booktitle = {Proceedings of the IEEE/CVF International Conference on Computer Vision (ICCV)},
  year      = {2023}
}

@article{depthanything3,
  title={Depth Anything 3: recovering the visual space from any views},
  author={Haotong Lin and Sili Chen and Jun Hao Liew and Donny Y. Chen and Zhenyu Li and Guang Shi and Jiashi Feng and Bingyi Kang},
  journal={arXiv preprint arXiv:2511.10647},
  year={2025}
}

@misc{opensora,
  author = {Zangwei Zheng and others},
  title = {Open-Sora: Democratizing Efficient Video Production for All},
  month = {March},
  year = {2024},
  url = {https://github.com/hpcaitech/Open-Sora}
}

@InProceedings{metric3d,
    author = {Yin, Wei and others},
    title = {Metric3D: Towards Zero-shot Metric 3D Prediction from A Single Image},
    booktitle = {Proceedings of the IEEE/CVF International Conference on Computer Vision (ICCV)},
    year = {2023},
    pages = {9043-9053}
}

@inproceedings{Li2026DROIDW,
  author    = {Li, Moyang and Zhu, Zihan and Pollefeys, Marc and Barath, Daniel},
  title     = {DROID-SLAM in the Wild},
  booktitle = {Proceedings of the IEEE/CVF Conference on Computer Vision and Pattern Recognition (CVPR)},
  year      = {2026}
}

@inproceedings{teed2021droidSalm,
  title={DROID-SLAM: Deep Visual SLAM for Monocular, Stereo, and RGB-D Cameras},
  author={Teed, Zachary and Deng, Jia},
  booktitle={Advances in Neural Information Processing Systems (NeurIPS)},
  year={2021}
}

@inproceedings{
  chen2025omnire,
  title={OmniRe: Omni Urban Scene Reconstruction},
  author={Ziyu Chen and Jiawei Yang and Jiahui Huang and Riccardo de Lutio and Janick Martinez Esturo and Boris Ivanovic and Or Litany and Zan Gojcic and Sanja Fidler and Marco Pavone and Li Song and Yue Wang},
  booktitle={The Thirteenth International Conference on Learning Representations (ICLR)},
  year={2025}
}

@inproceedings{yan2024street,
    title={Street Gaussians: Modeling Dynamic Urban Scenes with Gaussian Splatting}, 
    author={Yunzhi Yan and Haotong Lin and Chenxu Zhou and Weijie Wang and Haiyang Sun and Kun Zhan and Xianpeng Lang and Xiaowei Zhou and Sida Peng},
    booktitle={European Conference on Computer Vision (ECCV)},
    year={2024}
}

@inproceedings{orbis2_2026,
  author    = {Mittal, Sudhanshu and Mousakhan, Arian and Galesso, Silvio and 
               Farid, Karim and Dienert, Johannes and Sahay, Rajat and 
               Ali, Muhammad and Brox, Thomas},
  title     = {Orbis 2: A Hierarchical World Model for Driving},
  booktitle = {Advances in Neural Information Processing Systems (NeurIPS)},
  year      = {2026}
}

@INPROCEEDINGS{caesar2020nuscenes,
  title={nuScenes: A multimodal dataset for autonomous driving},
  author={Holger Caesar and Varun Bankiti and Alex H. Lang and Sourabh Vora and 
          Venice Erin Liong and Qiang Xu and Anush Krishnan and Yu Pan and 
          Giancarlo Baldan and Oscar Beijbom}, 
  booktitle={Proceedings of the IEEE/CVF Conference on Computer Vision and Pattern Recognition (CVPR)},
  year=2020
}

@InProceedings{mars,
  author    = {Li, Yiming and Li, Zhiheng and Chen, Nuo and Gong, Moonjun and Lyu, Zonglin Caps and Wang, Zehong and Jiang, Peili and Feng, Chen},
  title     = {Multiagent Multitraversal Multimodal Self-Driving: Open MARS Dataset},
  booktitle = {Proceedings of the IEEE/CVF Conference on Computer Vision and Pattern Recognition (CVPR)},
  month     = {June},
  year      = {2024},
  pages     = {22041-22051}
}

@article{team2026advancing,
      title={Advancing Open-source World Models}, 
      author={Robbyant Team and Zelin Gao and Qiuyu Wang and Yanhong Zeng and Jiapeng Zhu and Ka Leong Cheng and Yixuan Li and Hanlin Wang and Yinghao Xu and Shuailei Ma and Yihang Chen and Jie Liu and Yansong Cheng and Yao Yao and Jiayi Zhu and Yihao Meng and Kecheng Zheng and Qingyan Bai and Jingye Chen and Zehong Shen and Yue Yu and Xing Zhu and Yujun Shen and Hao Ouyang},
      journal={arXiv preprint arXiv:2601.20540},
      year={2026}
}

@inproceedings{
    dai2025fantasyworld,
    title={FantasyWorld: Geometry-Consistent World Modeling via Unified Video and 3D Prediction},
    author={Yixiang Dai and Fan Jiang and Chiyu Wang and Mu Xu and Yonggang Qi},
    booktitle={The Fourteenth International Conference on Learning Representations (ICLR)},
    year={2026},
}

@article{hyworld2025,
  title={HY-World 1.5: A Systematic Framework for Interactive World Modeling with Real-Time Latency and Geometric Consistency},
  author={Team HunyuanWorld},
  journal={arXiv preprint},
  year={2025}
}

@article{mao2025yume,
  title={Yume: An Interactive World Generation Model},
  author={Mao, Xiaofeng and Lin, Shaoheng and Li, Zhen and Li, Chuanhao and Peng, Wenshuo and He, Tong and Pang, Jiangmiao and Chi, Mingmin and Qiao, Yu and Zhang, Kaipeng},
  journal={arXiv preprint arXiv:2507.17744},
  year={2025}
}

@article{chen2025geodrive,
  title   = {GeoDrive: 3D Geometry-Informed Driving World Model with Precise Action Control},
  author  = {Chen, Anthony and Zheng, W. and Wang, Y. and Zhang, X. and Zhan, Kun and Jia, P. and others},
  journal = {arXiv preprint arXiv:2505.22421},
  year    = {2025}
}

@inproceedings{yan2024streetcrafter,
  title={StreetCrafter: Street View Synthesis with Controllable Video Diffusion Models},
  author={Yan, Yunzhi and Xu, Zhen and Lin, Haotong and Jin, Haian and Guo, Haoyu and Wang, Yida and Zhan, Kun and Lang, Xianpeng and Bao, Hujun and Zhou, Xiaowei and Peng, Sida},
  booktitle={Proceedings of the IEEE/CVF Conference on Computer Vision and Pattern Recognition (CVPR)},
  year={2025},
}

@article{fischer2024dynamic,
  title={Dynamic 3d gaussian fields for urban areas},
  author={Fischer, Tobias and Kulhanek, Jonas and Bulo, Samuel Rota and Porzi, Lorenzo and Pollefeys, Marc and Kontschieder, Peter},
  journal={arXiv preprint arXiv:2406.03175},
  year={2024}
}

@inproceedings{yang2023unisim,
  title={Unisim: A neural closed-loop sensor simulator},
  author={Yang, Ze and Chen, Yun and Wang, Jingkang and Manivasagam, Sivabalan and Ma, Wei-Chiu and Yang, Anqi Joyce and Urtasun, Raquel},
  booktitle={2023 IEEE/CVF Conference on Computer Vision and Pattern Recognition (CVPR)},
  pages={1389--1399},
  year={2023},
  organization={IEEE}
}

@inproceedings{tonderski2024neurad,
  title={Neurad: Neural rendering for autonomous driving},
  author={Tonderski, Adam and Lindstr{\"o}m, Carl and Hess, Georg and Ljungbergh, William and Svensson, Lennart and Petersson, Christoffer},
  booktitle={2024 IEEE/CVF Conference on Computer Vision and Pattern Recognition (CVPR)},
  pages={14895--14904},
  year={2024},
  organization={IEEE}
}

@inproceedings{huang2024vbench,
  title={Vbench: Comprehensive benchmark suite for video generative models},
  author={Huang, Ziqi and He, Yinan and Yu, Jiashuo and Zhang, Fan and Si, Chenyang and Jiang, Yuming and Zhang, Yuanhan and Wu, Tianxing and Jin, Qingyang and Chanpaisit, Nattapol and others},
  booktitle={2024 IEEE/CVF Conference on Computer Vision and Pattern Recognition (CVPR)},
  pages={21807--21818},
  year={2024},
  organization={IEEE}
}
\bibliographystyle{iclr2027_conference}

\newpage
\appendix
\section*{Appendix}
\addcontentsline{toc}{section}{Appendices}

\startcontents[appendices]
\printcontents[appendices]{l}{1}{\vspace{-1.5em}}

\newpage
\section{Overview of Persistent Scene Context Methods}
\label{app:related_memory}

Long-horizon world models need a representation of the scene that persists beyond the context window. Existing approaches differ in whether this memory is built from the model's own generated frames or grounded in real observations, and in whether it is aligned to past or to upcoming viewpoints. \Cref{tab:related_memory} summarizes them along these axes.

\begin{table}[ht!]
\centering
\caption{Comparison of scene memory approaches. 
\emph{Anchoring}: \emph{retrospective} methods condition on geometry aligned to past or current viewpoints, while \emph{prospective} methods render geometry to the future camera pose before generation. 
\emph{Memory input}: the data used to construct the persistent scene representation, not the full input to the generative model.
\emph{Dyn.\ objects}: whether the conditioning signal itself represents dynamic objects, or instead delegates dynamics entirely to the generative prior.
\emph{Dyn.\ scene}: whether the method's intended output contains a dynamic scene with moving objects, or is restricted to static environments.}
\label{tab:related_memory}
\resizebox{\linewidth}{!}{%
\begin{tabular}{llllll}
\toprule
Method & Memory input & Anchoring & Memory representation & Dyn.\ objects & Dyn.\ scene \\
\midrule
\multicolumn{6}{l}{\textit{Generated content memory}} \\[2pt]
StreamingT2V~\citep{henschel2024streamingt2v}  & Previously generated frames & Retrospective & Frame buffer + global token & \cmark & \cmark \\
WorldMem~\citep{xiaoworldmem}                  & Previously generated frames & Retrospective & Retrieved frame index       & \cmark & \cmark \\
SlowFastGen~\citep{hong2024slowfastgen}         & Previously generated frames & Retrospective & LoRA weights                & \cmark & \cmark \\
WorldWarp~\citep{worldwarp2025}                 & Previously generated frames & Prospective   & Online 3DGS                 & \xmark         & \xmark         \\
VMem~\citep{Li_2025_vmem}                       & Previously generated frames & Retrospective & Surfel map                  & \xmark         & \xmark         \\
MosaicMem~\citep{mosaicmem2026}                 & Previously generated frames & Retrospective & 3D memory patches           & \xmark         & \cmark \\
SpMem~\citep{wu2025videoworldmodels}            & Previously generated frames & Prospective   & Static point maps           & \xmark         & \cmark \\
\midrule
\multicolumn{6}{l}{\textit{Real-world observations}} \\[2pt]
CityRAG~\citep{chou2026cityrag}                 & Route imagery (multi-pass)  & Retrospective & Image database          & \cmark & \cmark \\
Seoul WM~\citep{seo2026grounding}               & Pre-built geospatial map    & Retrospective & Geospatial image map    & \cmark & \cmark \\
GEN3C~\citep{ren2025gen3c}                      & Monocular seed image        & Prospective   & Point cloud cache       & \xmark         & \xmark         \\
GeoDrive~\citep{chen2025geodrive}               & Monocular seed frame     & Prospective   & Decomposed point cloud  & \cmark & \cmark \\
StreetNVS~\citep{streetnvs2025}                 & LiDAR + multi-camera        & Prospective   & LiDAR point cloud       & \xmark         & \cmark \\
Vista4D~\citep{vista4d2025}                     & Training frames             & Prospective   & 4D point cloud          & \cmark & \cmark \\
VerseCrafter~\citep{zheng2026versecrafter}      & Input image + obj.\ trajs.  & Prospective   & 4D control maps & \cmark & \cmark \\
\textbf{Ours}                                   & Monocular driving log       & Prospective   & Static point cloud (SLAM)                & \xmark         & \cmark \\
\bottomrule
\end{tabular}}
\end{table}

\section{Reconstruction and Rendering Details}
\label{app:point_cloud_rendering}

\paragraph{Point-cloud filtering.}

Filtering is performed once, \emph{offline}, on the complete unprojected point cloud accumulated over the full traversal, rather than per rendered frame.
Five sequential filters clean the raw point cloud, ordered from cheap local per-pixel tests to more expensive cross-view and voxel-density-based checks.
The uncertainty and multi-view geometric filters reapply the consistency criteria that DROID-W already computes internally during optimization, now as explicit, tunable filters on the point cloud.
We also add a disparity/depth threshold, a photometric color-agreement test, and a final 3D outlier removal step.
Together, these filters suppress unreliable geometry arising from dynamic objects, occlusions, and mismatched correspondences, reducing erroneous reprojections and improving the stability of the subsequent rendering stage (thresholds in \Cref{tab:hp_rendering}):
\begin{enumerate}
    \item \textbf{Disparity filter.} Points whose source disparity is below a threshold, or whose depth exceeds $d_\text{max}$ are discarded.
    \item \textbf{Uncertainty filter.} Points whose source pixel has tracking uncertainty
    exceeding $\tau_\text{unc}$ are removed.
    \item \textbf{Multi-view geometric filter.} A point survives only if
    it reprojects into at least $N_\text{mv}$ other keyframes
    with relative depth error below $\tau_\text{mv}$.
    \item \textbf{Multi-view photometric filter.} A point is kept only if its reprojected color agrees (mean L1 $< \tau_\text{photo}$) with at least $N_\text{photo}$ other frames.
    \item \textbf{Voxel outlier filter.} Points whose voxel of side length $s_\text{vox}$ contains fewer than $k_\text{vox}$ points (including the point itself) are discarded.
\end{enumerate}

\paragraph{Per-frame projection.}
In contrast to filtering, projection is performed \emph{online}, once per target frame. Each world-space point is transformed into the camera frame of the target pose $P_n$, yielding camera coordinates $(X, Y, Z)$, and projected with the pinhole model
\begin{equation}
    u = f_x \frac{X}{Z} + c_x, \quad v = f_y \frac{Y}{Z} + c_y,
\end{equation}
where $f_x, f_y, c_x, c_y$ are the intrinsics of the target camera at the model's input resolution $H \times W$. During training and replay, $P_n$ is the DROID-W pose of the recorded frame. For counterfactual and perturbed rollouts, it is the modified pose.
Each point is projected to the nearest pixel. Points with $Z \le 0.01$ or outside the image are discarded. We use a single-pixel footprint and apply no hole filling. Instead of a hard z-buffer, all points projecting to the same pixel are composited with a depth-gated weighted blend. Let $Z_\text{front}$ be the smallest depth among them. Each point receives the weight
\begin{equation}
  w(Z) = \frac{1}{Z^2}\,\mathbb{1}\bigl[Z \le (1+\delta)\,Z_\text{front}\bigr],
\end{equation}
and the pixel color is the weighted average $\bar{\mathbf{c}} = \sum_i w(Z_i)\,\mathbf{c}_i / \sum_i w(Z_i)$. We use a permissive gate ($\delta{=}2$), so occlusion is resolved mainly by the inverse-square weighting: a point at three times the front depth receives $1/9$ of the front point's weight. Compared to a nearest-point z-buffer, this averaging suppresses flickering caused by near-coincident points on the sparse cloud. The accompanying depth map stores $Z_\text{front}$ per pixel.
Rendering from the complete scene point cloud, rather than only the points observed up to the current frame, matches the simulation setting we target: the route has already been traversed, and the reconstructed map is available before rollout begins, so the model replays the logged route, or counterfactual variants of it, such as altered speed, lane offsets, or other traffic participants.

\section{Base World Model Details}\label{app:wm_details}

We evaluate our conditioning on three base world models spanning different scales and architectures: Orbis~v1~\citep{mousakhan2025orbis}, Orbis~v2~\citep{orbis2_2026}, and Cosmos2.5-Predict-2B~\citep{nvidia2025cosmos}. We summarize only the components relevant to our conditioning and refer to the original works for details. For all learned variants, we initialize from the released weights and fine-tune the full backbone jointly with the conditioning modules; latent blending uses the released weights unchanged. Geo-reference frames are encoded with the frozen tokenizer of the respective model.

\textbf{Orbis~v1.}
A 469M-parameter continuous-latent autoregressive world model trained from scratch on 280h of front-camera driving video. Frames are encoded by a hybrid VQGAN-style tokenizer (86M encoder / 91M decoder) into semantic (DINOv2-distilled) and detail token streams. The generator is a causal STDiT of $L{=}24$ blocks (Swin-windowed spatial attention, causal temporal attention, adaLN-Zero~\citep{peebles2023dit} timestep conditioning), trained with a rectified-flow objective~\citep{lipman2022flow, liu2022flow, esser2024scaling}: it learns a velocity field $v_\theta(\mathbf{z}_t;\mathbf{z}^{1:N_{\mathrm{ctx}}},t)$ that transports noise to a clean latent via Euler integration. At inference, the model rolls out frame-by-frame: given $N_{\mathrm{ctx}}$ context latents it denoises one new frame, appends it to the context, drops the oldest frame, and repeats, generating long-horizon videos one step at a time.

\textbf{Orbis~v2.}
A hierarchical extension of Orbis~v1 with a 512M high-level abstract predictor and a 555M low-level detail predictor, 1067M params total, both flow-matching-based. The abstract predictor operates on a compact 16-dim DINOv2-aligned latent at 2Hz, predicting subgoals over a 3s strided context. The detail predictor generates pixel-aligned frames at 10Hz conditioned on these subgoals, reusing the Orbis~v1 tokenizer. Training uses diffusion-forcing pretraining followed by teacher-forcing fine-tuning, on 5,890h of video (OpenDV, BDD100K, PhysicalAI-AV, NATIX) at $512\times288$. Steering control is added via adaLN on the abstract branch, fine-tuned on 500h with IMU-derived trajectories. Rollout is likewise autoregressive but at two nested rates: the abstract predictor advances its sliding window one frame at a time to emit a new subgoal, and the detail predictor advances its own window 5 frames at a time, generating the intervening fine-grained frames conditioned on the latest predicted subgoal.

\textbf{Cosmos2.5-Predict-2B.}
NVIDIA's 2B-parameter flow-matching video DiT unifying Text2World, Image2World, and Video2World generation. The backbone has 32 blocks, model dimension 2048, 16 attention heads, AdaLN-LoRA conditioning, and 3D RoPE. Video is compressed via a causal WAN2.1 VAE ($4\times8\times8$) plus $1\times2\times2$ patchification, generating $\sim$5.8s clips (93 frames, 16fps). Pretrained on 200M curated clips spanning driving, robotics, and human-dynamics domains, then refined with supervised fine-tuning, RL-based post-training, and timestep distillation. Since the fixed context window only covers $\sim$5.8s, longer rollouts are produced chunk-autoregressively whereby the last frames of the previously generated chunk are fed back via a frame-replacement strategy to condition the denoising of the next chunk.

\section{Conditioning Mechanisms Details}\label{sec:appendix_cond_mechanism}

\subsection{Latent Blending}
\label{sec:latent_blending}

Latent blending injects the geo-reference signal without training or architectural changes by correcting the sampler state in latent space between ODE steps. The rectified-flow forward process noises a clean latent $Z_0$ as $z_t = (1-t)\,Z_0 + t\,\epsilon$ with $\epsilon \sim \mathcal{N}(0, I)$, so that $t{=}0$ is clean and $t{=}1$ is pure noise. After each Euler step that brings the state to timestep $t'$, we construct an anchor by applying this forward process to the geo-reference latent,
\begin{equation}
  Z^{\text{anc}}(t') = (1-t')\,Z^{\text{geo}} + t'\,\epsilon,
  \label{eq:anchor}
\end{equation}
i.e.\ the state that a denoising trajectory ending at $Z^{\text{geo}}$ would visit at $t'$.
During the first $\lfloor \rho_{\text{geo}} S \rfloor$ of the $S$ ODE steps, the state is interpolated toward the anchor at covered positions,
\begin{equation}
  z_{t'}[M] \leftarrow w(t')\, Z^{\text{anc}}(t')[M] + \bigl(1-w(t')\bigr)\, z_{t'}[M],
  \qquad w(t') = \min(s_{\text{geo}}\, t',\, 1),
  \label{eq:anchor_blend}
\end{equation}
where $[M]$ selects the tokens of the $F_{\text{pred}}$ predicted frames that are covered by the geo-reference, broadcast over channels. Uncovered tokens and context frames are left unchanged. Blending is therefore restricted to the noise-dominated regime, where it fixes the coarse layout at little cost to diversity, while the remaining steps refine appearance freely. With $S{=}30$, $\rho_{\text{geo}}{=}0.2$ and $s_{\text{geo}}{=}1$, blending is active for the first 6 steps with $w \in [0.8, 1]$.

\subsection{Concatenation}
\label{sec:concatenation}

%& channel concat.\ at patch embed., input only
The most direct injection point is the patch embedding layer itself: rather than modulating features deeper in the network, the geo-reference signal is fused with the noisy input latent along the channel axis before the first projection. Every block downstream is an unmodified STDiT block, which makes this the cheapest learned variant. \newline
The patch embedding is a $1{\times}1$ convolution $W_{\mathrm{embed}} \in \mathbb{R}^{D \times C}$ mapping tokenizer-latent channels to the backbone width $D$. To admit the geo-reference latent, $W_{\mathrm{embed}}$ is widened along the input axis to $W_{\mathrm{embed}}' \in \mathbb{R}^{D \times 2C}$,
\vspace{-0.05cm}
\begin{equation}
  W_{\mathrm{embed}}' = \bigl[\,W_{\mathrm{embed}} \;\big|\; \mathbf{0}\,\bigr],
  \label{eq:concat_widen}
\end{equation}
copying the pretrained weights into the first $C$ columns and zero-initializing the new ones (bias unchanged), so the model computes the pretrained function at initialization. Only the input side changes. The output head is untouched.

\subsection{AdaLN Adapter}
\label{sec:adaln_adapter}

Each STDiT block of Orbis~v1 modulates three sub-layers $\ell \in \{\text{sa}, \text{smlp}, \text{tmlp}\}$ (spatial self-attention, spatial MLP, temporal MLP) via adaLN-Zero~\citep{peebles2023dit}. For each sub-layer, it produces modulation vectors $\boldsymbol{\gamma}^{k,\ell} \in \mathbb{R}^D$, $k \in \{\text{shift}, \text{scale}, \text{gate}\}$, which are shared by all $N$ tokens.
Our adapter augments this global signal with a per-token additive offset derived from the geo-reference, injected at every layer. 
%
%Our adapter adds a per-token offset derived from the geo-reference to this shared signal. 
The idea is related to SPADE~\citep{park2019spade}, but the offset is added to the adaLN signal rather than replacing it, so timestep information is preserved at every token.
\newline
Each block contains a lightweight MLP applied independently to every
geometry token $\mathbf{g}_{f,i}$,
\vspace{-0.05cm}
\begin{equation}
  \Delta(\mathbf{g}_{f,i}) = W_2\,\mathrm{SiLU}(W_1 \mathbf{g}_{f,i} + \mathbf{b}_1) + \mathbf{b}_2,
  \quad
  W_1 \in \mathbb{R}^{B_{\mathrm{neck}} \times C},\;
  W_2 \in \mathbb{R}^{9D \times B_{\mathrm{neck}}},
  \label{eq:adapter_mlp}
\end{equation}
where $B_{\mathrm{neck}}$ is the bottleneck width. The output is split into nine offsets $\boldsymbol{\delta}^{k,\ell}_{f,i} \in \mathbb{R}^{D}$, one for each modulation vector $k$ and sub-layer $\ell$, which are added to the base modulation:
\vspace{-0.05cm}
\begin{equation}
  \boldsymbol{\Gamma}^{k,\ell}_{f,i}
  = \boldsymbol{\gamma}^{k,\ell} + \boldsymbol{\delta}^{k,\ell}_{f,i},
  \qquad
  \boldsymbol{\delta}^{k,\ell}_{f,i} = \mathbf{0}
  \;\text{ for context frames},
  \label{eq:eff_mod}
\end{equation}
i.e.\ only the $F_{\text{pred}}$ predicted frames receive a geometric
offset, while context frames pass through the base modulation
unchanged. Sub-layer $\ell$, with function $\mathcal{F}_\ell$, then updates each token as
\vspace{-0.05cm}
\begin{equation}
  \mathbf{x}_{f,i} \leftarrow \mathbf{x}_{f,i}
  + \boldsymbol{\Gamma}^{\text{gate},\ell}_{f,i} \odot
    \mathcal{F}_\ell\!\Bigl(
      \mathrm{LN}(\mathbf{x})_{f,i} \odot \bigl(1 + \boldsymbol{\Gamma}^{\text{scale},\ell}_{f,i}\bigr)
      + \boldsymbol{\Gamma}^{\text{shift},\ell}_{f,i}
    \Bigr).
  \label{eq:eff_modulate}
\end{equation}
Causal temporal self-attention is unmodulated in the base block and remains so. Each of the $L$ blocks contains one adapter, and the final layer, which uses only a shift and scale receives an adapter with a $2D$-dimensional output.
The last layer of each adapter MLP is initialized with $\mathcal{N}(0, 0.02^2)$ weights and zero bias, so the offsets are small but non-zero at initialization.

\section{Geo-Reference Signal Quality}\label{app:geo_ref_quality}

Coverage of the projected geo-reference signal ranges from roughly 50\% to 78\% of each frame across the single-camera splits, with reprojection quality tracking SLAM trajectory accuracy far less than expected: nuScenes attains the best trajectory estimate (0.74\,m ATE) with a masked PSNR of 22.96\,dB, while MARS, despite the lowest trajectory error (0.42\,m) and highest coverage (68.5\%), yields by far the weakest masked PSNR (12.83\,dB), degrading further under cross-traversal reconstruction (MARS~(swm): 77.5\% coverage, 12.36\,dB). In the multi-view setting, the forward camera clearly outperforms the oblique left/right views (e.g., 21.83\,dB vs.\ 13.25/12.56\,dB masked PSNR on the nuPlan train split), pulling the multi-view mean well below the single-view front-camera numbers. Full results are in \Cref{tab:dataset-coverage}.

% Dataset coverage / PSNR summary — generated by dataset_coverage_table.py
\begin{table}[h]
\centering
\caption{Coverage and reprojection quality of the geo-reference signal.%
}
\label{tab:dataset-coverage}
\footnotesize
\setlength{\tabcolsep}{4pt}
\renewcommand{\arraystretch}{1.05}
\begin{tabular}{lllrcrrrr}
\toprule
 & & & & & & & \multicolumn{2}{c}{PSNR} \\
\cmidrule(lr){8-9}
View & Split & Dataset & Traj.\ RMSE & Scenes & Frames & Coverage & Full & Masked \\
\midrule
\multirow{6}{*}{Single}
  & Train & nuPlan & 3.25\,m & 1007 & 2{,}882{,}118 & 50.4\% & 8.88\,dB & 21.72\,dB \\
\cmidrule{2-9}
  & \multirow{4}{*}{Val} & nuPlan turns & 3.34\,m & 128 & 270{,}263 & 62.6\% & 9.88\,dB & 22.42\,dB \\
  & & Waymo & 2.70\,m & 182 & 35{,}854 & 52.9\% & 9.07\,dB & 24.43\,dB \\
  & & nuScenes & 0.74\,m & 150 & 34{,}948 & 55.5\% & 10.27\,dB & 22.96\,dB \\
  & & MARS & 0.42\,m & 401 & 88{,}621 & 68.5\% & 7.28\,dB & 12.83\,dB \\
\cmidrule{2-9}
  & Cross-trav. & MARS & — & 30 & 10{,}950 & 77.5\% & 7.81\,dB & 12.36\,dB \\
\midrule
\multirow{8}{*}{Multi}
  & \multirow{4}{*}{Train} & nuPlan (left) & 3.25\,m & 986 & 2{,}844{,}008 & 26.6\% & 7.60\,dB & 13.25\,dB \\
  & & nuPlan (front) & 3.25\,m & 986 & 2{,}844{,}008 & 50.5\% & 8.89\,dB & 21.83\,dB \\
  & & nuPlan (right) & 3.25\,m & 986 & 2{,}844{,}008 & 31.2\% & 7.77\,dB & 12.56\,dB \\
  & & nuPlan (mean) & 3.25\,m & 986 & 2{,}844{,}008 & 36.1\% & 8.09\,dB & 15.88\,dB \\
\cmidrule{2-9}
  & \multirow{4}{*}{Val} & nuPlan (left) & 3.24\,m & 225 & 505{,}139 & 35.7\% & 7.98\,dB & 13.20\,dB \\
  & & nuPlan (front) & 3.24\,m & 225 & 504{,}569 & 59.4\% & 9.56\,dB & 22.45\,dB \\
  & & nuPlan (right) & 3.24\,m & 225 & 505{,}139 & 37.3\% & 8.03\,dB & 12.82\,dB \\
  & & nuPlan (mean) & 3.24\,m & 225 & 504{,}949 & 44.1\% & 8.52\,dB & 16.16\,dB \\
\bottomrule
\end{tabular}
\end{table}

\begin{figure}[ht!]
    \centering
    \includegraphics[width=1\linewidth]{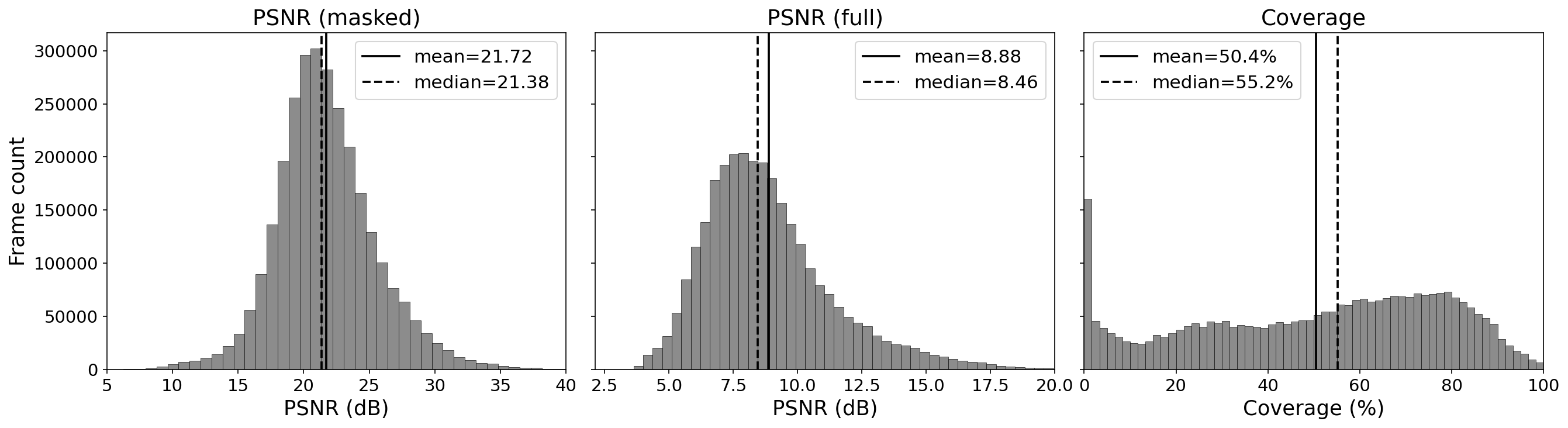}
    \caption{Distribution of geo-reference frame statistics over the training set, computed per frame: pixel coverage of the validity mask $\mathbf{M}$, the PSNR within masked region, and across the full geo-reference frame. Coverage varies substantially across frames, motivating conditioning mechanisms that tolerate partial and unreliable geometry rather than enforcing it uniformly.}
    \label{fig:geo_ref_train_statics}
\end{figure}

\begin{figure}[ht!]
    \centering
    \includegraphics[width=1\linewidth]{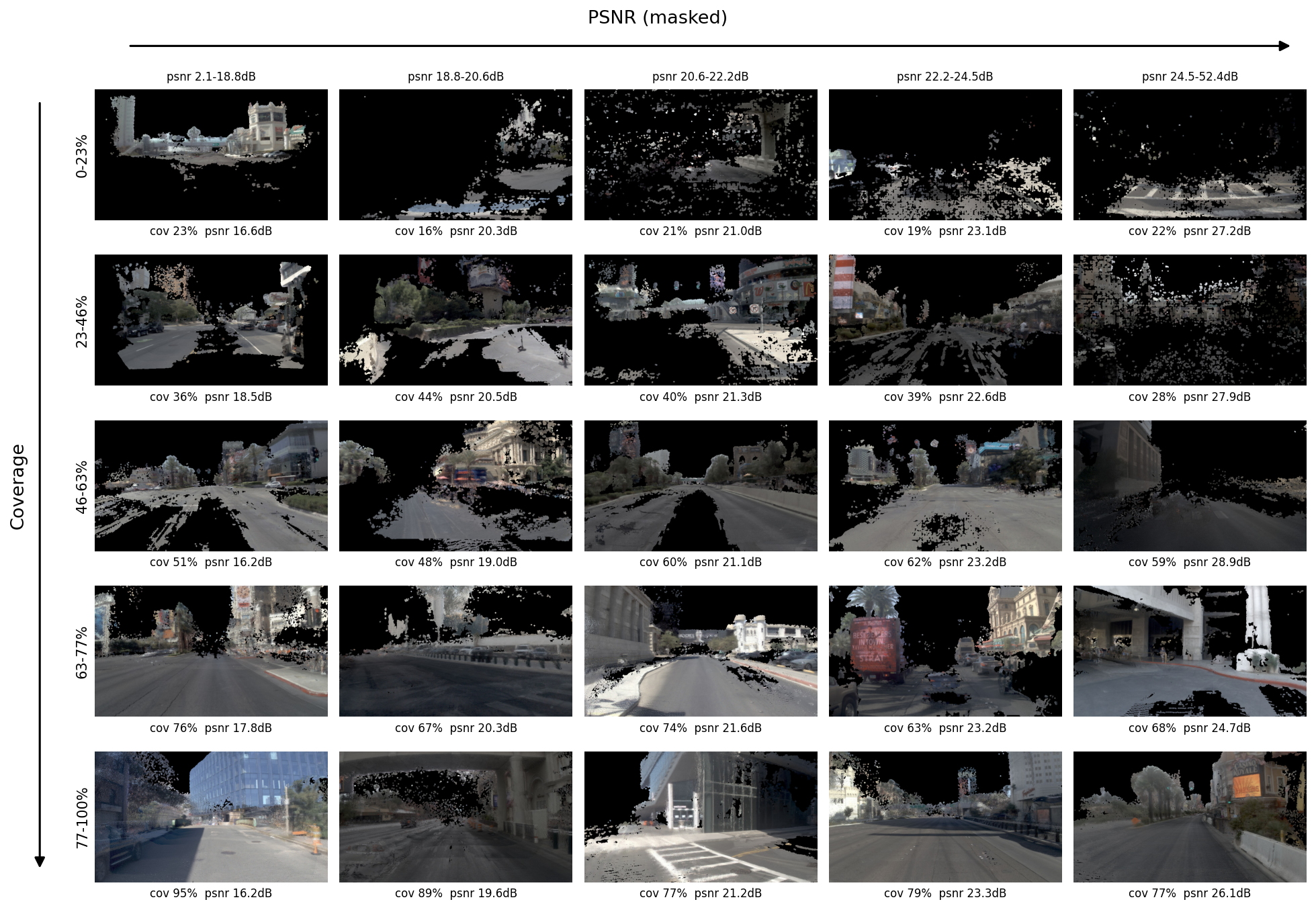}
    \caption{The geo-reference frames differ in coverage and quality. While some frames have very low coverage (e.g., due to being at the beginning of the collected trajectory), other frames provide a dense signal. The quality, measured via PSNR in the covered region, differs due to floaters or dynamic objects missing from the image. While dynamic objects are removed from the point cloud before rendering, parked cars remain in the scene.}
     \label{fig:geo_ref_psnr_coverage_grid}
\end{figure}

\begin{figure}[ht!]
    \centering
    \includegraphics[width=1\linewidth]{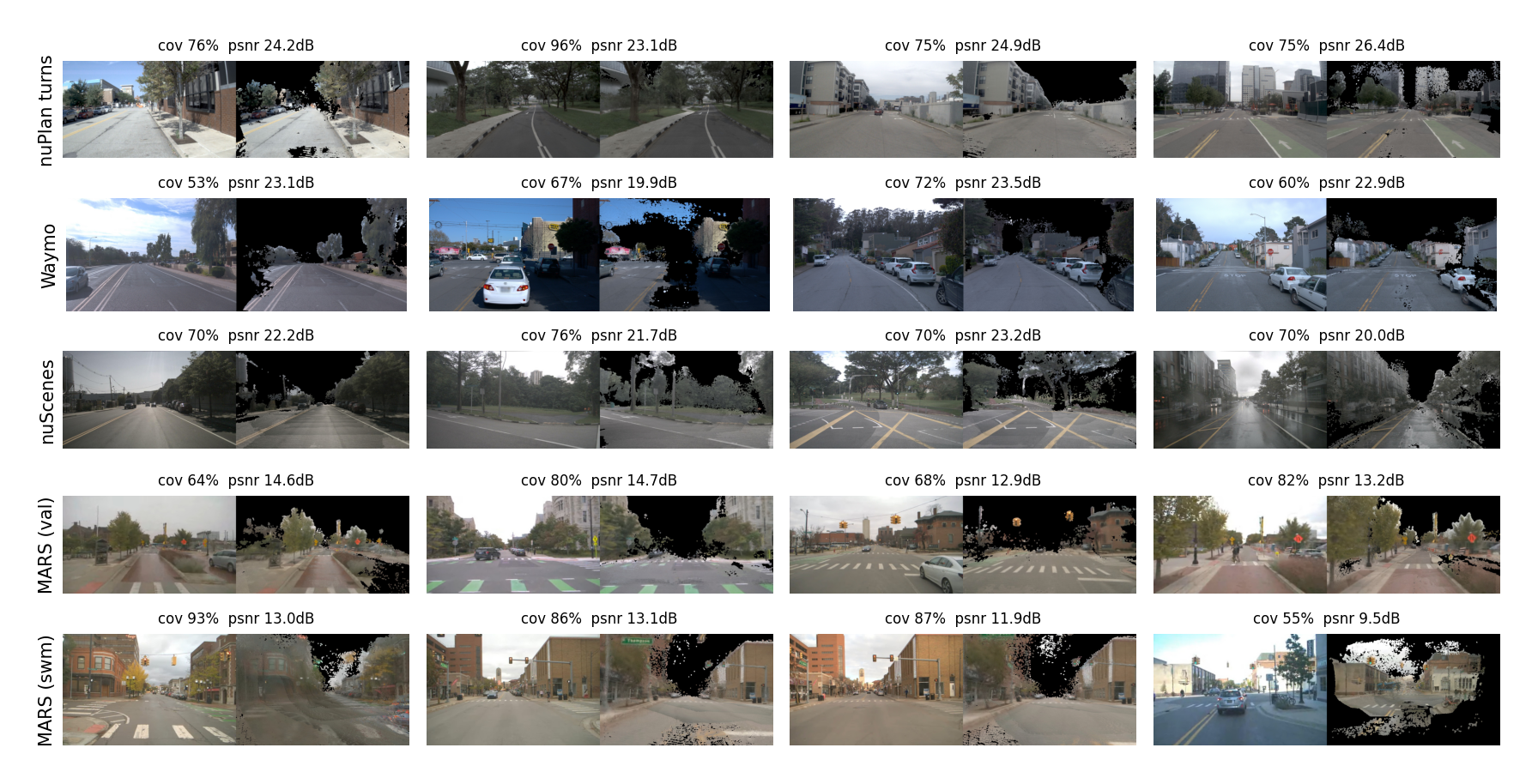}
    \caption{Example geo-reference validation frames from nuPlan turns, Waymo, nuScenes, our generated MARS validation set, and \textsc{Ann-Arbor-City-Bench} from 30 MARS cross-traversal pairs. }
     \label{fig:geo_ref_val}
\end{figure}

\clearpage
\newpage
\section{Hyperparameters}
\label{app:hp}

\subsubsection*{DROID-W SLAM}

\vspace{-0.25cm}
\begin{table}[ht!]
\centering
\caption{DROID-W SLAM hyperparameters.}
\label{tab:hp_droidw}
\begin{tabular}{lll}
\toprule
Parameter & Symbol & Value \\
\midrule
\multicolumn{3}{l}{\textit{Tracker}} \\
Metric3D input resolution & $H \times W$ & $616 \times 1064$ \\
Metric depth scale (raw prediction $\hat{d}$) & --- & $\hat{d} \cdot f_x / 1000$ \\
Metric depth clamp range & --- & $[0,\,300]$\,m \\
DINOv2-small patch stride & $s$ & 14 \\
\midrule
\multicolumn{3}{l}{\textit{Motion filter}} \\
Minimum flow to admit keyframe & $\tau_{\text{motion}}$ & 2.8\,px \\
\midrule
\multicolumn{3}{l}{\textit{Frontend (local BA)}} \\
Local BA iterations before keyframe decision & $k_1$ & 6 \\
Local BA iterations after keyframe decision & $k_2$ & 4 \\
Temporal edge window & $T_w$ & 25 frames \\
Keyframe pruning flow threshold & $\tau_{\text{kf}}$ & 3.5\,px \\
\midrule
\multicolumn{3}{l}{\textit{Backend (global BA)}} \\
Global BA trigger interval & $N_{\text{ba}}$ & 20 keyframes \\
Final BA coarse steps & $N_c$ & 12 \\
Final BA fine steps & $N_f$ & 20 \\
\bottomrule
\end{tabular}
\end{table}
\vspace{-0.25cm}

\subsubsection*{Point Cloud Rendering}

\vspace{-0.25cm}
\begin{table}[ht!]
\centering
\caption{Point-cloud filtering and rendering hyperparameters.}
\label{tab:hp_rendering}
\begin{tabular}{lll}
\toprule
Parameter & Symbol & Value \\
\midrule
\multicolumn{3}{l}{\textit{Point-cloud filters}} \\
Disparity gate (fraction of mean disparity) & $\tau_{\text{disp}}$ & $10^{-7}$ \\
Maximum projection depth & $d_{\text{max}}$ & 200\,m \\
Max.\ tracking uncertainty & $\tau_{\text{unc}}$ & 2.0 \\
Min.\ keyframes for multi-view survival & $N_{\text{mv}}$ & 5 \\
Max.\ relative depth error (multi-view) & $\tau_{\text{mv}}$ & 0.03 \\
Max.\ mean L1 reprojection color error & $\tau_{\text{photo}}$ & 0.20 \\
Min.\ frames for photometric survival & $N_{\text{photo}}$ & 2 \\
Voxel outlier min.\ neighbors & $k$ & 2 \\
Voxel outlier side length & $\ell$ & 0.08\,m \\
\midrule
\multicolumn{3}{l}{\textit{Weighted depth blending}} \\
Splat radius & $r$ & 1\,px \\
Hard depth gate threshold & $\delta$ & 2.0 \\
Soft depth decay rate ($\lambda{=}0$: hard gate) & $\lambda$ & 0 \\
\bottomrule
\end{tabular}
\end{table}
\vspace{-0.25cm}

\subsubsection*{Latent Blending}

\vspace{-0.25cm}
\begin{table}[ht!]
\centering
\caption{Latent blending hyperparameters.}
\label{tab:hp_anchor}
\begin{tabular}{lll}
\toprule
Parameter & Symbol & Value \\
\midrule
Flow-matching schedule minimum noise & $\sigma_{\min}$ & $10^{-5}$ \\
Number of ODE integration steps & $S$ & 30 \\
Sampler stochasticity ($\eta{=}0$: deterministic ODE) & $\eta$ & 0 \\
Fraction of steps with blending correction & $\rho_{\text{geo}}$ & 0.2 \\
Blend scale & $s_{\text{geo}}$ & 1.0 \\
\bottomrule
\end{tabular}
\end{table}

\subsubsection*{Cross-Attention and AdaLN Training}

\begin{table}[ht!]
\centering
\caption{Training hyperparameters shared by the adaLN adapter and the cross-attention adapter, and architecture constants specific to each.}
\label{tab:hp_adapters}
\begin{tabular}{llll}
\toprule
Parameter & Symbol & AdaLN & CrossAttn \\
\midrule
\multicolumn{4}{l}{\textit{Architecture}} \\
STDiT blocks & $L$ & 24 & 24 \\
Token grid & $H_{\text{lat}} \times W_{\text{lat}}$ & \multicolumn{2}{c}{$18 \times 32$} \\
Tokenizer latent channels & $C$ & \multicolumn{2}{c}{32} \\
Adapter MLP bottleneck width & $B_{\text{neck}}$ & 256 & --- \\
Adapter output layer init & --- & $\mathcal{N}(0,0.02^2)$, $\mathbf{b}_2{=}\mathbf{0}$ & --- \\
Cross-attention gate init & $\alpha$ & --- & 0 \\
Shared geo-reference projection & $W_{\text{proj}}$ & --- & $\mathbb{R}^{D \times C}$, shared across blocks \\
\midrule
\multicolumn{4}{l}{\textit{Training}} \\
Optimizer & --- & \multicolumn{2}{c}{AdamW} \\
Learning rate & $\mathrm{lr}$ & \multicolumn{2}{c}{$5 \times 10^{-5}$} \\
Weight decay & --- & \multicolumn{2}{c}{$0.01$} \\
LR warmup & --- & \multicolumn{2}{c}{5{,}000 steps (linear)} \\
Geo-ref dropout rate & --- & \multicolumn{2}{c}{10\%} \\
Pose perturbation radius (optional) & $R$ & \multicolumn{2}{c}{3\,m} \\
Yaw / pitch jitter (optional) & --- & \multicolumn{2}{c}{$\pm2^\circ$ / $\pm1^\circ$} \ \\
\bottomrule
\end{tabular}
\end{table}

\section{Evaluation Metrics}\label{app:metrics}

\subsection{Consistency metrics}

All consistency metrics compare a generated frame $G_t$ to its temporally aligned ground-truth frame $I_t$ and are averaged over all frames within the respective horizon.

\textbf{PSNR.} The peak signal-to-noise ratio measures pixel-wise reconstruction fidelity,
\[
\text{PSNR}(G_t, I_t)
= 10 \log_{10} \frac{\mathrm{MAX}^2}{\mathrm{MSE}(G_t, I_t)},
\]
where $\mathrm{MAX}$ is the maximum pixel intensity and $\mathrm{MSE}$ the mean squared error over all pixels and channels. PSNR rewards exact pixel alignment and therefore penalizes even sub-pixel spatial or sub-second temporal offsets between the generated and observed future.

\textbf{LPIPS.} The learned perceptual image patch
similarity~\citep{lpips} measures the distance between deep features of
a pre-trained network $\phi$,
\[
\text{LPIPS}(G_t, I_t)
= \sum_l \frac{1}{H_l W_l} \sum_{h,w}
  \bigl\| w_l \odot \bigl(\phi_l(G_t)_{hw} - \phi_l(I_t)_{hw}\bigr) \bigr\|_2^2,
\]
where $\phi_l$ denotes unit-normalized activations of layer $l$ and
$w_l$ learned per-channel weights. We use the VGG variant.
LPIPS correlates better with human judgments than pixel-wise metrics and
degrades more gracefully under small misalignments.

\textbf{mPSNR} and \textbf{mLPIPS}, PSNR and LPIPS~\citep{lpips} restricted to the static scene: dynamic agents (e.g. pedestrians and vehicle classes) are segmented with YOLOv8-seg in both frames and their union excluded, so the score reflects geometric fidelity rather than incidental disagreement in agent placement. We report this score to compare with the Seoul World Model paper \citep{seo2026grounding}.

\textbf{LightGlue keypoint matches.} We extract keypoints with SuperPoint and match them between the generated and ground-truth frames with LightGlue~\citep{lightglue}, reporting the number of matches with confidence above $0.5$. 
A high match count requires that the local scene structure is preserved at approximately corresponding locations, making this a structure-aware consistency measure that tolerates moderate appearance differences.

\textbf{Depth error.} To isolate geometric from photometric consistency, we predict monocular depth for both frames with DepthAnything~V3~\citep{depthanything3} and report the absolute relative error after per-frame median-scale alignment,
\[
\text{AbsRel}(G_t, I_t)
= \frac{1}{|\Omega|} \sum_{p \in \Omega}
  \frac{\bigl| s \, \hat{d}(p) - d(p) \bigr|}{d(p)},
\qquad
s = \frac{\operatorname{median}(d)}{\operatorname{median}(\hat{d})},
\]
where $\hat{d}$ and $d$ are the depth maps predicted from the generated and ground-truth frame, and $\Omega$ is the set of valid pixels.
The scale factor $s$ resolves the scale ambiguity of monocular depth.
We deliberately use an estimator that does not share a backbone with the Metric3D~\citep{metric3d} prior used in DROID-W, so that the conditioning signal and its evaluator remain independent.

\subsection{Perceptual quality metrics}

These metrics compare feature \emph{distributions} of generated and real frames and are computed over the full evaluation set within each horizon.

\textbf{Fréchet Inception Distance (FID).} FID~\citep{fid} fits Gaussians $\mathcal{N}(\boldsymbol{\mu}_g, \boldsymbol{\Sigma}_g)$ and $\mathcal{N}(\boldsymbol{\mu}_r, \boldsymbol{\Sigma}_r)$ to Inception-v3 features of generated and real frames and reports their Fréchet distance,
\[
\text{FID}
= \|\boldsymbol{\mu}_g - \boldsymbol{\mu}_r\|_2^2
+ \operatorname{Tr}\!\bigl(\boldsymbol{\Sigma}_g + \boldsymbol{\Sigma}_r
  - 2(\boldsymbol{\Sigma}_g \boldsymbol{\Sigma}_r)^{1/2}\bigr).
\]
It captures per-frame image realism but is blind to temporal artifacts such as flicker.

\textbf{Img.Q.}~\citep{huang2024vbench} is VBench's no-reference Imaging Quality score (MUSIQ-SPAQ~\citep{ke2021musiq}, averaged per frame then per video). We report this score to compare with the Seoul World Model paper \citep{seo2026grounding}.

\textbf{Fréchet Video Distance (FVD).} FVD~\citep{fvd} applies the same Fréchet formulation to spatio-temporal features of an I3D network computed on video clips rather than individual frames, and is therefore sensitive to temporal artifacts and motion realism in addition to per-frame quality.

\textbf{FVD slope.} Following Orbis~v2~\citep{orbis2_2026}, we split each rollout into $T$ consecutive 2\,s chunks (10 frames at 5\,Hz), compute FVD per chunk, and report $(\text{FVD}_T - \text{FVD}_1)/(T-1)$. Lower values indicate slower degradation over the rollout.

\subsection{Trajectory adherence metrics}

We recover the camera path of each generated rollout with VGGT~\citep{vggt}. As reference, we run VGGT on the corresponding ground-truth frames, so that both trajectories share VGGT's reconstruction errors, and align the generated path to the reference with a Sim(3) Umeyama fit. Since VGGT reconstructions are defined only up to scale, ADE is expressed in the units of the reference reconstruction rather than in meters.

\textbf{Average Displacement Error (ADE).} Given the aligned generated positions $\hat{\mathbf{p}}_1, \dots, \hat{\mathbf{p}}_T$ and the reference positions $\mathbf{p}_1, \dots, \mathbf{p}_T$,
\[
\text{ADE} = \frac{1}{T} \sum_{t=1}^T \|\hat{\mathbf{p}}_t - \mathbf{p}_t\|_2 .
\]

% TODO: add what we do for the counterfactuals!

% \textbf{Average Displacement Error (ADE).} Given a predicted trajectory $\hat{\tau} = (\hat{\mathbf{p}}_1, \dots, \hat{\mathbf{p}}_T)$ and a ground-truth trajectory $\tau = (\mathbf{p}_1, \dots, \mathbf{p}_T)$, with positions $\hat{\mathbf{p}}_t, \mathbf{p}_t \in \mathbb{R}^2$, the ADE is defined as the average Euclidean distance between corresponding points:
% \[
% \text{ADE}(\hat{\tau}, \tau) = \frac{1}{T} \sum_{t=1}^T \|\hat{\mathbf{p}}_t - \mathbf{p}_t\|_2.
% \]
This metric quantifies pointwise deviations and is sensitive to minor spatial misalignments.

\section{Additional Ablations}

\subsection{Geo-Reference Encoding}

We compare three encodings of the geo-reference frames $R_n$, all with gated cross-attention on Orbis~v1 (\Cref{tab:encoders}).

\paragraph{(i) Orbis tokenizer (default).} RGB renders are encoded with the frozen hybrid tokenizer of Orbis~v1~\citep{bai2024factorized} in its continuous mode. It concatenates a DINOv2-distilled~\citep{oquab2023dinov2} semantic branch and a detail branch, so the condition lies in the same latent space as the generated frames.

\paragraph{(ii) DINOv2.} RGB renders are encoded with a frozen DINOv2 encoder~\citep{oquab2023dinov2}. Its features serve directly as condition tokens, providing a purely semantic representation without a reconstruction objective. Frames encoded this way are never compressed unlike the ones encoded via the Orbis tokenizer. 

\paragraph{(iii) Depth + Canny.} Depth maps rendered from the filtered point cloud are replicated to three channels and encoded with the Orbis tokenizer, together with Canny edge maps of the RGB renders. % TODO: how are depth and edge tokens combined?
This signal carries geometry and edges but no color.

% Encoder ablation — generated by make_results.py
\begin{table}[ht!]
\centering
\caption{%
  Geo-reference encoding ablation on \textsc{nuPlan-Turns} (Orbis~v1, cross-attention): RGB renders encoded with the Orbis tokenizer or a frozen DINOv2 encoder, and depth and Canny-edge renders encoded with the Orbis tokenizer. Metrics over frames 0--49 (short) and 50--99 (long) at 5\,Hz. Best per column in \textbf{bold}, second best \underline{underlined}.
}
\label{tab:encoders}
\footnotesize
\setlength{\tabcolsep}{3pt}
\renewcommand{\arraystretch}{1.05}
\begin{tabular}{l|cc|cc|cc|cc|cc}
\toprule
& \multicolumn{2}{c|}{LPIPS $\downarrow$} & \multicolumn{2}{c|}{FVD $\downarrow$} & \multicolumn{2}{c|}{KP Matches $\uparrow$} & \multicolumn{2}{c|}{Depth AbsRel $\downarrow$} & \multicolumn{2}{c}{ADE $(\times10^{-3})$ $\downarrow$} \\
Encoder & Short & Long & Short & Long & Short & Long & Short & Long & Short & Long \\
\midrule
  Orbis tokenizer & \textbf{0.393} & \textbf{0.437} & \textbf{128.5} & \underline{188.4} & \textbf{225.1} & \textbf{148.4} & \textbf{0.167} & \underline{0.206} & \textbf{3.6} & \textbf{9.9} \\
  DINOv2 & \underline{0.405} & \underline{0.443} & \underline{139.3} & \textbf{184.8} & \underline{193.0} & \underline{121.0} & \underline{0.170} & \textbf{0.200} & \underline{3.8} & \underline{10.4} \\
  Depth+Canny & 0.429 & 0.503 & 166.0 & 297.6 & 172.4 & 81.0 & 0.189 & 0.254 & 3.9 & 10.9 \\
\bottomrule
\end{tabular}
\end{table}

\subsection{Multi-Camera Conditioning.}
A natural extension is synchronized multi-camera generation with the front, left, and right cameras. We add another cross-attention block performing epipolar band cross-attention between nearby frames. We train this model with and without geo-reference conditioning. The shared point cloud is rendered separately for each camera, using the rig extrinsics for the side views, and each camera has its own cross-attention conditioning module.
Geo-reference conditioning improves every metric in every view (\Cref{tab:multiview}). The front view benefits most: long-horizon KP matches triple (60.1\,$\to$\,180.0) and long-horizon FVD drops by 42\% (360.7\,$\to$\,209.6), consistent with the front camera covering the most geometry-rich part of the reconstruction. For the side cameras, long-horizon KP matches roughly double, with smaller absolute gains, as the side views cover sparser regions of the reconstruction. Averaged over views, cross-attention lowers long-horizon FVD from 272.1 to 191.2 and raises long-horizon KP matches from 49.4 to 117.2, confirming that the shared geometric anchor enforces structural consistency across the full rig.
%
% Multiview rollout — generated by make_results.py
\begin{table}[ht!]
\centering
\caption{%
  Multi-view rollouts on \textsc{nuPlan-Turns} for the left, front, and right cameras, without (Base) and with cross-attention conditioning. \emph{Average} is the mean over the three views. Metrics over frames 0--49 (short) and 50--99 (long) at 5\,Hz. Cross-attention improves every metric in every view, best average values in \textbf{bold}.
}
\label{tab:multiview}
\footnotesize
\setlength{\tabcolsep}{3pt}
\renewcommand{\arraystretch}{1.05}
\begin{tabular}{ll|cc|cc|cc|cc|cc}
\toprule
& & \multicolumn{2}{c|}{LPIPS $\downarrow$} & \multicolumn{2}{c|}{FVD $\downarrow$} & \multicolumn{2}{c|}{KP Matches $\uparrow$} & \multicolumn{2}{c|}{Depth AbsRel $\downarrow$} & \multicolumn{2}{c}{ADE $(\times10^{-3})$ $\downarrow$} \\
Variant & View & Short & Long & Short & Long & Short & Long & Short & Long & Short & Long \\
\midrule
\multirow{4}{*}{Base}
  & Left & 0.478 & 0.556 & 145.1 & 210.7 & 130.8 & 47.2 & 0.231 & 0.303 & 13.3 & 28.5 \\
  & Front & 0.448 & 0.541 & 185.3 & 360.7 & 171.1 & 60.1 & 0.214 & 0.314 & 4.9 & 14.7 \\
  & Right & 0.487 & 0.564 & 156.9 & 245.0 & 123.3 & 40.9 & 0.231 & 0.297 & 10.3 & 26.1 \\
  & Average & 0.471 & 0.554 & 162.4 & 272.1 & 141.7 & 49.4 & 0.225 & 0.305 & 9.5 & 23.1 \\
\midrule
\multirow{4}{*}{Cross-attn}
  & Left & 0.432 & 0.499 & 126.2 & 175.9 & 165.8 & 87.1 & 0.194 & 0.245 & 12.3 & 27.2 \\
  & Front & 0.337 & 0.416 & 135.4 & 209.6 & 265.0 & 180.0 & 0.144 & 0.208 & 4.2 & 12.7 \\
  & Right & 0.436 & 0.506 & 130.6 & 188.0 & 165.2 & 84.5 & 0.186 & 0.244 & 9.8 & 24.8 \\
  & Average & \textbf{0.402} & \textbf{0.474} & \textbf{130.7} & \textbf{191.2} & \textbf{198.7} & \textbf{117.2} & \textbf{0.175} & \textbf{0.233} & \textbf{8.7} & \textbf{21.6} \\
\bottomrule
\end{tabular}
\end{table}

\newpage
\section{Extended Results}

\subsection{Long-Horizon Rollout Performance Per Dataset}

% FVD of the GeoRef Signal is uninformative since the 

In \Cref{tab:main_datasets} we report the values obtained on nuPlan turns, the longest sequences from the Waymo validation set, the MARS dataset, and on nuScenes validation. 

% Comparison against baselines — generated by make_results.py
\vspace{-0.5cm}
\begin{table}[ht!]
\centering
\caption{%
  Comparison against baselines on \textsc{nuPlan-Turns}, Waymo, \textsc{MARS}, and nuScenes. Short ($s$) / long ($l$) horizons: frames 0--39 / 40--79 (8\,s each, 16\,s total) for every dataset (both at 5\,Hz). ADE in $\times10^{-3}$\,m. Best per column and dataset in \textbf{bold}, second best \underline{underlined}. FVD slope over $T{=}8$ chunks (frames 0--79), see \Cref{app:metrics}.%
}
\label{tab:main_datasets}
\footnotesize
\setlength{\tabcolsep}{3pt}
\renewcommand{\arraystretch}{1.05}
\begin{tabular}{ll|cc|ccc|cc|cc|cc}
\toprule
& & \multicolumn{2}{c|}{LPIPS $\downarrow$} & \multicolumn{3}{c|}{FVD $\downarrow$} & \multicolumn{2}{c|}{KP Matches $\uparrow$} & \multicolumn{2}{c|}{Depth AbsRel $\downarrow$} & \multicolumn{2}{c}{ADE $\downarrow$} \\
 & Method & Short & Long & Short & Long & Slope & Short & Long & Short & Long & Short & Long \\
\midrule
\multirow{7}{*}{\rotatebox{90}{\textsc{nuPlan-T.}}}
  & Orbis~v1 & 0.451 & 0.545 & 249.9 & 440.4 & 56.3 & 181.1 & 68.4 & 0.217 & 0.305 & 5.0 & 13.9 \\
  & Orbis~v2 & 0.440 & 0.536 & 222.3 & 333.8 & 33.9 & 178.7 & 63.6 & 0.218 & 0.310 & 4.7 & 12.1 \\
  & Cosmos 2.5 & 0.483 & 0.548 & 297.8 & 444.9 & 39.3 & 241.5 & 85.4 & 0.232 & 0.312 & 6.7 & 11.3 \\
\cmidrule{2-13}
  & GeoRef Signal & 0.451 & 0.438 & 1325.1 & 1291.3 & -- & 212.9 & 226.3 & 0.352 & 0.318 & 4.8 & 8.4 \\
  & Ours (Orbis~v1) & 0.387 & 0.428 & \textbf{120.0} & \textbf{174.1} & \textbf{13.7} & 237.7 & 159.7 & 0.162 & 0.197 & \underline{3.0} & \underline{8.2} \\
  & Ours (Orbis~v2) & \textbf{0.304} & \textbf{0.360} & 131.9 & \underline{183.3} & \underline{16.2} & \underline{298.1} & \underline{249.4} & \textbf{0.104} & \textbf{0.132} & 3.4 & 9.1 \\
  & Ours (Cosmos 2.5) & \underline{0.321} & \underline{0.392} & \underline{128.0} & 237.7 & 22.7 & \textbf{387.0} & \textbf{290.1} & \underline{0.122} & \underline{0.162} & \textbf{2.7} & \textbf{7.2} \\
\midrule
\multirow{7}{*}{\rotatebox{90}{Waymo}}
  & Orbis~v1 & 0.371 & 0.469 & 353.5 & 579.7 & 67.2 & 237.6 & 122.3 & 0.184 & 0.280 & 3.9 & 13.7 \\
  & Orbis~v2 & 0.336 & 0.446 & 293.4 & 440.4 & 49.2 & 243.9 & 122.8 & 0.167 & 0.267 & 3.4 & 12.1 \\
  & Cosmos 2.5 & 0.382 & 0.478 & 347.5 & 496.0 & 46.3 & \underline{342.9} & 171.1 & 0.177 & 0.275 & 4.6 & 10.1 \\
\cmidrule{2-13}
  & GeoRef Signal & 0.541 & 0.533 & 2118.0 & 1686.8 & -- & 164.0 & 171.7 & 0.311 & 0.350 & 3.0 & 5.2 \\
  & Ours (Orbis~v1) & \underline{0.313} & \underline{0.388} & 276.2 & \underline{359.9} & \textbf{25.6} & 286.6 & 208.6 & 0.149 & 0.216 & \textbf{1.6} & \textbf{4.4} \\
  & Ours (Orbis~v2) & \textbf{0.267} & \textbf{0.341} & \textbf{249.8} & \textbf{346.2} & \underline{30.7} & 312.8 & \underline{251.2} & \textbf{0.116} & \textbf{0.169} & \underline{1.9} & \underline{4.5} \\
  & Ours (Cosmos 2.5) & 0.314 & 0.398 & \underline{257.7} & 415.4 & 38.9 & \textbf{424.1} & \textbf{306.9} & \underline{0.121} & \underline{0.171} & 2.0 & 6.5 \\
\midrule
\multirow{7}{*}{\rotatebox{90}{\textsc{MARS}}}
  & Orbis~v1 & 0.453 & 0.531 & 311.1 & 528.5 & 64.3 & 216.5 & 116.6 & 0.247 & 0.336 & 4.8 & 11.0 \\
  & Orbis~v2 & 0.412 & 0.512 & 285.3 & 536.6 & 57.2 & 218.5 & 104.5 & 0.223 & 0.333 & 3.3 & 7.3 \\
  & Cosmos 2.5 & 0.477 & 0.550 & 323.8 & 526.9 & 45.1 & 266.2 & 119.0 & 0.256 & 0.352 & 5.4 & 6.9 \\
\cmidrule{2-13}
  & GeoRef Signal & 0.497 & 0.477 & 2878.3 & 2480.9 & -- & 212.9 & \underline{223.6} & 0.328 & 0.334 & 3.5 & 4.1 \\
  & Ours (Orbis~v1) & \textbf{0.315} & \textbf{0.394} & \textbf{207.4} & \textbf{365.1} & \underline{39.5} & \textbf{287.5} & 216.7 & \textbf{0.145} & \underline{0.213} & \textbf{1.9} & \textbf{3.4} \\
  & Ours (Orbis~v2) & \underline{0.366} & \underline{0.430} & \underline{247.4} & \underline{404.1} & \textbf{38.1} & \underline{272.6} & \textbf{224.3} & \textbf{0.145} & \textbf{0.200} & \underline{2.5} & \underline{3.9} \\
  & Ours (Cosmos 2.5) & 0.432 & 0.471 & 479.3 & 712.8 & 69.3 & 271.6 & 210.1 & \underline{0.204} & 0.253 & 9.4 & 8.5 \\
\midrule
\multirow{7}{*}{\rotatebox{90}{nuScenes}}
  & Orbis~v1 & 0.392 & 0.486 & 380.9 & 609.1 & 68.2 & 233.9 & 112.0 & 0.207 & 0.306 & 5.4 & 20.1 \\
  & Orbis~v2 & 0.384 & 0.480 & 330.8 & 476.4 & 43.1 & 232.1 & 102.7 & 0.200 & 0.289 & 4.8 & 14.0 \\
  & Cosmos 2.5 & 0.391 & 0.478 & 377.4 & 547.6 & 43.5 & \underline{352.0} & 170.3 & 0.195 & 0.278 & 6.2 & 14.6 \\
\cmidrule{2-13}
  & GeoRef Signal & 0.537 & 0.524 & 1858.1 & 1562.4 & -- & 164.5 & 178.9 & 0.372 & 0.367 & 4.3 & 10.8 \\
  & Ours (Orbis~v1) & 0.364 & 0.434 & 304.9 & \textbf{400.9} & \textbf{30.1} & 259.3 & 152.9 & 0.173 & 0.241 & \textbf{2.7} & \underline{10.0} \\
  & Ours (Orbis~v2) & \underline{0.314} & \textbf{0.387} & \underline{290.3} & \underline{414.3} & \underline{36.7} & 300.9 & \underline{219.1} & \underline{0.137} & \underline{0.185} & \textbf{2.7} & \textbf{9.1} \\
  & Ours (Cosmos 2.5) & \textbf{0.296} & \underline{0.388} & \textbf{275.0} & 454.5 & 38.2 & \textbf{441.4} & \textbf{307.6} & \textbf{0.124} & \textbf{0.172} & \underline{2.8} & 10.5 \\
\bottomrule
\end{tabular}
\end{table}

\subsection{Coverage vs.\ Quality.} 

\Cref{app:tab_ablation-coverage} varies the coverage of the geo-reference signal via the multi-view depth threshold $\tau_\text{mv}$. Denser signals improve the rollout, but the additional points are less reliable: PSNR over covered pixels drops from 22.8\,dB at 34\% coverage to 21.9\,dB at 79--87\%, and the looser filtering admits floaters and artifacts (\Cref{fig:coverage}). Our conservative coverage score of 62\% is a trade-off between signal quality and coverage.

\begin{table}[ht!]
\centering
\caption{%
    Geo-reference coverage ablation on \textsc{nuPlan-Turns} (Orbis~v1, cross-attention). Coverage is varied via the multi-view depth threshold $\tau_\text{mv}$. \emph{Oracle}: shows the model being conditioned on ground truth images, also containing dynamic objects. Metrics over frames 0--49 (short) and 50--99 (long) at 5\,Hz. Best per column in \textbf{bold}, second best \underline{underlined}; Oracle is excluded from the ranking.
}
\label{app:tab_ablation-coverage}
\footnotesize
\setlength{\tabcolsep}{3pt}
\renewcommand{\arraystretch}{1.05}
\resizebox{\textwidth}{!}{%
\begin{tabular}{l|c|cc|cc|cc|cc|cc}
\toprule
& PSNR (covered) & \multicolumn{2}{c|}{LPIPS $\downarrow$} & \multicolumn{2}{c|}{FVD $\downarrow$} & \multicolumn{2}{c|}{KP Matches $\uparrow$} & \multicolumn{2}{c|}{Depth AbsRel $\downarrow$} & \multicolumn{2}{c}{ADE $(\times10^{-3})$ $\downarrow$} \\
Coverage & (dB) $\uparrow$ & Short & Long & Short & Long & Short & Long & Short & Long & Short & Long \\
\midrule
  6\% & 22.49 & 0.446 & 0.523 & 182.3 & 360.0 & 161.3 & 69.4 & 0.216 & 0.300 & 4.2 & 11.8 \\
  24\% & \underline{22.74} & 0.427 & 0.498 & 165.0 & 297.5 & 175.0 & 82.6 & 0.192 & 0.267 & 3.7 & 10.6 \\
  34\% & \textbf{22.79} & 0.418 & 0.484 & 159.1 & 264.4 & 184.8 & 93.3 & 0.185 & 0.252 & 3.7 & 10.4 \\
  54\% & 22.55 & 0.401 & 0.451 & 137.8 & 208.1 & 210.9 & 126.5 & 0.172 & 0.221 & \underline{3.6} & 10.1 \\
  62\% (Ours) & 22.42 & 0.393 & 0.437 & 128.5 & 188.4 & 225.1 & 148.4 & 0.167 & 0.206 & \underline{3.6} & \underline{9.9} \\
  79\% & 21.88 & \underline{0.383} & \textbf{0.419} & \textbf{123.0} & \textbf{171.5} & \underline{247.3} & \textbf{177.3} & \underline{0.157} & \textbf{0.193} & \underline{3.6} & 10.1 \\
  87\% & 21.86 & \textbf{0.383} & \underline{0.421} & \underline{125.6} & \underline{180.1} & \textbf{248.8} & \underline{175.4} & \textbf{0.157} & \underline{0.193} & \textbf{3.4} & \textbf{9.5} \\
\midrule
  Oracle & -- & 0.353 & 0.361 & 109.0 & 138.8 & 322.8 & 292.9 & 0.138 & 0.149 & 3.0 & 8.0 \\
\bottomrule
\end{tabular}
}
\end{table}

\subsection{Robustness to Localization Error}
\Cref{app:tab_alignment} reports the full set of metrics for the localization-error experiment of \Cref{sec:localization} on Orbis~v1. Each model is evaluated on both aligned and misaligned geo-reference renders, so that the cost of training with pose perturbations and the cost of encountering a displaced signal at inference can be read separately. The unconditioned backbone is included as the score a model would reach by ignoring the signal entirely.

% GeoRef alignment — generated by make_results.py
\begin{table}[ht!]
\centering
\caption{%
  The nominal reprojection (\emph{aligned eval}) and one rendered from a virtual viewpoint displaced onto a $3$\,m circle (\emph{misaligned eval}). \emph{Orbis~v1} is the unconditioned backbone, which takes no geo-ref signal at all — it is what a run would score if it had learned to ignore a signal it cannot trust. Metrics averaged over frames 0--49 ($s$) and 50--99 ($l$) at 5\,Hz. ADE in $\times10^{-3}$\,m.%
}
\label{app:tab_alignment}
\footnotesize
\setlength{\tabcolsep}{3pt}
\renewcommand{\arraystretch}{1.05}
\resizebox{\textwidth}{!}{%
    \begin{tabular}{ll|cc|cc|cc|cc|cc}
    \toprule
    & & \multicolumn{2}{c|}{LPIPS $\downarrow$} & \multicolumn{2}{c|}{FVD $\downarrow$} & \multicolumn{2}{c|}{KP Matches $\uparrow$} & \multicolumn{2}{c|}{Depth AbsRel $\downarrow$} & \multicolumn{2}{c}{ADE $(\times10^{-3})$ $\downarrow$} \\
    Training & Evaluation & Short & Long & Short & Long & Short & Long & Short & Long & Short & Long \\
    \midrule
    \multirow{1}{*}{}
      & Orbis~v1 & 0.468 & 0.554 & 276.1 & 501.3 & 160.7 & 61.4 & 0.232 & 0.316 & 6.0 & 16.9 \\
    \midrule
    \multirow{2}{*}{Pose error 0m}
      & Pose error 0m & 0.393 & 0.437 & 128.5 & 188.4 & 225.1 & 148.4 & 0.167 & 0.206 & 3.6 & 9.9 \\
      & Pose error 0--3m & 0.477 & 0.500 & 502.5 & 728.4 & 126.6 & 84.3 & 0.234 & 0.254 & 19.1 & 28.0 \\
    \midrule
    \multirow{2}{*}{Pose error 0--3m}
      & Pose error 0m & 0.456 & 0.522 & 175.4 & 285.3 & 158.2 & 65.6 & 0.221 & 0.290 & 4.3 & 11.5 \\
      & Pose error 0--3m & 0.458 & 0.525 & 177.4 & 299.3 & 156.7 & 65.9 & 0.223 & 0.297 & 4.2 & 12.1 \\
    \bottomrule
    \end{tabular}
}
\end{table}

\subsection{Scene Dynamics}\label{sec:dynamic_vehicles}
\begin{figure}
    \centering
    \includegraphics[width=1\linewidth]{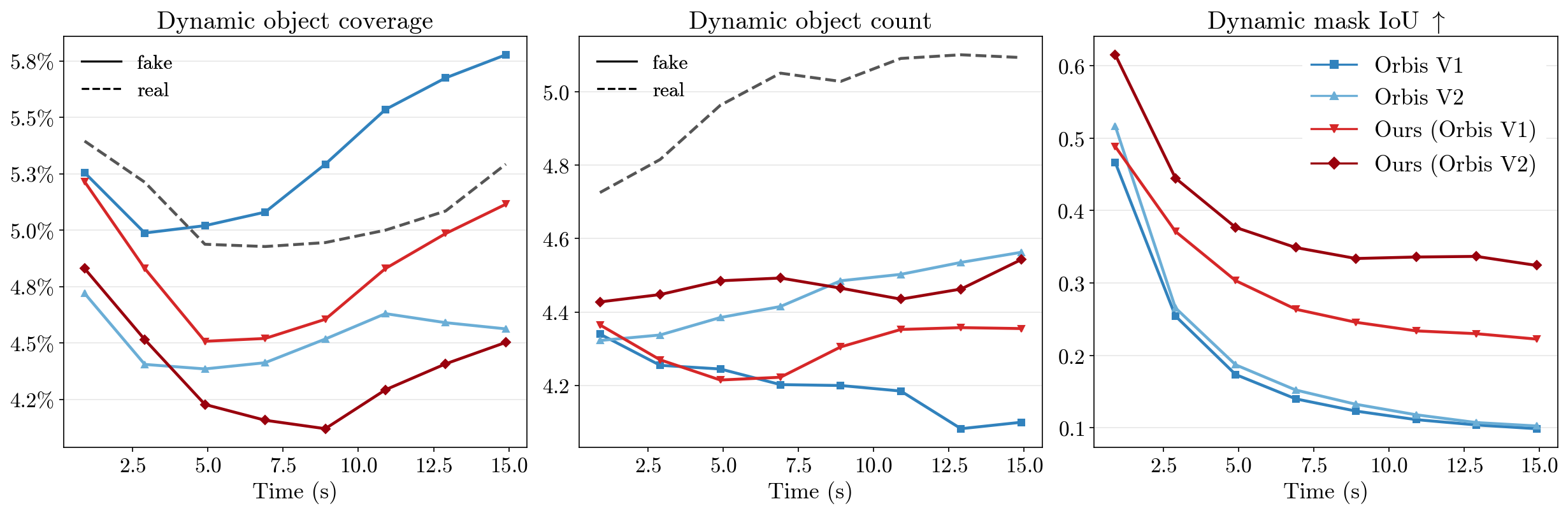}
    \caption{Dynamic object count over the rollout, comparing the base world models with their anchored variants. Values are the difference between the generated and the recorded frames, where zero is ideal.}
    \label{fig:dynamics}
\end{figure}

\Cref{fig:dynamics} tracks the number of dynamic objects over the rollout, measured as the difference between the generated and the recorded frames. The anchored models stay within the range of their unconditioned counterparts throughout, so conditioning on static geometry does not remove moving agents as the rollout proceeds.

\subsection{Counterfactual Evaluation}

Throughout, \emph{GT} denotes the real recorded driving log (real camera
frames, real ego odometry), the same log droidw reconstructed the persistent
geometry from. \emph{Nominal rollout} and \emph{counterfactual rollout} both
denote videos \emph{generated} by our model -- one conditioned on the geo-ref
render along the unperturbed (replay) trajectory, the other along the
perturbed (lateral/wiggle/speed) trajectory. Neither is ground truth; GT
appears only as the fixed reference both are measured against.

Measuring compliance as the raw displacement between the counterfactual and
nominal rollout is not enough: VGGT's own pose-estimation noise can push the
reconstructed trajectory in an arbitrary direction, and this reconstruction
error is easily mistaken for a genuine lateral shift.

A more fundamental problem is that ADE cannot see lateral compliance at all.
Umeyama's similarity alignment jointly fits scale, rotation, and translation,
which removes a pure translation exactly; since each rollout is additionally
re-origined at its own first generated frame, any lateral offset -- which
after droidw's ramp is a fixed offset from the nominal path -- is fit away
entirely.

To recover a genuine compliance signal we anchor every rollout to real,
shared geometry instead of aligning it to itself:

\begin{enumerate}
\item Run VGGT jointly over the 3 real conditioning frames and the 100
  generated frames, for the nominal and counterfactual rollout separately.
  Since the conditioning frames are real and identical across both, VGGT's
  own canonicalization to its first input frame fixes a shared origin and
  heading for both reconstructions.
\item Fit a single scalar scale $s^*$ per rollout on the forward
  (along-track) component only, by least squares against GT odometry:
  \begin{equation}
    s^* = \frac{\sum_t f(t)\, f^{\text{gt}}(t)}{\sum_t f(t)^2},
  \end{equation}
  where $f(t)$ is VGGT's forward coordinate for that rollout at frame $t$ and
  $f^{\text{gt}}(t)$ is the corresponding real GT forward distance. Because
  only the along-track component enters this fit, $s^*$ cannot absorb a
  sideways shift.
\item Gate out sequences where VGGT disagrees with GT motion, using the
  \emph{nominal} rollout's own fit quality only (outcome-independent, since
  it does not depend on whether the counterfactual complied):
  \begin{equation}
    s^* > 0 \quad\text{and}\quad
    R^2 = 1 - \frac{\sum_t \left(s^* f(t) - f^{\text{gt}}(t)\right)^2}
                   {\sum_t \left(f^{\text{gt}}(t) - \overline{f^{\text{gt}}}\right)^2}
    \;\geq\; 0.9.
  \end{equation}
\item Measure the signed cross-track distance of each rollout's scaled VGGT
  trajectory from the real GT path, separately for the counterfactual and
  the nominal rollout:
  \begin{equation}
    d_{\text{cf}}(t) = s^*_{\text{cf}}\, g_{\text{cf}}(t),
    \qquad
    d_{\text{nom}}(t) = s^*_{\text{nom}}\, g_{\text{nom}}(t),
  \end{equation}
  where $g(t)$ is VGGT's lateral coordinate at frame $t$ for that rollout.
  Both are measured against the same real GT path, not against each other,
  so any lateral bias intrinsic to VGGT's reconstruction of that scene --
  present regardless of which trajectory was commanded -- appears in both.
\item Take the paired difference
  \begin{equation}
    \Delta(t) = d_{\text{cf}}(t) - d_{\text{nom}}(t),
  \end{equation}
  which cancels shared VGGT reconstruction bias between the two generated
  rollouts and isolates the shift induced specifically by the counterfactual
  command.
\end{enumerate}

\paragraph{Wiggle.} The wiggle perturbation is a sine-wave lateral offset
(period 10\,m with a maximal offset of 0.5 meters), so its commanded shift alternates sign along the rollout.
Reporting a single scalar $\Delta$ (as for lateral) is meaningless here: the
positive and negative half-cycles average toward zero by construction, even
under full compliance. Instead, for each sample $i$ we use droidw's own
per-frame commanded offset trace $e_i(t)$ -- the actual lateral component of
the perturbed trajectory at frame $t$, which already accounts for how the
ramp and the vehicle's heading through turns modulate the raw sine amplitude
-- and fit a single regression slope through the origin between the measured
paired shift and the commanded trace:
\begin{equation}
  \beta_i = \frac{\sum_t \Delta_i(t)\, e_i(t)}{\sum_t e_i(t)^2}.
\end{equation}
$\beta_i = 1$ means the measured shift tracks the commanded sine curve
exactly (full compliance); $\beta_i = 0$ means no relationship. We report the
median $\beta_i$ across samples, together with the fraction of samples with
$\beta_i > 0$ and its binomial $p$-value.

\paragraph{Speed.} A speed change is not a translation: it changes how far
along the route -- through the turn -- the rollout has progressed by a given
frame, which is a change in the trajectory's own shape that Umeyama's
similarity alignment cannot fit away. Plain ADE is therefore usable for
speed, provided it is measured against the correctly speed-scaled target
rather than the nominal-speed one. For each commanded multiplier
$\mu \in \{0.85, 1.15\}$ we build a target $\text{gtcf}_\mu(t)$ by
resampling real GT odometry along its own arclength at rate $\mu$ (the same
route, traversed $\mu\times$ faster or slower), and compute per-sample ADE
(VGGT trajectory re-origined at frame 0, aligned with $\mathrm{umeyama\_2d}$)
of both the counterfactual and the nominal rollout against this \emph{same}
target:
\begin{equation}
  a_{\text{own}} = \mathrm{ADE}\!\left(V_{\text{cf}},\, \text{gtcf}_\mu\right),
  \qquad
  a_{\text{nom}} = \mathrm{ADE}\!\left(V_{\text{nom}},\, \text{gtcf}_\mu\right).
\end{equation}
Compliance is the paired comparison $a_{\text{own}} < a_{\text{nom}}$
(fraction of samples, binomial $p$-value): a rollout that actually changed
speed should track the $\mu$-scaled target more closely than one that did
not. We additionally check specificity against the \emph{opposite}
multiplier's target $\text{gtcf}_{\mu'}$, confirming
$a_{\text{own}} < a_{\text{opp}} = \mathrm{ADE}(V_{\text{cf}}, \text{gtcf}_{\mu'})$.

% Counterfactual trajectories -- generated by build_table.py
% (fidelity columns from make_results.py's table_ablation_counterfactuals.tex,
%  compliance column from lateral_offset.py / speed_ade.py -- see README.md)
\begin{table}[ht!]
\centering
\caption{%
  Counterfactual trajectory perturbations of \textsc{Ours} on \textsc{nuPlan-Turns}.
  FID/FVD/ADE are computed against the \emph{nominal} ground-truth rollout and
  quantify how far the generation drifts from the unperturbed future (lower is
  not better here -- nominal is the reference, not a target). ADE cannot see
  lateral compliance at all (see text), so Compliance is a separate
  measurement, in different units per row: for Lateral, the median
  counterfactual-minus-nominal sideways shift in metres, sign matching the
  commanded direction (samples passing a nominal-only VGGT quality gate; see
  fraction of samples in that direction and its $p$-value); for Wiggle, whose
  shift alternates sign and so cannot be summarized as a single offset, the
  per-sample regression slope of measured shift on the commanded sine curve
  (1 = fully follows, 0 = ignores); for Speed, the fraction of samples closer
  to their GT-odometry speed target than the nominal rollout is. ADE in
  $\times10^{-3}$\,m.%
}
\label{tab:counterfactual-compliance}
\footnotesize
\setlength{\tabcolsep}{3pt}
\renewcommand{\arraystretch}{1.05}

\begin{tabular}{l|cc|cc|cc|c}
\toprule
& \multicolumn{2}{c|}{FID $\downarrow$} & \multicolumn{2}{c|}{FVD $\downarrow$} & \multicolumn{2}{c|}{ADE $(\times10^{-3})$ $\downarrow$} & Compliance \\
Perturbation & Short & Long & Short & Long & Short & Long & ($p$) \\
\midrule
  \multicolumn{8}{l}{\textit{CFG = 1 (default guidance)}} \\
\midrule
  Lateral $+1$\,m & 18.4 & 26.0 & 155.0 & 224.5 & 4.4{\scriptsize$\,\pm\,13.9$} & 10.8{\scriptsize$\,\pm\,13.7$} & +0.66\,m (181/237, $p=2e{-}16$) \\
  Lateral $+0.5$\,m & 17.0 & 24.5 & 139.3 & 202.7 & 4.0{\scriptsize$\,\pm\,12.8$} & 10.3{\scriptsize$\,\pm\,14.1$} & +0.47\,m (185/237, $p=1e{-}18$) \\
  Nominal & 16.2 & 23.5 & 128.5 & 188.4 & 3.6{\scriptsize$\,\pm\,13.8$} & 9.9{\scriptsize$\,\pm\,13.4$} & -- \\
  Lateral $-0.5$\,m & 16.9 & 24.5 & 140.9 & 202.8 & 4.0{\scriptsize$\,\pm\,14.2$} & 10.2{\scriptsize$\,\pm\,13.4$} & -0.42\,m (176/237, $p=4e{-}14$) \\
  Lateral $-1$\,m & 18.3 & 26.1 & 156.8 & 224.0 & 4.4{\scriptsize$\,\pm\,14.4$} & 11.3{\scriptsize$\,\pm\,14.8$} & -0.75\,m (176/237, $p=4e{-}14$) \\
  Wiggle $0.5$\,m & 17.2 & 24.0 & 171.9 & 250.4 & 4.8{\scriptsize$\,\pm\,14.1$} & 11.8{\scriptsize$\,\pm\,14.6$} & slope 0.86 (210/237, $p=3e{-}36$) \\
  Speed $0.85\times$ & 19.5 & 27.5 & 168.8 & 208.7 & 6.3{\scriptsize$\,\pm\,15.9$} & 11.9{\scriptsize$\,\pm\,13.9$} & 60\% (239/400, $p=1e{-}4$) \\
\midrule
  \multicolumn{8}{l}{\textit{CFG = 2}} \\
\midrule
  Lateral $+1$\,m & 17.2 & 23.5 & 146.3 & 200.7 & 4.4{\scriptsize$\,\pm\,16.0$} & 10.1{\scriptsize$\,\pm\,14.2$} & +0.71\,m (178/237, $p=5e{-}15$) \\
  Lateral $+0.5$\,m & 15.7 & 21.2 & 126.8 & 165.8 & 3.8{\scriptsize$\,\pm\,14.7$} & 9.0{\scriptsize$\,\pm\,13.1$} & +0.52\,m (179/237, $p=2e{-}15$) \\
  Nominal & 14.8 & 19.3 & 115.2 & 151.2 & 3.6{\scriptsize$\,\pm\,15.9$} & 9.5{\scriptsize$\,\pm\,14.0$} & -- \\
  Lateral $-0.5$\,m & 15.7 & 21.6 & 125.8 & 169.3 & 3.9{\scriptsize$\,\pm\,13.3$} & 9.5{\scriptsize$\,\pm\,14.3$} & -0.40\,m (176/237, $p=4e{-}14$) \\
  Lateral $-1$\,m & 17.5 & 23.9 & 147.7 & 200.7 & 4.5{\scriptsize$\,\pm\,15.2$} & 10.4{\scriptsize$\,\pm\,14.8$} & -0.89\,m (183/237, $p=1e{-}17$) \\
  Wiggle $0.5$\,m & 15.6 & 21.3 & 155.0 & 213.2 & 4.7{\scriptsize$\,\pm\,13.1$} & 10.5{\scriptsize$\,\pm\,14.4$} & slope 1.18 (218/237, $p=5e{-}44$) \\
  Speed $0.85\times$ & 18.1 & 23.0 & 157.6 & 181.9 & 7.1{\scriptsize$\,\pm\,14.7$} & 10.7{\scriptsize$\,\pm\,13.3$} & 62\% (250/400, $p=7e{-}7$) \\
\bottomrule
\end{tabular}
\end{table}

\subsection{Quality beyond the reconstructed region}
\label{app:beyondmap}

The geo-reference only contains geometry where the original drive observed the scene. To measure what happens when the ego leaves this region, we reconstruct the map from only the first $L \in \{1, 2, 4, 8\}$\,s of each clip and keep it frozen. We use the same GeoSCope (Orbis v1) checkpoint without retraining. We evaluate on the 172 nuPlan-Turns clips whose 20\,s window lies within a single chunk of the existing SLAM reconstruction, so the frozen maps can be built without re-running SLAM.

\begin{figure}[ht!]
    \centering
    \includegraphics[width=1\linewidth]{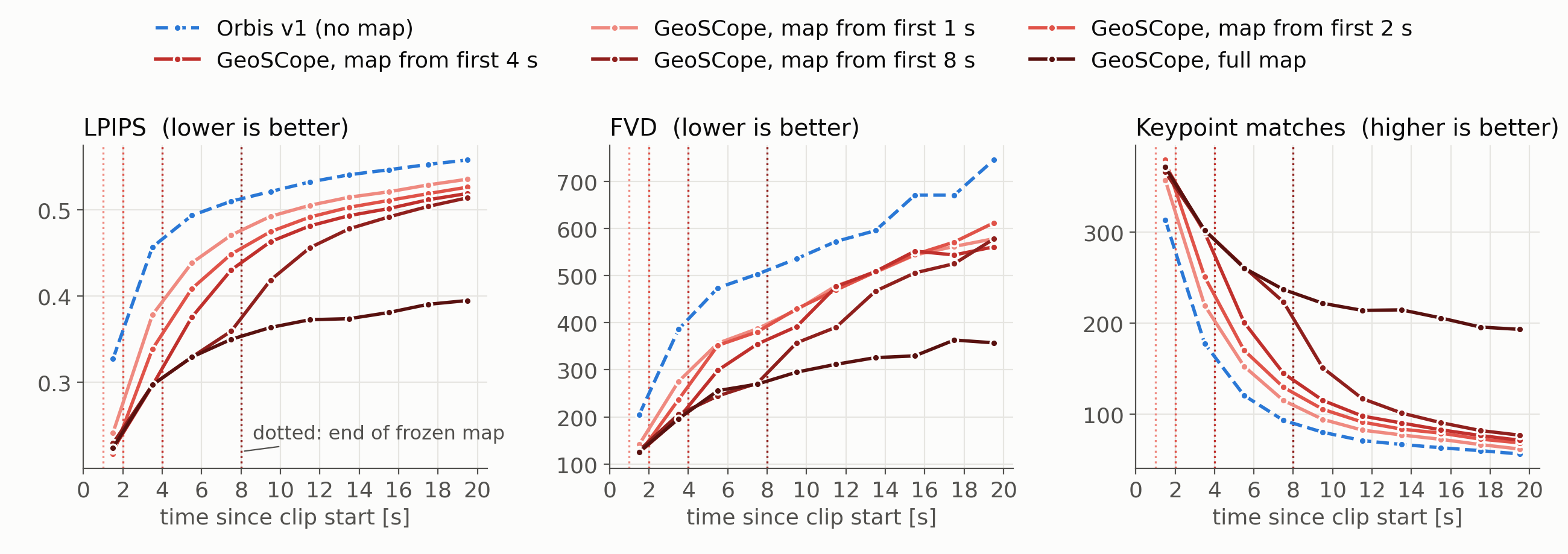}
    \caption{Rollout quality when the ego drives out of a frozen map (172 nuPlan-Turns clips; LPIPS, FVD and LightGlue keypoint matches over 2\,s windows). GeoSCope uses a map built from only the first $L$\,s of the clip (red, light to dark: $L = 1, 2, 4, 8$\,s, full clip) and Orbis v1 uses no map (blue, dashed). Dotted lines mark the last frame used to build each frozen map.}
    \label{fig:beyondmap}
\end{figure}

Figure~\ref{fig:beyondmap} shows the result. While the map still covers the view, each frozen-map curve follows the full-map curve. Beyond it, quality degrades smoothly toward the base model but stays better than it. Averaged over the 10--20\,s window, which lies outside every frozen map, FVD is 493--541 with a frozen map, 651 for Orbis v1 and 338 with the full map. Beyond the map, GeoSCope thus falls back to the learned prior of the world model rather than failing.

\subsection{Agent-Level Evaluation}
\label{app:agent_level}

We evaluate generated vehicles against the nuPlan lidar tracks and the HD map on the 400 nuPlan-Turns clips.

\textbf{Pipeline.}
We detect vehicles with YOLOv8x and lift each detection to a metric ground-plane position, fusing the box bottom on the ground plane with the box height under a class-size prior. Using the recorded ego pose and camera calibration, positions are placed in the world frame of the logged drive (range $\leq$25\,m). They are then tracked with Hungarian matching and a constant-velocity Kalman smoother, and matched to lidar tracks by 2D IoU. A track is moving if its median speed exceeds 3\,m/s. We use the first 4\,s of each rollout, as later traffic legitimately diverges from the log. Errors therefore reflect both agent placement and how closely the rollout follows the logged ego path. The same pipeline on the recorded video gives the measurement floor.

\textbf{Metrics.}
\emph{Position error}: centre distance to the matched lidar track. \emph{Speed error}: for moving matched tracks. \emph{'In lane'} and \emph{'on sidewalk'}: share of moving-track observations inside lane or walkway polygons of the HD map. \emph{Unmatched by proximity}: share of tracks more than 2.5\,m from every lidar box in at least 60\% of their frames.

\begin{table}[h]
\centering
\caption{Vehicle tracks in the first 4\,s of 400 nuPlan-Turns rollouts. The same pipeline on the recorded video gives the measurement floor.}
\label{tab:agent_level}
%\small
\resizebox{\linewidth}{!}{%
\begin{tabular}{lccccc}
\toprule
Model & Pos.\ err.\ (m) $\downarrow$ & Speed err., moving (m/s) $\downarrow$ & In lane $\uparrow$ & On sidewalk $\downarrow$ & Unmatched by proximity $\downarrow$ \\
\midrule
Recorded video (floor) & 1.02 & 0.72 & 97\% & 0\% & 7\% \\
\midrule
Orbis v1 & 2.39 & 1.60 & 81\% & 4\% & 36\% \\
+ GeoSCope & \textbf{1.23} & \textbf{1.18} & \textbf{93\%} & \textbf{0\%} & \textbf{12\%} \\
\midrule
Orbis v2 & 2.28 & 1.50 & 85\% & 5\% & 34\% \\
+ GeoSCope & \textbf{1.19} & \textbf{0.96} & \textbf{95\%} & \textbf{1\%} & \textbf{13\%} \\
\bottomrule
\end{tabular}}
\end{table}

\textbf{Results.}
GeoSCope roughly halves position error for both backbones, to 1.19--1.23\,m compared with 1.02\,m on recorded video, and reduces speed error (Table~\ref{tab:agent_level}). Moving vehicles in lane rise from 81--85\% to 93--95\%, vehicles on sidewalks drop from 4--5\% to 0--1\%, and unmatched tracks drop from 34--36\% to 12--13\%. Interaction dynamics are not improved. We select 51 clips whose lidar log contains crossing traffic: 92 vehicles within 40\,m that move faster than 2\,m/s at $60$--$120^\circ$ to the ego heading for at least 0.4\,s. A crossing vehicle counts as \emph{found} if a generated track is matched to it, and its crossing is \emph{preserved} if that track also moves at $60$--$120^\circ$ to the ego in most of its frames. GeoSCope finds slightly more crossing vehicles than its base model (17\% vs.\ 11\% for Orbis v1, 26\% vs.\ 21\% for Orbis v2; 26\% on recorded video, where the 25\,m tracking range and occlusions limit detection). Of the found vehicles, the share that keeps its crossing motion is unchanged (31--33\% vs.\ 30--32\%; 58\% on recorded video).

\textbf{Counterfactual consistency.}
Without ground truth, we measure consistency in the world frame. For GeoSCope (Orbis v1), each of the 400 clips is generated on the logged path and with the ego path shifted laterally by about 2\,m to either side, keeping context and seed fixed. In none of the 1{,}200 rollouts (including the logged path) does a tracked vehicle overlap the ego. Parked vehicles stay anchored and moving vehicles largely stay in lane (Table~\ref{tab:agent_cf}). Reactions of other agents come from the world-model prior and are not evaluated.

\begin{table}[h]
\centering
\caption{World-frame consistency of GeoSCope (Orbis v1) under lateral counterfactuals, 400 nuPlan-Turns clips, 0--6\,s. \emph{Parked anchored}: share of logged parked vehicles in view ($\leq$30\,m) with a tracked vehicle within 2\,m.}
\label{tab:agent_cf}
\small
\begin{tabular}{lcccc}
\toprule
Ego path & Ego overlaps $\downarrow$ & Parked anchored $\uparrow$ & Moving in lane $\uparrow$ & Moving on sidewalk $\downarrow$ \\
\midrule
Logged & 0 & 0.51 & 93\% & 0.9\% \\
Lateral, left & 0 & 0.46 & 90\% & 1.3\% \\
Lateral, right & 0 & 0.45 & 82\% & 2.3\% \\
\bottomrule
\end{tabular}
\end{table}

\subsection{Starting Without Recorded Frames}
\label{app:no_context}

We replace all context frames with Gaussian noise, so the model sees only the map, and evaluate on the 400 nuPlan-Turns clips with the protocol of Table~\ref{tab:long_horizon}.

\begin{table}[h]
\centering
\caption{Starting without recorded frames, 400 nuPlan-Turns clips (protocol of Table~\ref{tab:long_horizon}, short / long = 0--8\,s / 8--16\,s).}
\label{tab:no_context}
\small
\begin{tabular}{llccc}
\toprule
Model & Context & LPIPS $\downarrow$ & FVD $\downarrow$ & KP matches $\uparrow$ \\
\midrule
Orbis v2 & recorded & 0.440 / 0.536 & 222.3 / 333.8 & 178.7 / 63.6 \\
+ GeoSCope & none (map only) & 0.405 / 0.383 & 222.7 / 211.1 & 222.7 / 239.2 \\
+ GeoSCope & recorded & \textbf{0.304 / 0.360} & \textbf{131.9 / 183.3} & \textbf{298.1 / 249.4} \\
\bottomrule
\end{tabular}
\end{table}

The map-only start is worse than recorded context over the short horizon, as appearance and traffic must be invented, but on par with or better than the base model with recorded context (Table~\ref{tab:no_context}). Over the long horizon, it approaches GeoSCope with recorded context. The generated vehicles are plausible but do not match the real traffic, so off-log starts still need real context to initialise agents. Box-conditioned models could supply agents explicitly in future work.

\clearpage
\newpage
\section{Qualitative Results}

\subsection{Conditioning mechanism}

\begin{figure}[ht!]
    \centering
    \includegraphics[width=0.95\linewidth]{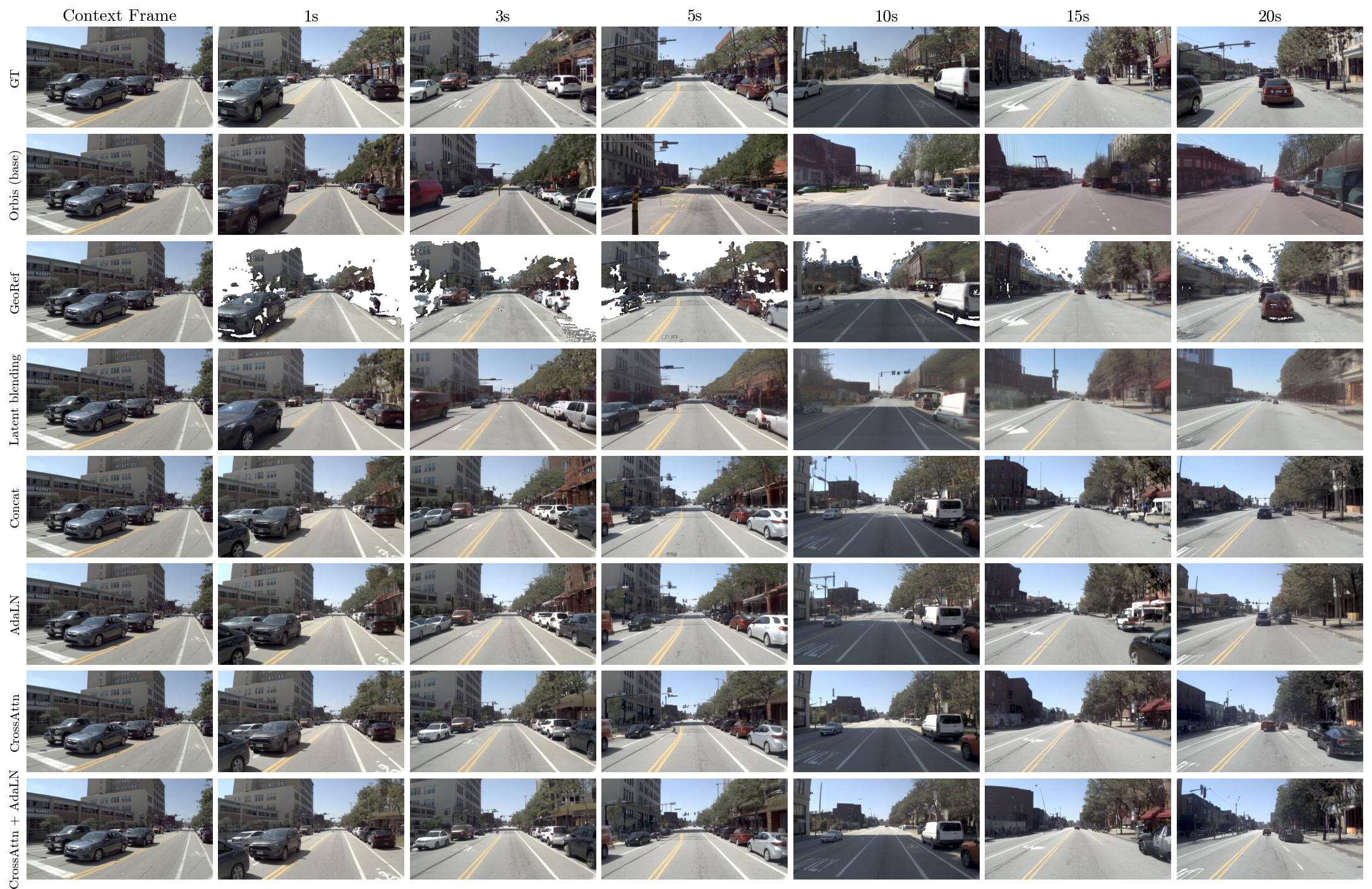}
    \vspace{-0.25cm}
    \caption{Qualitative comparison of conditioning mechanisms on nuPlan-Turns (sequence 123).}
    \label{fig:conditioning_mechanism}
\end{figure}

\begin{figure}[ht!]
    \centering
    \includegraphics[width=0.95\linewidth]{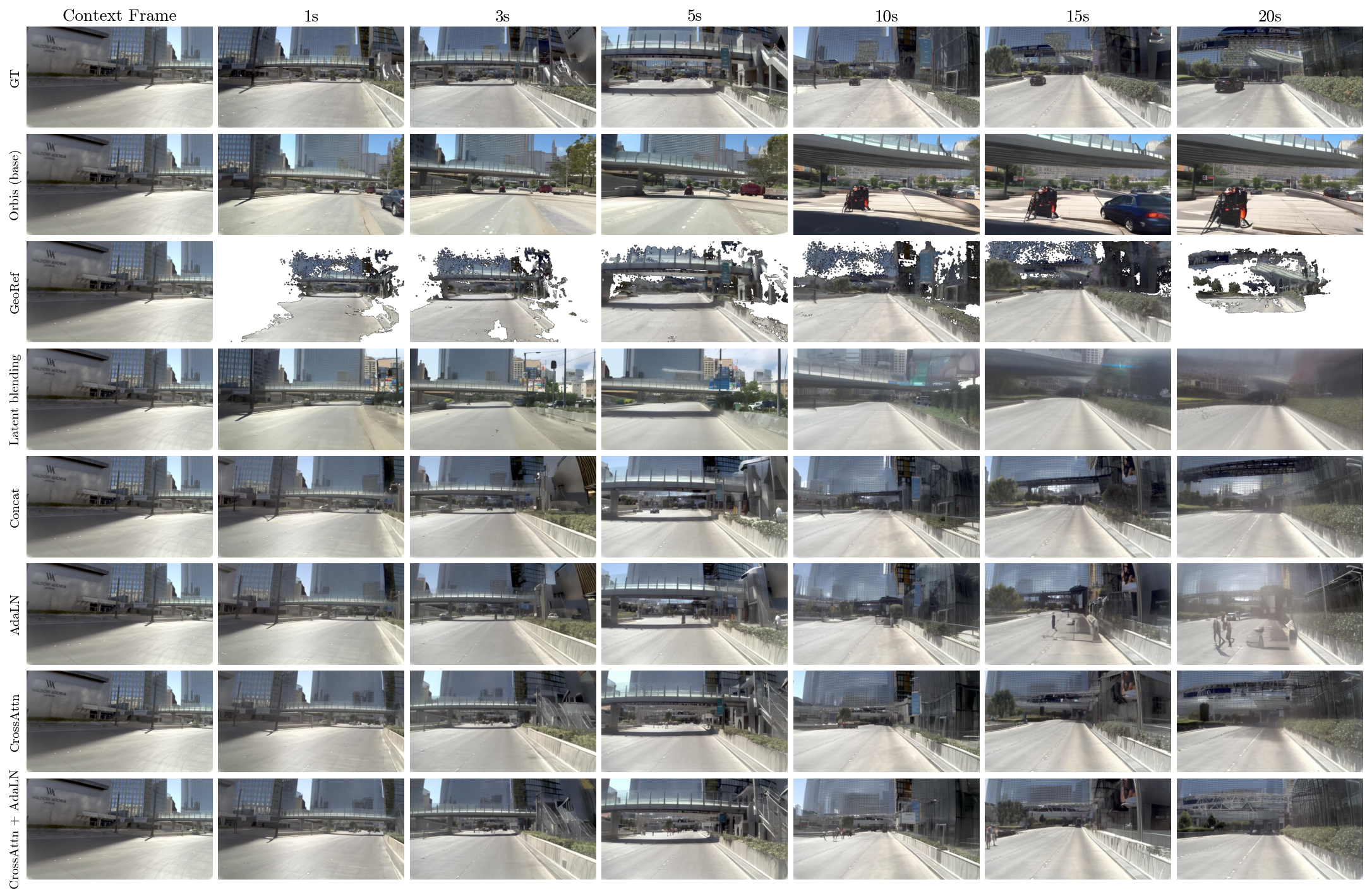}
    \vspace{-0.25cm}
    \caption{Qualitative comparison of conditioning mechanisms on nuPlan-Turns (sequence 131).}
    \label{fig:conditioning_mechanism2}
\end{figure}

\Cref{fig:conditioning_mechanism,fig:conditioning_mechanism2} show representative rollouts for each injection mechanism. Cross-attention recovers fine scene structure that latent blending and concatenation are unable to retain over time.

\newpage
\subsection{Long-Horizon Rollouts}

\begin{figure}[ht!]
    \centering
    \includegraphics[width=0.95\linewidth]{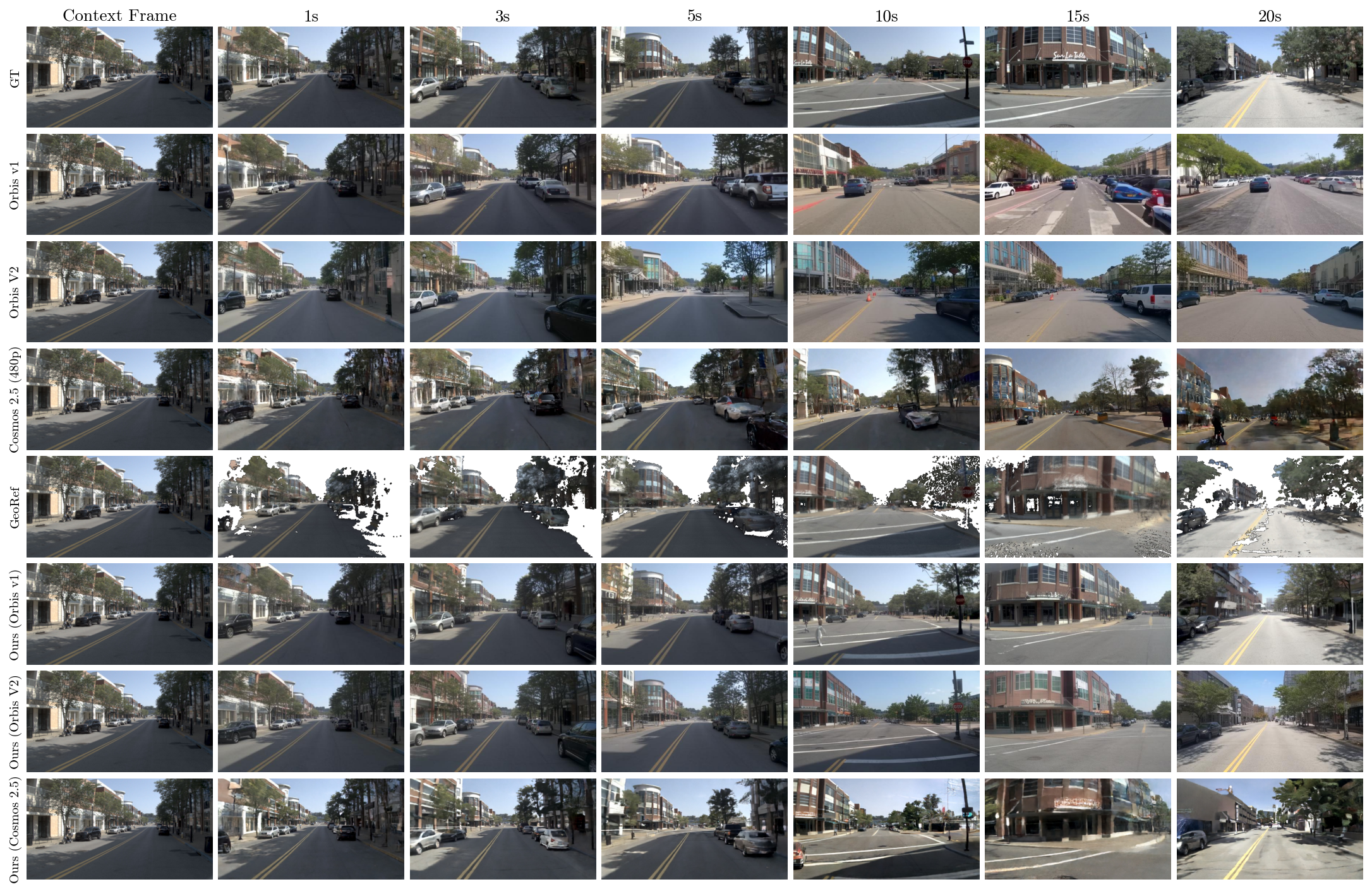}
    \vspace{-0.25cm}
    \caption{Qualitative comparison of the different methods on nuPlan-Turns (sequence 001).}
    \label{fig:long_horizon1}
\end{figure}

\begin{figure}[ht!]
    \centering
    \includegraphics[width=0.95\linewidth]{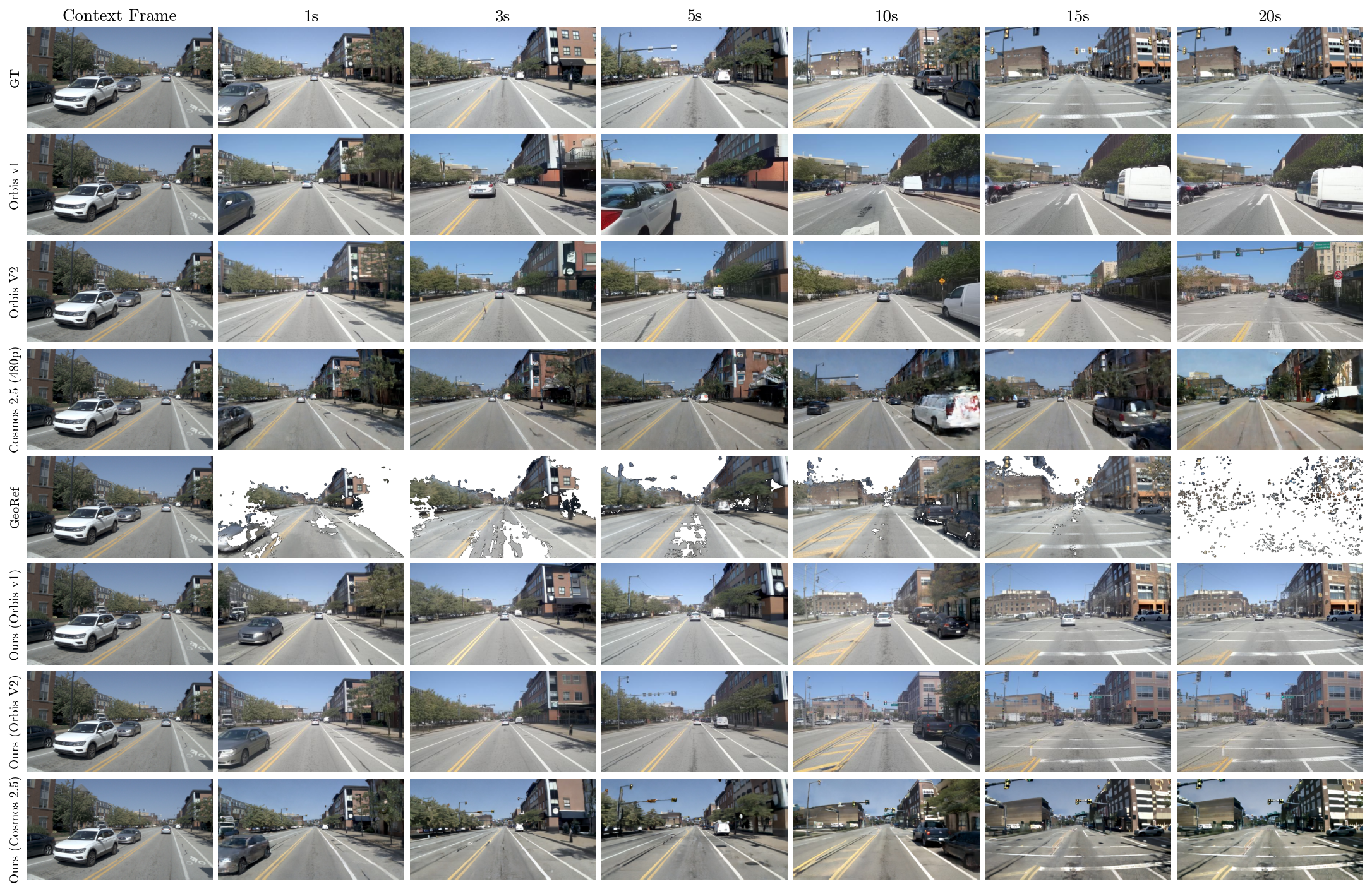}
    \vspace{-0.25cm}
    \caption{Qualitative comparison of the different methods on nuPlan-Turns (sequence 357).}
    \label{fig:long_horizon2}
\end{figure}

\Cref{fig:long_horizon1,fig:long_horizon2} highlight how the geo-reference information anchors the rollouts in the actual surroundings of the original drive.  

\subsection{Coverage of the Geometric Reference}

\begin{figure}[ht!]
    \centering
    \includegraphics[width=1\linewidth]{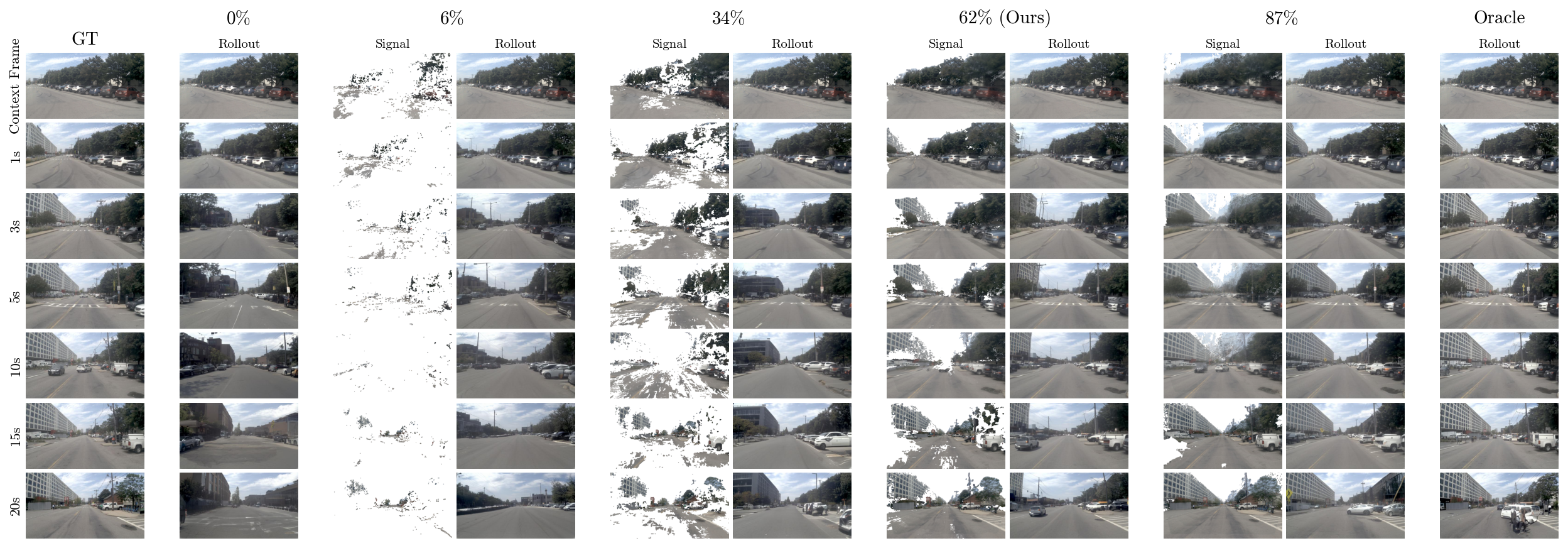}
    \vspace{-0.15cm}
    \caption{Qualitative effect of geo-reference coverage on nuPlan-Turns (sequence 116).}
    \label{fig:coverage1}
\end{figure}

\begin{figure}[ht!]
    \centering
    \includegraphics[width=1\linewidth]{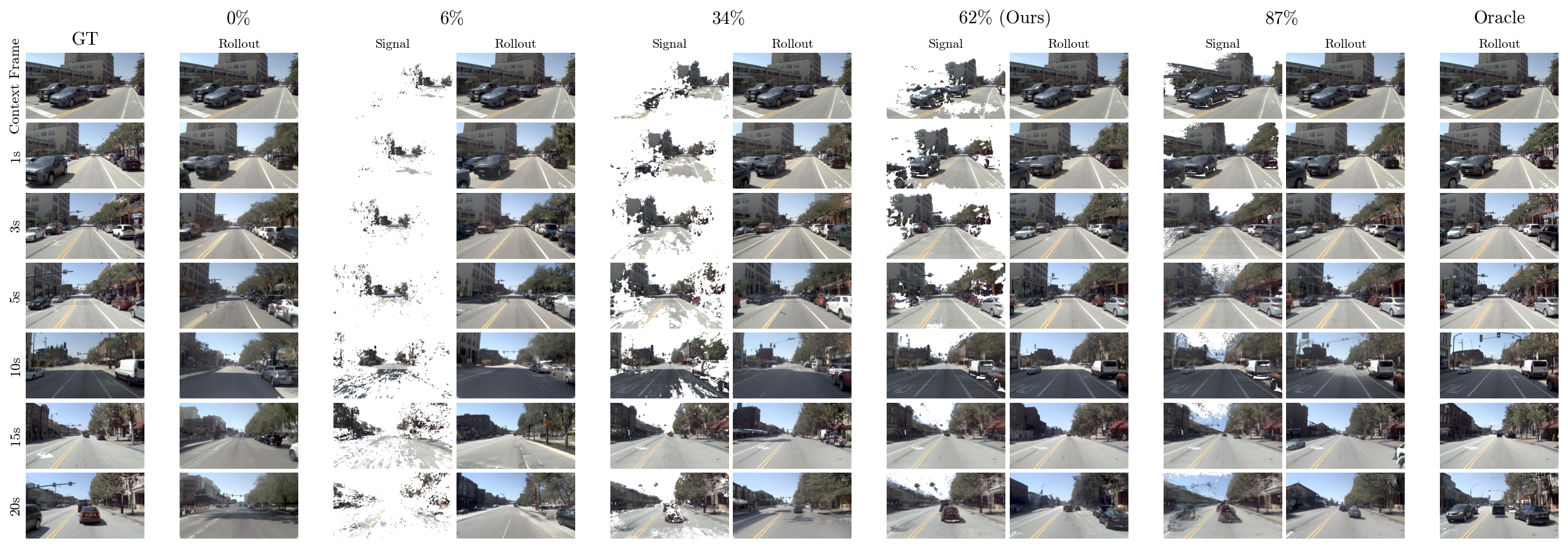}
    \vspace{-0.15cm}
    \caption{Qualitative effect of geo-reference coverage on nuPlan-Turns (sequence 123).}
    \label{fig:coverage2}
\end{figure}

\begin{figure}[ht!]
    \centering
    \includegraphics[width=1\linewidth]{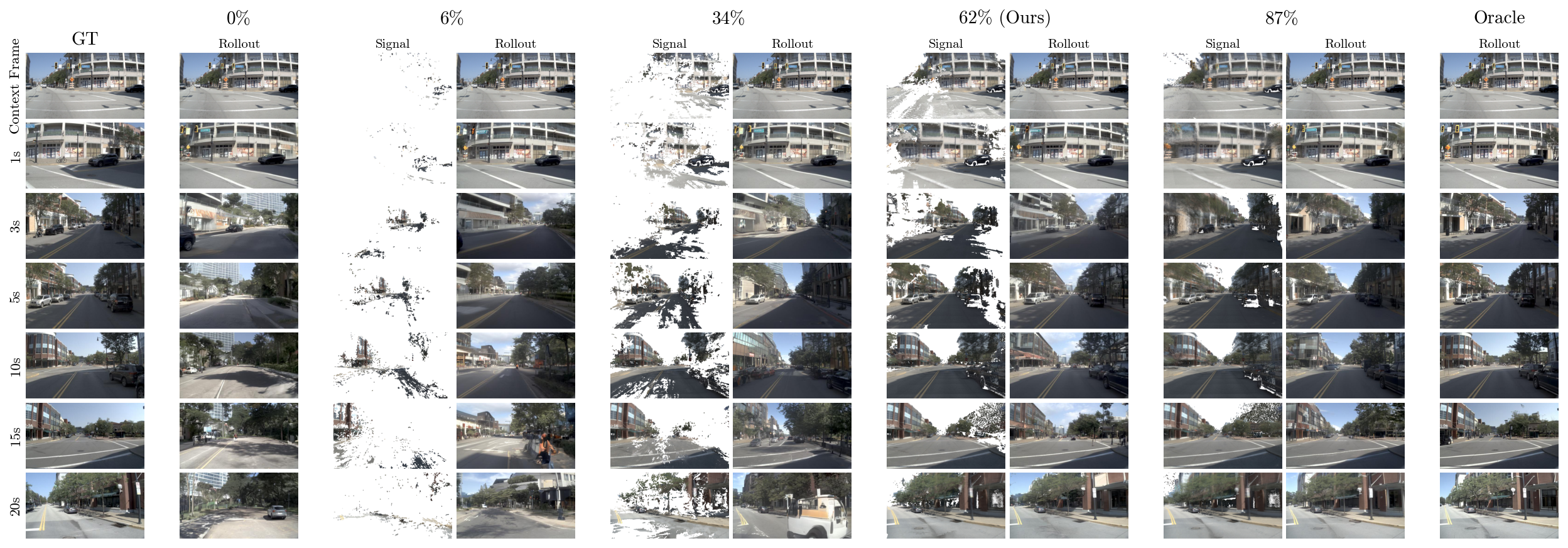}
    \vspace{-0.15cm}
    \caption{Qualitative effect of geo-reference coverage on nuPlan-Turns (sequence 000).}
    \label{fig:coverage3}
\end{figure}

\Cref{fig:coverage1,fig:coverage2,fig:coverage3} illustrate the coverage sweep of \Cref{app:tab_ablation-coverage}: as point-cloud coverage increases, background structure sharpens and long-horizon drift visibly decreases.

\newpage
\subsection{Localization error}

\begin{figure}[ht!]
    \centering
    \includegraphics[width=1\linewidth]{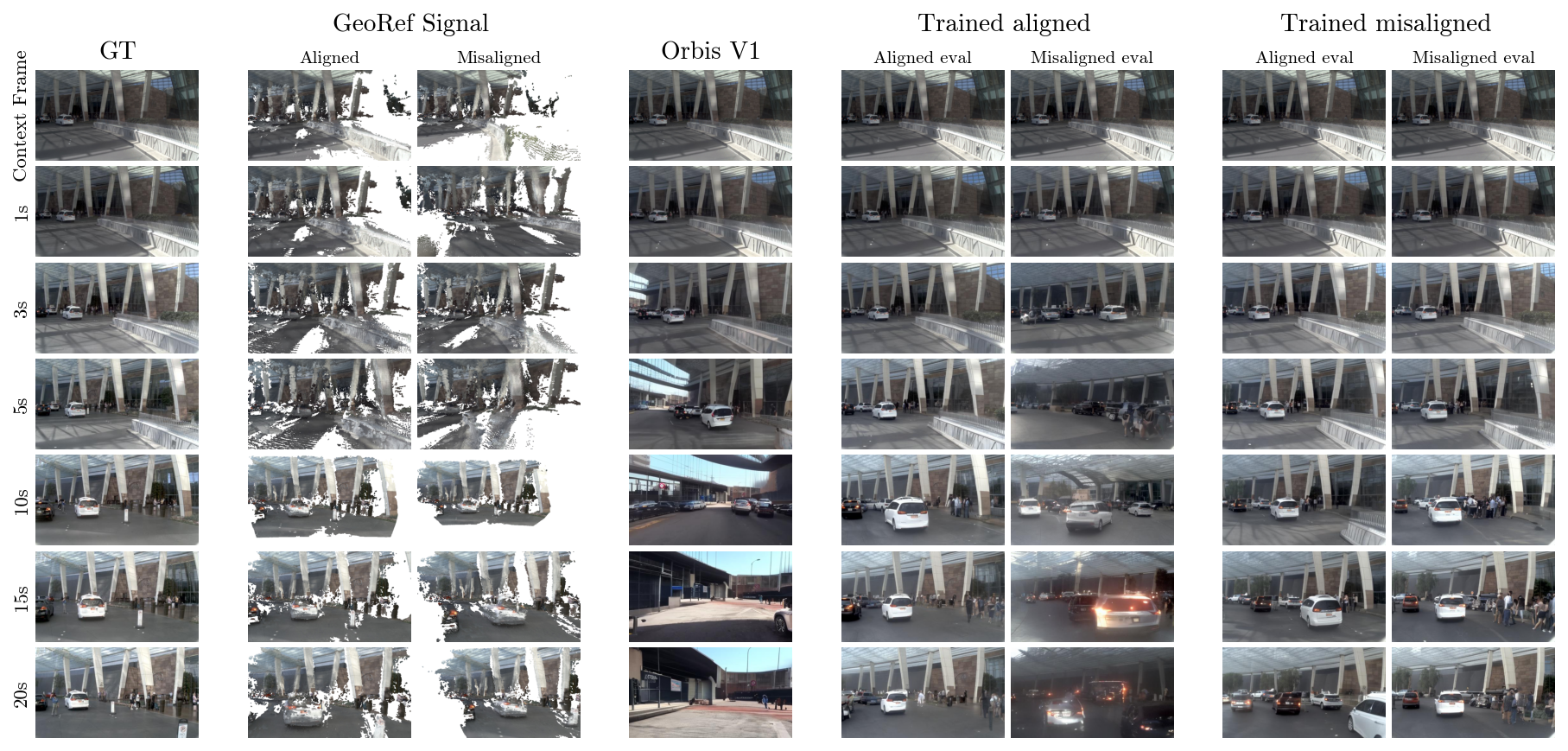}
    \caption{Rollout under a $3$\,m pose perturbation on nuPlan-Turns. The aligned-trained model follows the displaced geometry, while the perturbation-trained model stays on the recorded route.}
    \label{fig:alignment}
\end{figure}

\begin{figure}[ht!]
    \centering
    \includegraphics[width=1\linewidth]{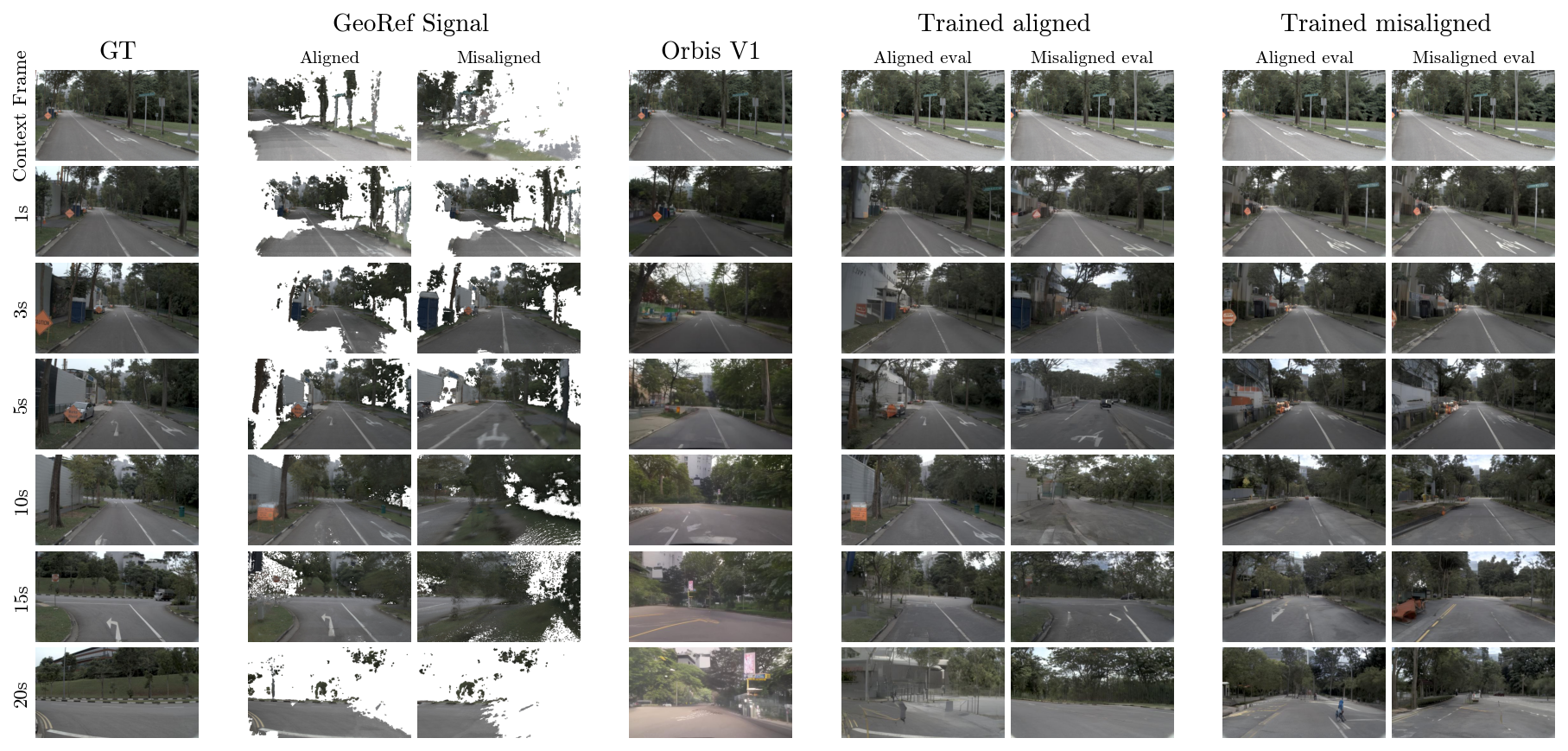}
    \caption{Rollout under a $3$\,m pose perturbation on nuPlan-Turns.}
    \label{fig:alignment2}
\end{figure}

\Cref{fig:alignment,fig:alignment2} show rollouts under a 3m pose perturbation (\Cref{sec:localization}, \Cref{app:tab_alignment}): the aligned-trained model visibly chases the displaced geometry, while the perturbation-trained model ignores the misalignment and stays on the recorded route.

\newpage
\subsection{Counterfactual Rollout}

\begin{figure}[ht!]
    \centering
    \includegraphics[width=0.85\linewidth]{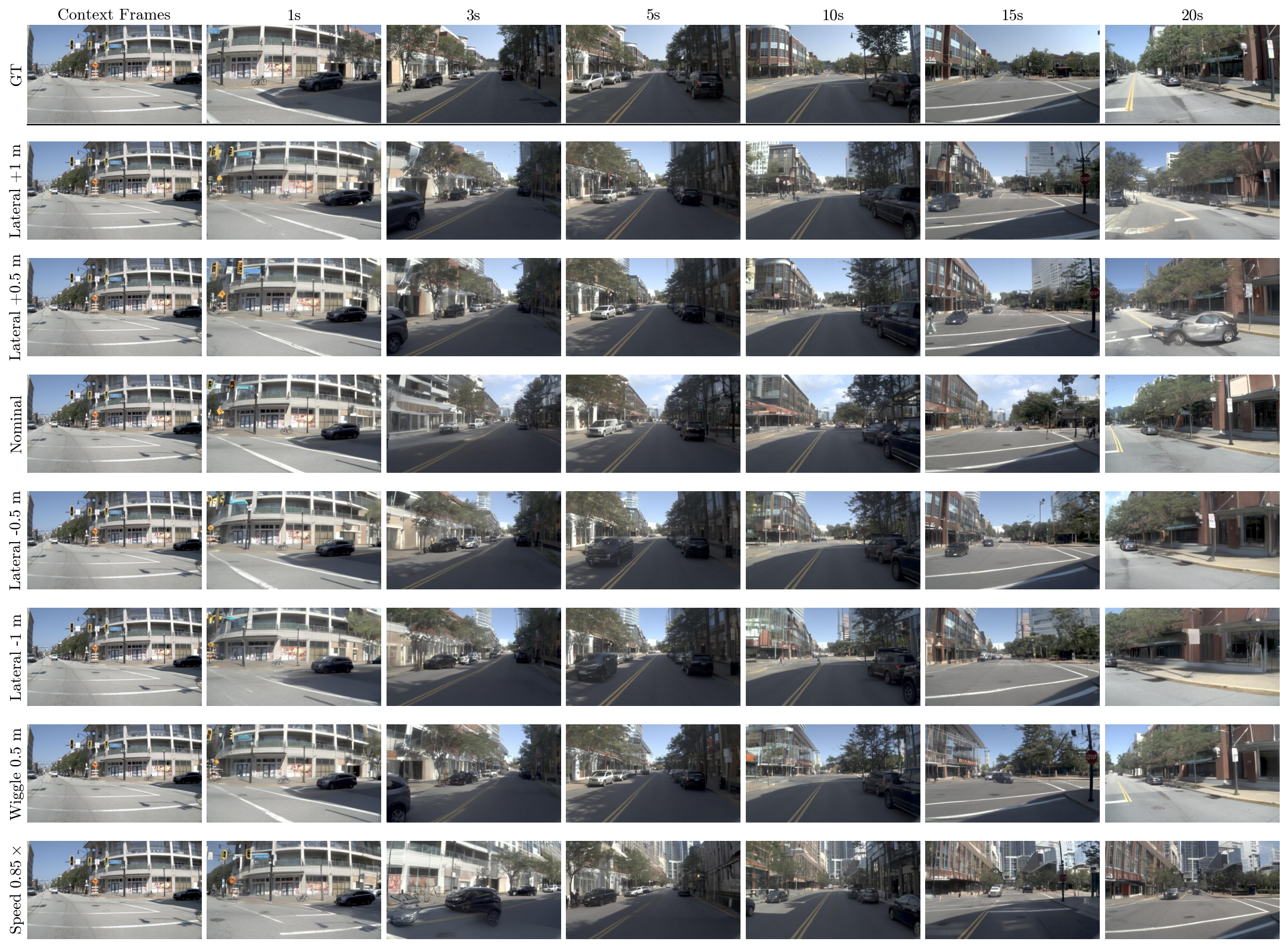}
    \caption{Counterfactual rollouts on nuPlan-Turns. Rows show lateral offsets from $+1$ to $-1$\,m, followed by the sinusoidal wiggle and the reduced-speed command.}
    \label{fig:counterfactual}
\end{figure}

\begin{figure}[ht!]
    \centering
    \includegraphics[width=0.85\linewidth]{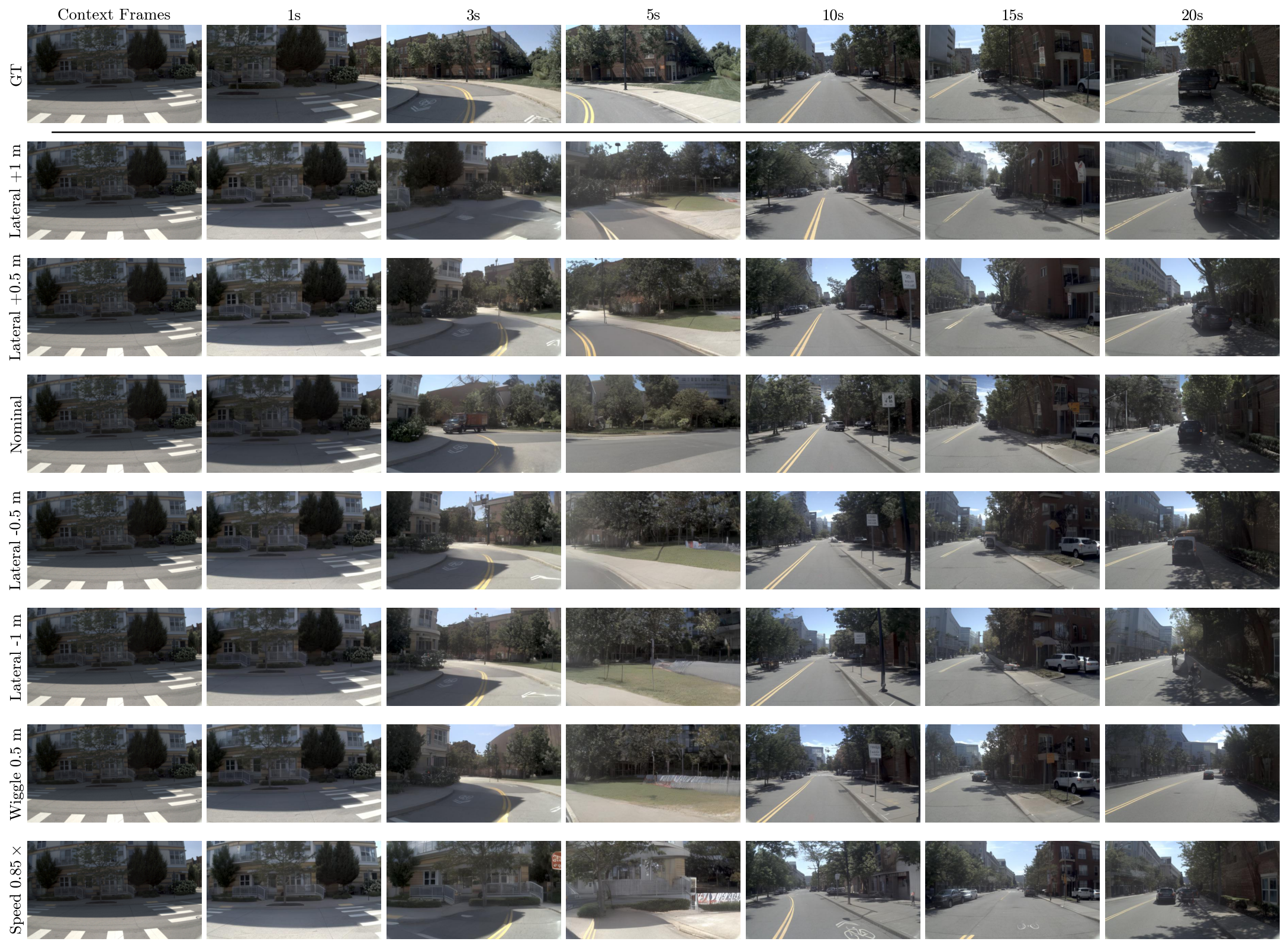}
    \caption{Counterfactual rollouts on nuPlan-Turns}
    \label{fig:counterfactual2}
\end{figure}

Figures~\ref{fig:counterfactual} and~\ref{fig:counterfactual2} illustrate counterfactual rollouts with lateral offsets ranging from $+1$ to $-1$ meters. The bottom two rows highlight the resulting sinusoidal lateral trajectory (``wiggle") and the corresponding reduction in speed. We refer the reader to the supplementary videos for a more detailed qualitative assessment.

\newpage
\subsection{Multi-seed Rollout Diversity}

\begin{figure}[ht!]
    \centering
    \includegraphics[width=1\linewidth]{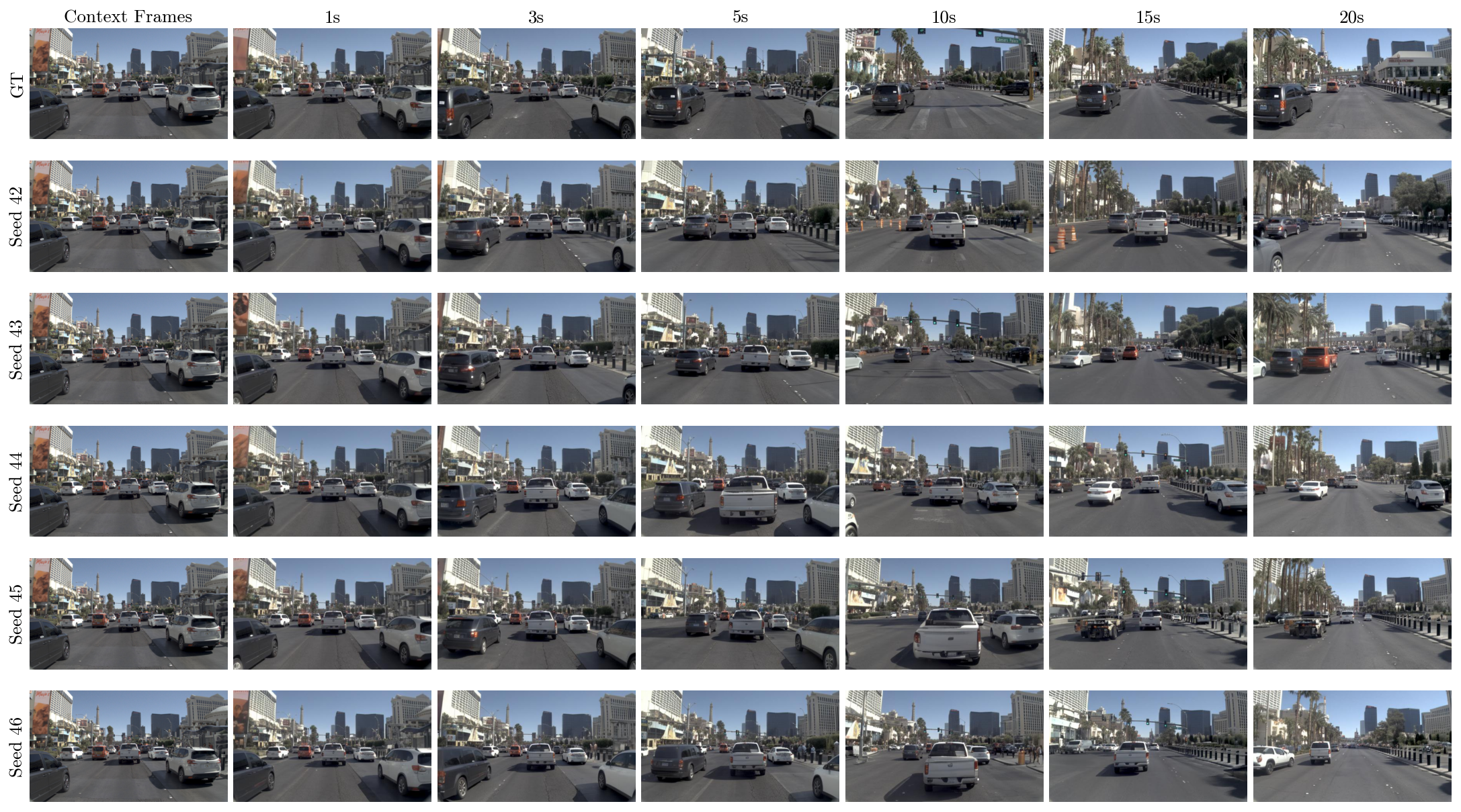}
    \caption{Rollouts from different sampling seeds under identical geo-reference conditioning on nuPlan-Turns. Scene structure is shared across seeds, while the dynamic content differs.}
    \label{fig:seed_diversity}
\end{figure}

\begin{figure}[ht!]
    \centering
    \includegraphics[width=1\linewidth]{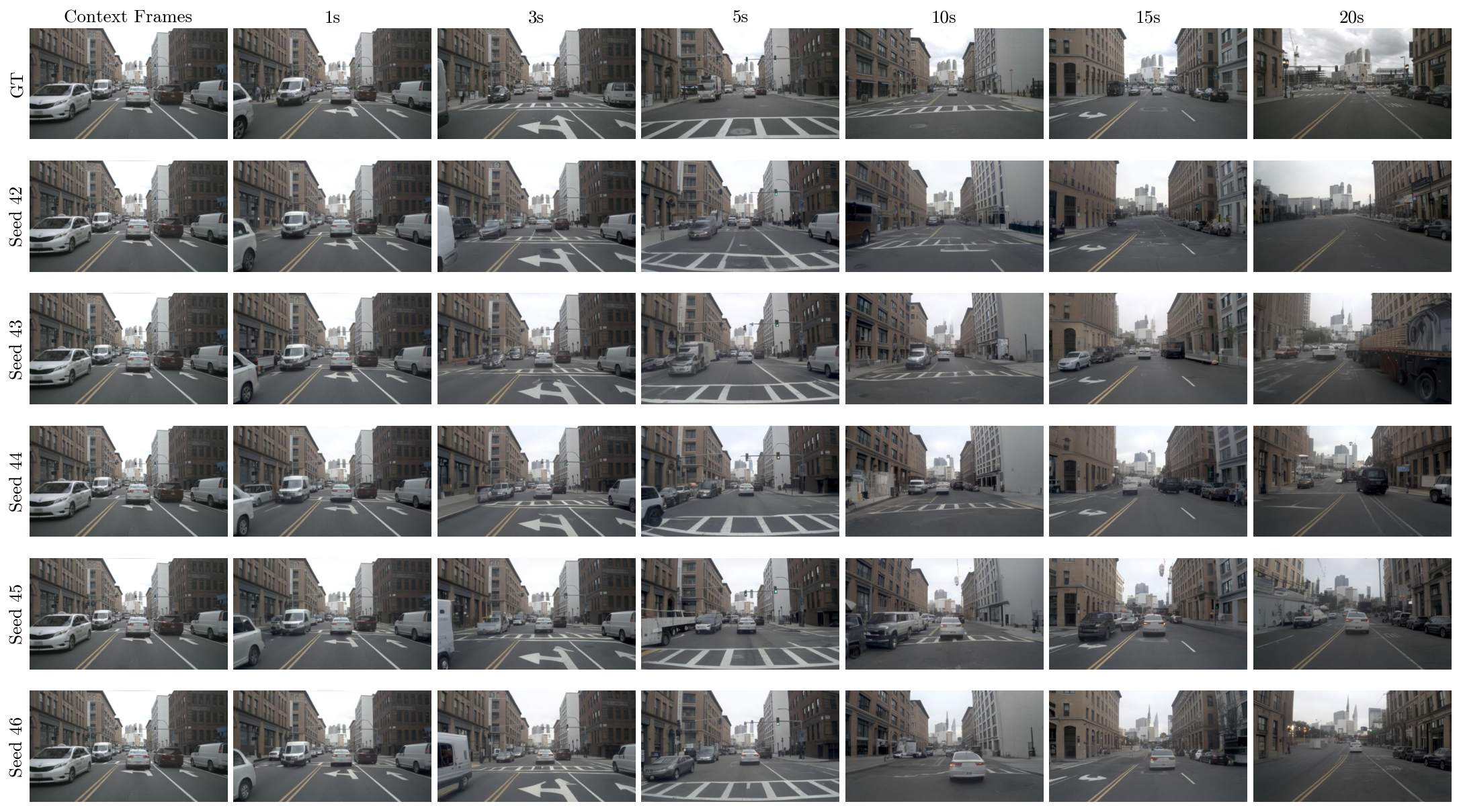}
    \caption{Rollouts from different sampling seeds under identical geo-reference
conditioning on nuScenes}
    \label{fig:seed_diversity2}
\end{figure}

\Cref{fig:seed_diversity,fig:seed_diversity2} vary only the sampling seed under identical geo-reference conditioning on nuPlan and nuScenes, showing that the geo-reference signal constrains scene structure without collapsing the diversity of dynamic content across samples.

\newpage
\subsection{Multi-view rollouts}

\begin{figure}[ht!]
    \centering
    \includegraphics[width=0.9\linewidth]{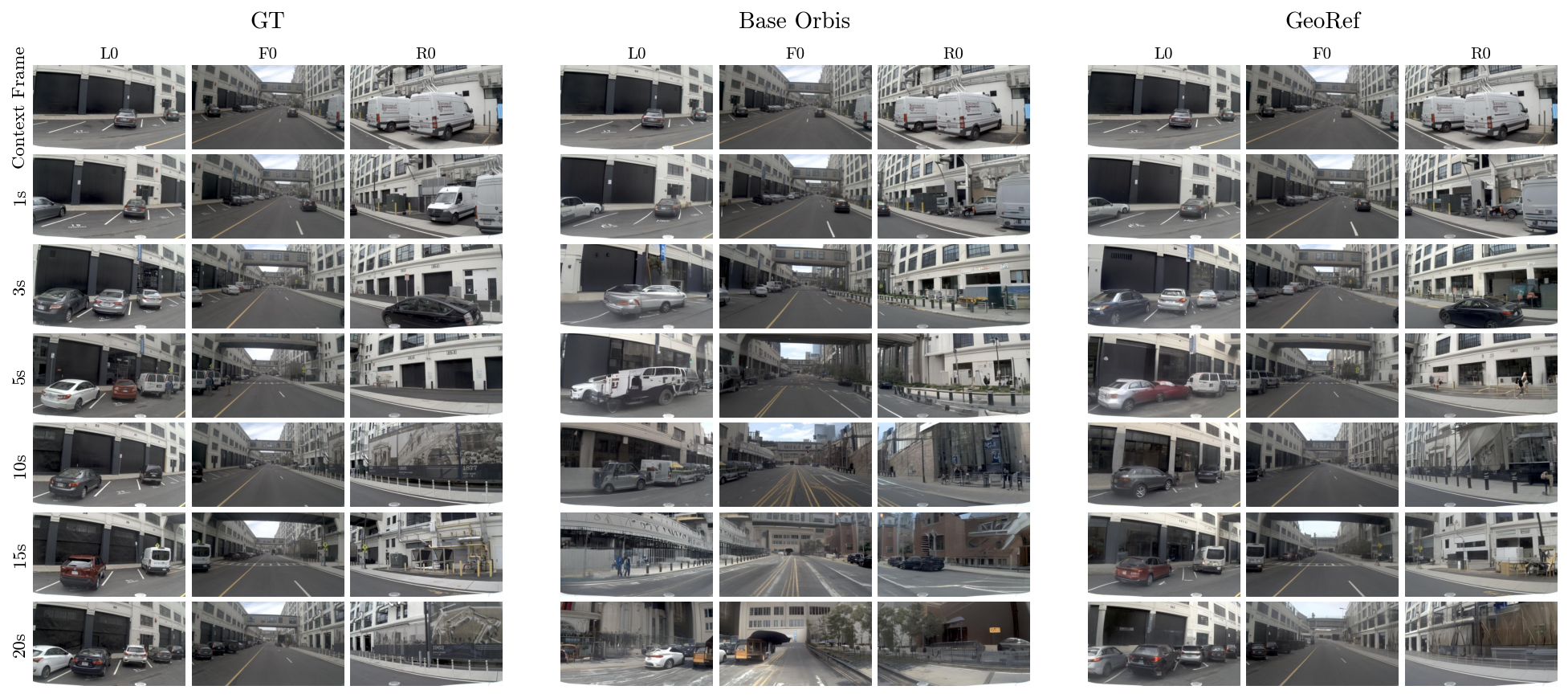}
    \caption{Synchronized left, front and right rollouts on nuPlan-Turns, without and with geo-reference conditioning.}
    \label{fig:multiview}
\end{figure}

\begin{figure}[ht!]
    \centering
    \includegraphics[width=0.9\linewidth]{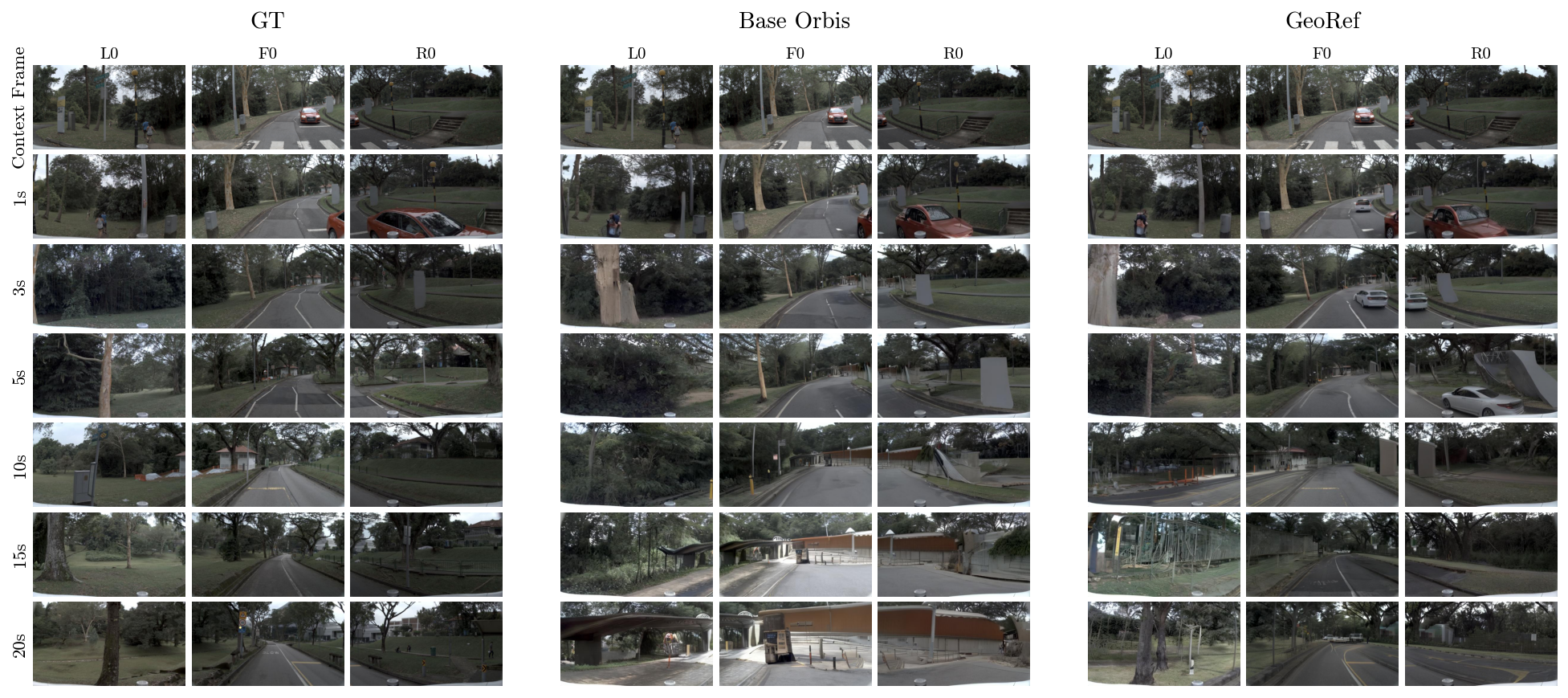}
    \caption{Synchronized multi-view rollouts on nuPlan-Turns.}
    \label{fig:multiview2}
\end{figure}

\begin{figure}[ht!]
    \centering
    \includegraphics[width=0.9\linewidth]{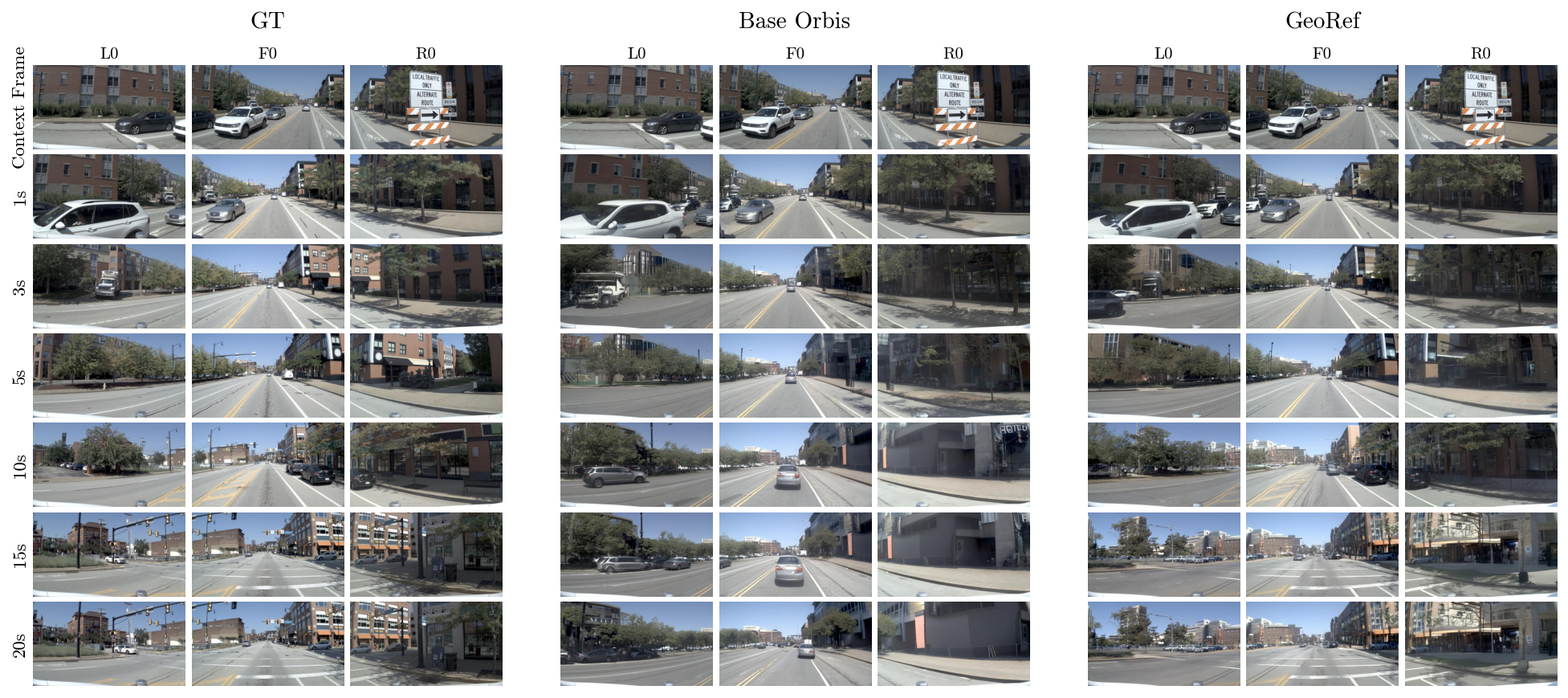}
    \caption{Synchronized multi-view rollouts on nuPlan-Turns.}
    \label{fig:multiview3}
\end{figure}

\Cref{fig:multiview,fig:multiview2,fig:multiview3} show synchronized front/left/right rollouts corresponding to \Cref{tab:multiview}: cross-attention conditioning keeps geometry consistent across all three views, most visibly on the forward-facing camera.
\vspace{-0.25cm}

\vspace{-0.25cm}

\vspace{-0.25cm}

\clearpage
\newpage
\subsection{Weather and Lighting Differences}

\begin{figure}[ht!]
    \centering
    \includegraphics[width=0.9\linewidth]{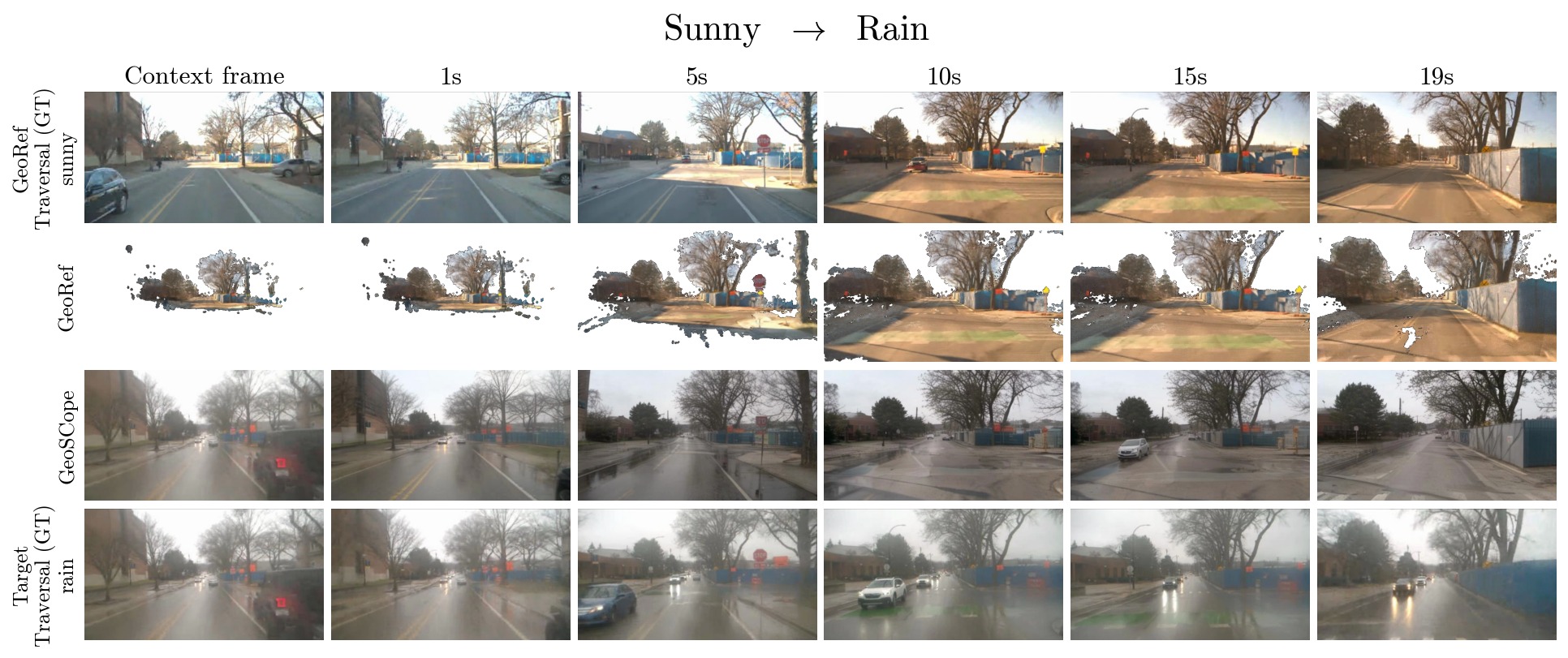}
    \caption{ Sunny map, rainy target drive. GeoSCope keeps the wet road and overcast sky of the target drive while following the mapped road layout.}
    \label{fig:lightning1}
\end{figure}

\begin{figure}[ht!]
    \centering
    \includegraphics[width=0.9\linewidth]{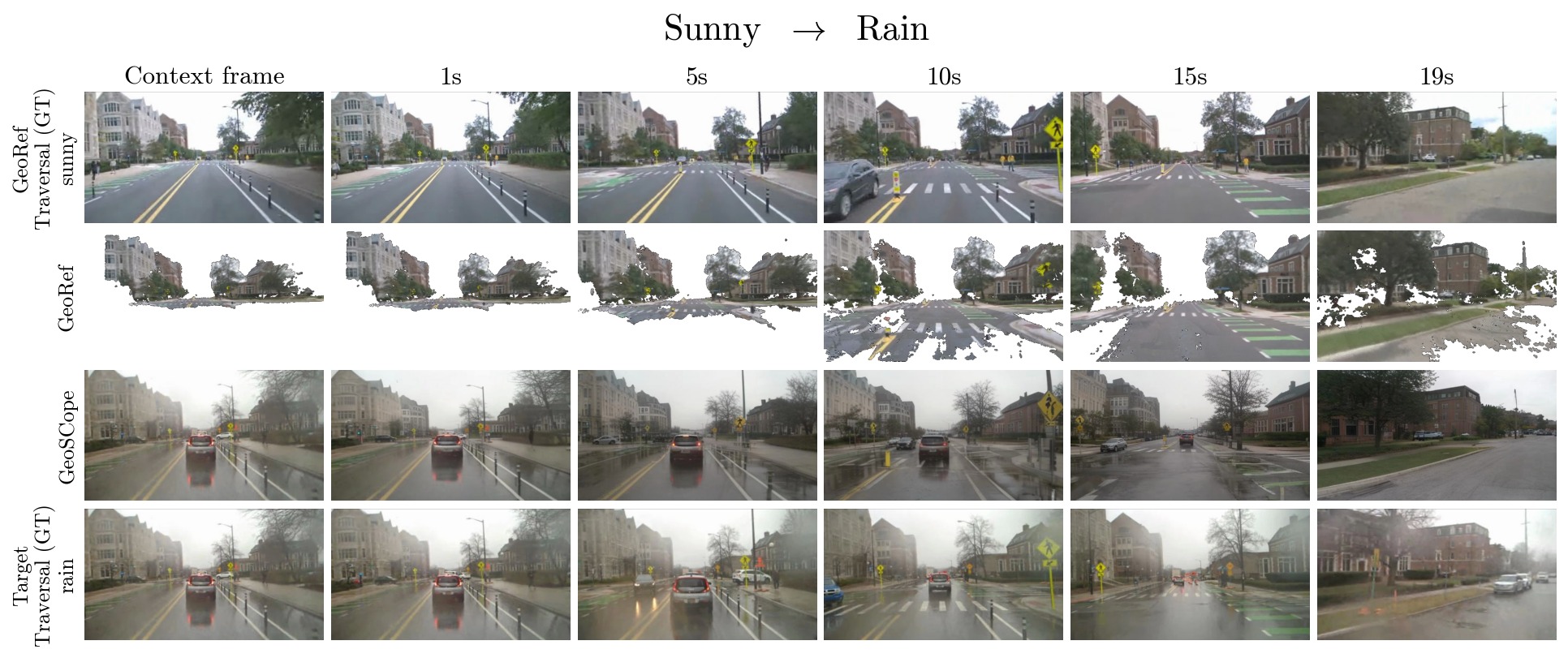}
    \caption{ Sunny map, rainy target drive. The rollout stays wet and hazy like the target drive and follows the mapped crosswalks, signs and buildings.}
    \label{fig:lightning2}
\end{figure}

\begin{figure}[ht!]
    \centering
    \includegraphics[width=0.9\linewidth]{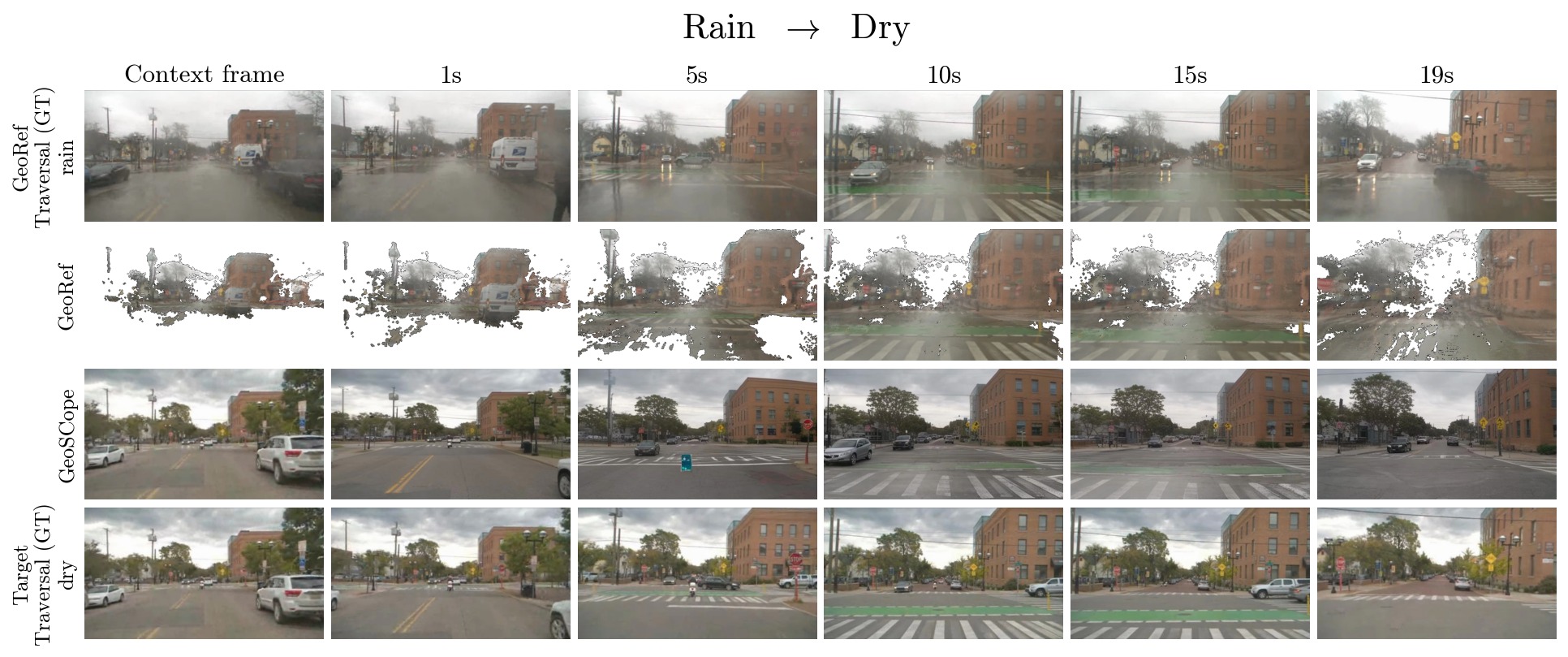}
    \caption{Rainy, foggy map, sunny target drive. GeoSCope produces a clear rollout and follows the mapped crosswalks, pedestrian sign and stone buildings through the intersection.}
    \label{fig:lightning3}
\end{figure}

\begin{figure}[ht!]
    \centering
    \includegraphics[width=0.9\linewidth]{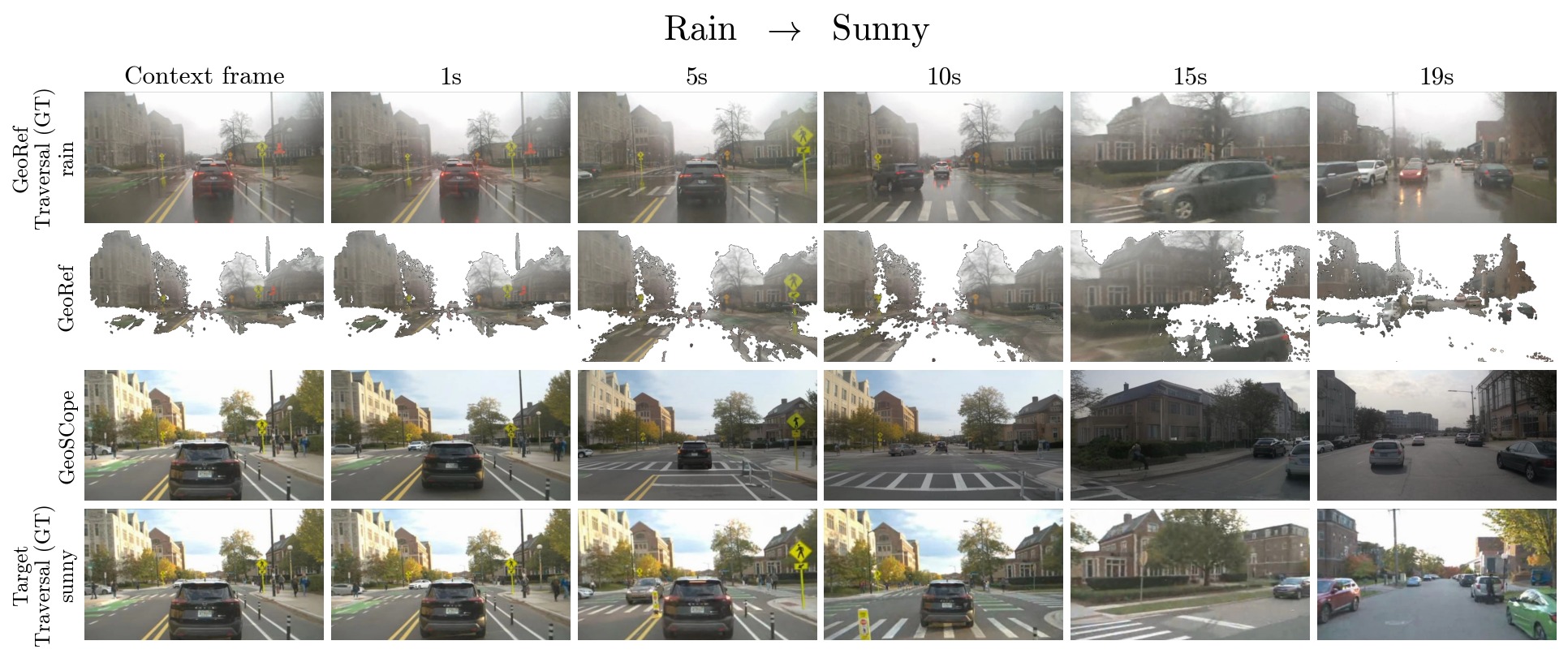}
    \caption{Rainy map with rain on the lens, dry overcast target drive. The rollout stays dry while following the mapped brick corner building, crosswalks, bike lane and signs.}
    \label{fig:lightning4}
\end{figure}

\stopcontents[appendices]

\end{document}